\documentclass[acmtog,nonacm]{acmart}

\usepackage{hhline}
\usepackage[ruled]{algorithm2e} %
\usepackage{subcaption}
\usepackage{pifont}
\usepackage{arydshln}
\usepackage{multirow}
\usepackage{multicol}
\usepackage{xcolor}
\usepackage[export]{adjustbox}
\usepackage[nor malem]{ulem}
\usepackage{cleveref}
\usepackage{xspace}
\usepackage{rotating}
\usepackage{xhfill}
\usepackage{tikz,bm}

\AtBeginDocument{%
  \providecommand\BibTeX{{%
    \normalfont B\kern-0.5em{\scshape i\kern-0.25em b}\kern-0.8em\TeX}}}

\setcopyright{acmcopyright}
\copyrightyear{2023}
\acmYear{2023}
\acmDOI{XXXXXXX.XXXXXXX}

\crefname{section}{Sec.}{Secs.}
\Crefname{section}{Section}{Sections}
\Crefname{table}{Table}{Tables}
\crefname{table}{Tab.}{Tabs.}

\definecolor{mydarkblue}{rgb}{0,0.08,1}
\definecolor{mydarkgreen}{rgb}{0.02,0.6,0.02}
\definecolor{mydarkorange}{rgb}{0.40,0.2,0.02}

\makeatletter
\DeclareRobustCommand\onedot{\futurelet\@let@token\@onedot}
\def\@onedot{\ifx\@let@token.\else.\null\fi\xspace}

\definecolor{mygreen}{RGB}{112,173,71}
\definecolor{myblue}{RGB}{91,155,213}
\definecolor{myorange}{RGB}{237,125,49}
\definecolor{myred}{RGB}{255,22,67}
\definecolor{honey}{RGB}{201,89,95}
\definecolor{amber}{RGB}{180,148,41}
\definecolor{darkblue}{RGB}{84,96,126}
\definecolor{darkgreen}{HTML}{008000}

\def\limitarrowmain#1{%
\begin{tikzpicture}
\draw[-{stealth[scale=4]}] (-7,0) to (-0.5,0);
\end{tikzpicture}}

\begin{document}

\title{ConceptLab: Creative Concept Generation using VLM-Guided Diffusion Prior Constraints}

\author{Elad Richardson}
\affiliation{%
  \institution{Tel Aviv University}
}

\author{Kfir Goldberg}
\affiliation{%
  \institution{Tel Aviv University, WSC Sports}
}

\author{Yuval Alaluf}
\affiliation{%
  \institution{Tel Aviv University}
}

\author{Daniel Cohen-Or}
\affiliation{%
  \institution{Tel Aviv University}
}

\begin{abstract}
Recent text-to-image generative models have enabled us to transform our words into vibrant, captivating imagery. The surge of personalization techniques that has followed has also allowed us to imagine unique concepts in new scenes. However, an intriguing question remains: How can we generate a \textit{new}, imaginary concept that has never been seen before?
In this paper, we present the task of \textit{creative text-to-image generation}, where we seek to generate new members of a broad category  (e.g., generating a pet that differs from all existing pets).
We leverage the under-studied Diffusion Prior models and show that the creative generation problem can be formulated as an optimization process over the output space of the diffusion prior, resulting in a set of ``prior constraints''.
To keep our generated concept from converging into existing members, we incorporate a question-answering Vision-Language Model (VLM) that adaptively adds new constraints to the optimization problem, encouraging the model to discover increasingly more unique creations. 
Finally, we show that our prior constraints can also serve as a strong mixing mechanism allowing us to create hybrids between generated concepts, introducing even more flexibility into the creative process.
\end{abstract}

\begin{teaserfigure}
\centering
{
\setlength{\tabcolsep}{0pt}
\begin{tabular}{c c@{\hspace{4pt}} c c@{\hspace{4pt}} c c@{\hspace{4pt}} c c}
\includegraphics[width=0.115\textwidth]{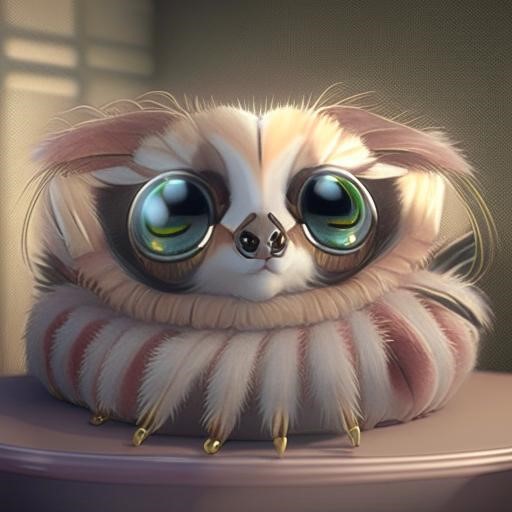}
&
\includegraphics[width=0.115\textwidth]{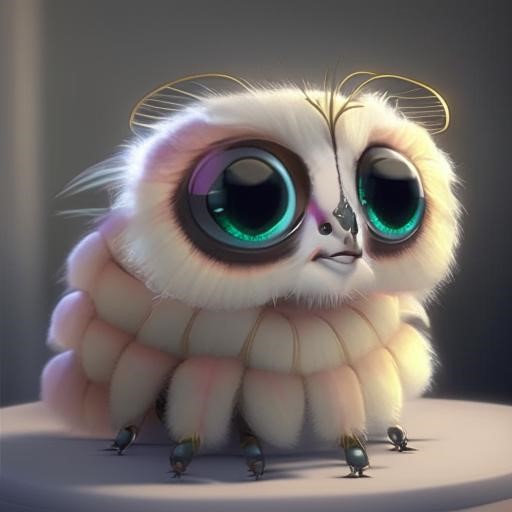}
&
\includegraphics[width=0.115\textwidth]{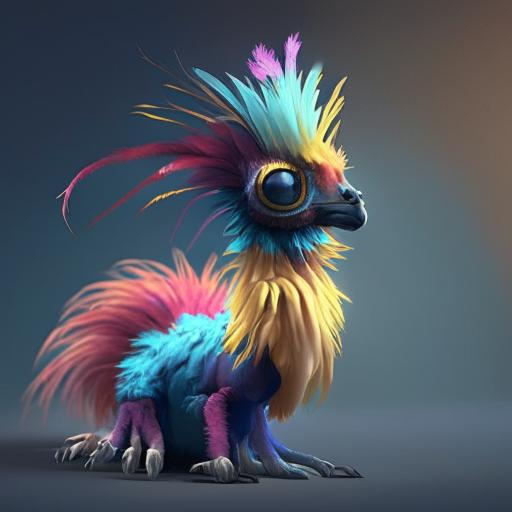}
&
\includegraphics[width=0.115\textwidth]{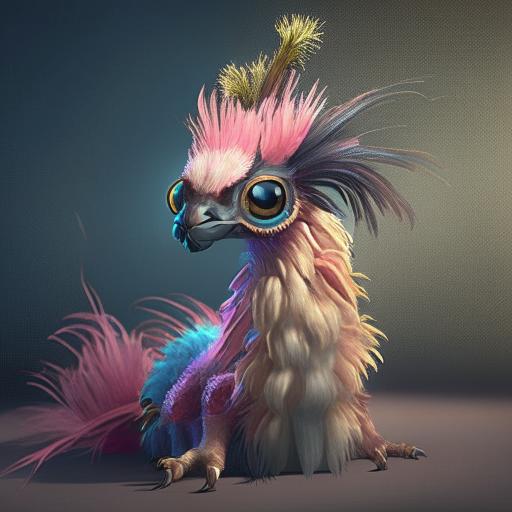}
        &
\includegraphics[width=0.115\textwidth]{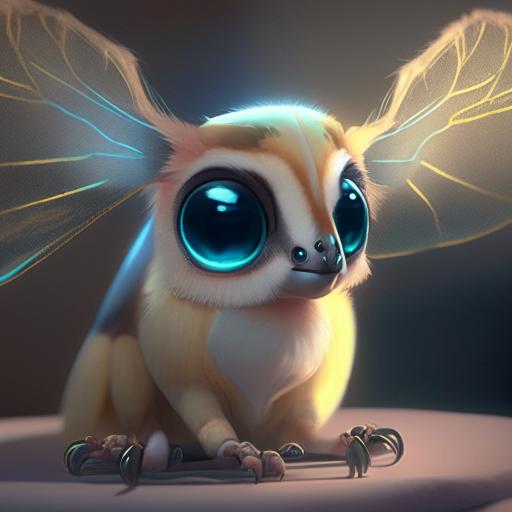}
&
\includegraphics[width=0.115\textwidth]{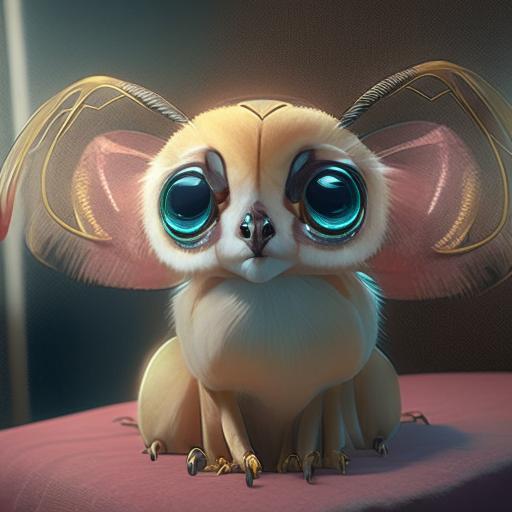}
&
\includegraphics[width=0.115\textwidth]{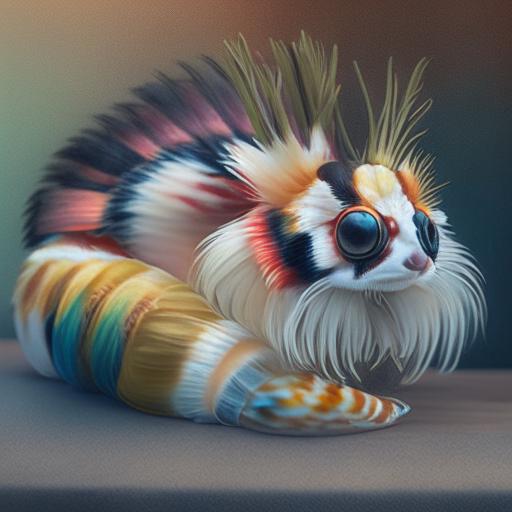}
&
\includegraphics[width=0.115\textwidth]{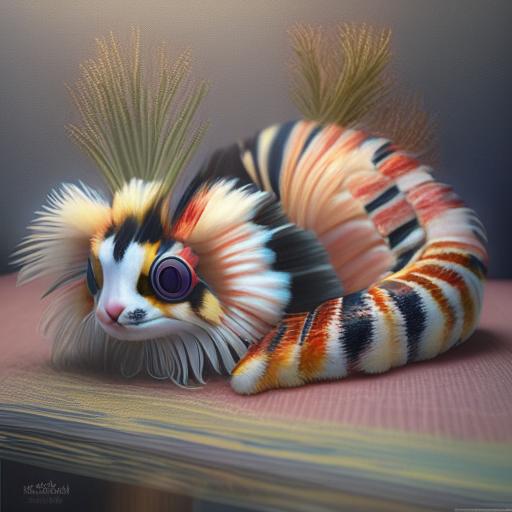} \\   
\includegraphics[width=0.115\textwidth]{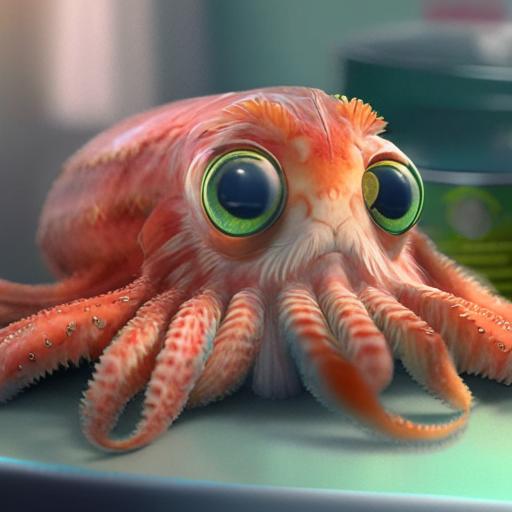}
&
\includegraphics[width=0.115\textwidth]{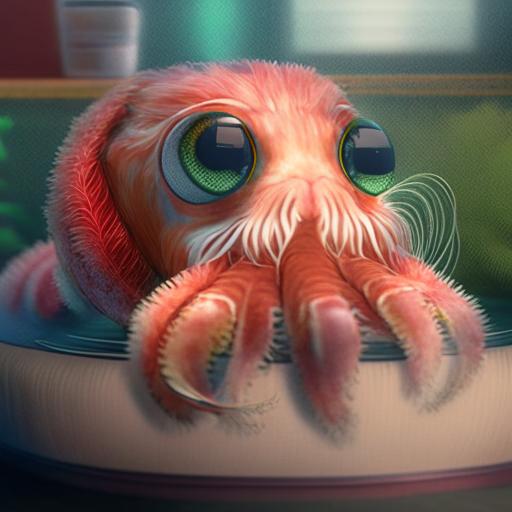}
&
\includegraphics[width=0.115\textwidth]{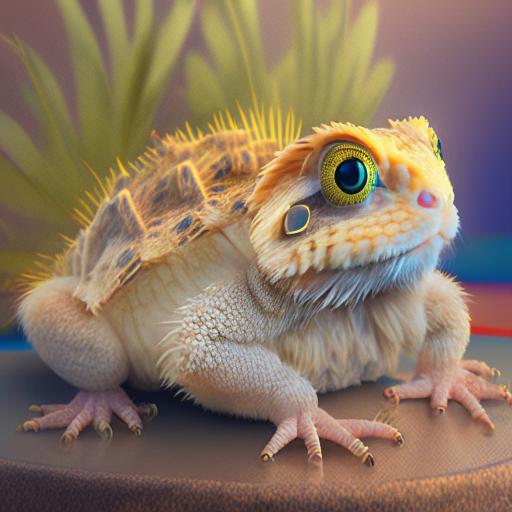}
&
\includegraphics[width=0.115\textwidth]{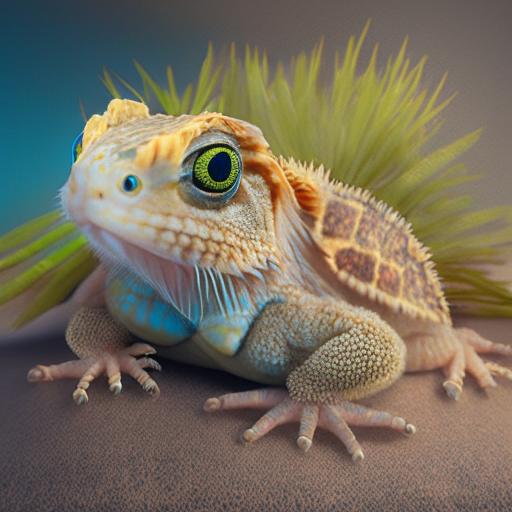}
&        
\includegraphics[width=0.115\textwidth]{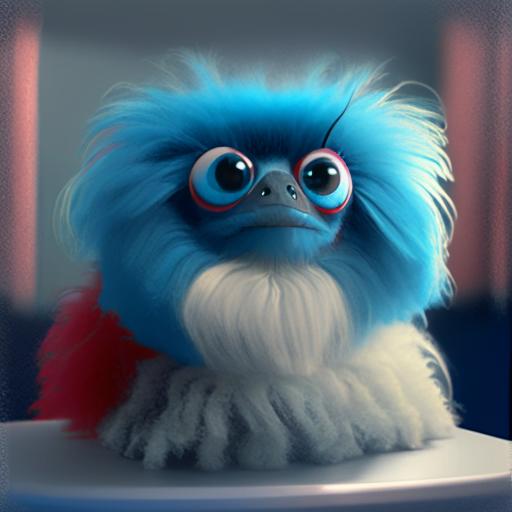}
&
\includegraphics[width=0.115\textwidth]{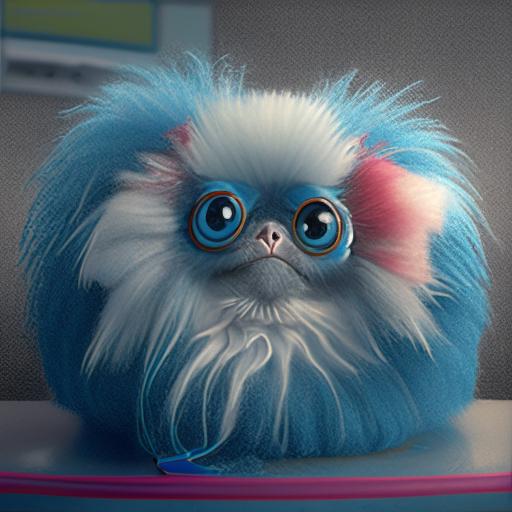}
&          
\includegraphics[width=0.115\textwidth]{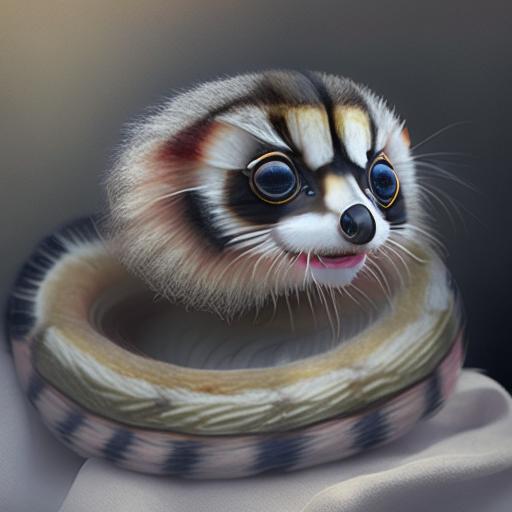}
&
\includegraphics[width=0.115\textwidth]{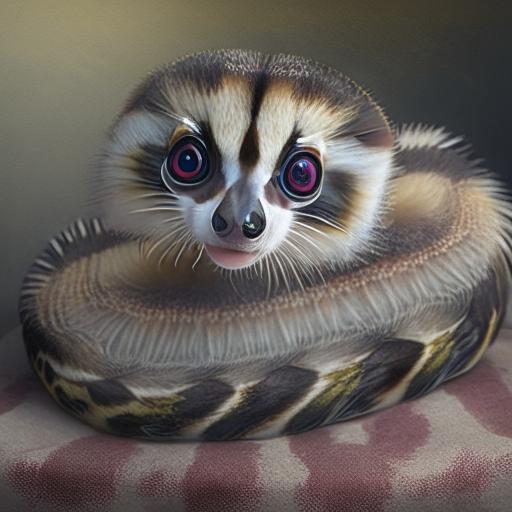}   
\end{tabular}
}
\caption{New ``pets'' generated using ConceptLab. Each pair depicts a learned concept that was optimized to be unique and distinct from existing members of the pet category. Our method can generate a variety of novel concepts from a single broad category.}
\Description{figure description}
\label{fig:teaser}
\end{teaserfigure}

\maketitle
\vspace{-0.2cm}
\section{Introduction}~\label{sec:intro}
The quest for creative generation in computer graphics has sparked the study of computational creativity~\cite{xu2012fit,cohen2016inspired,hertzmann2018can, sims1991artificial,sims1994evolving}, which involves algorithms that simulate creative behaviors or try to enhance and augment the human creative process. Thanks to the rapid advancements in powerful text-to-image generative models, we now have an unprecedented ability to transform language into incredible, diverse images~\cite{ramesh2022hierarchical,nichol2021glide,rombach2022high,saharia2022photorealistic,balaji2023ediffi,kandinsky2,ding2022cogview2}, opening up new possibilities for generating creative content.
Building on these models, recent personalization techniques~\cite{gal2023image,ruiz2022dreambooth,kumari2022customdiffusion,gal2023encoderbased,wei2023elite} have also enabled us to create personalized concepts and incorporate them into the generative process. Yet, an interesting question remains: can we use these powerful models to generate a novel creative concept that was not explicitly described to the model?

\begin{figure*}
    \centering
    \includegraphics[width=0.95\textwidth]{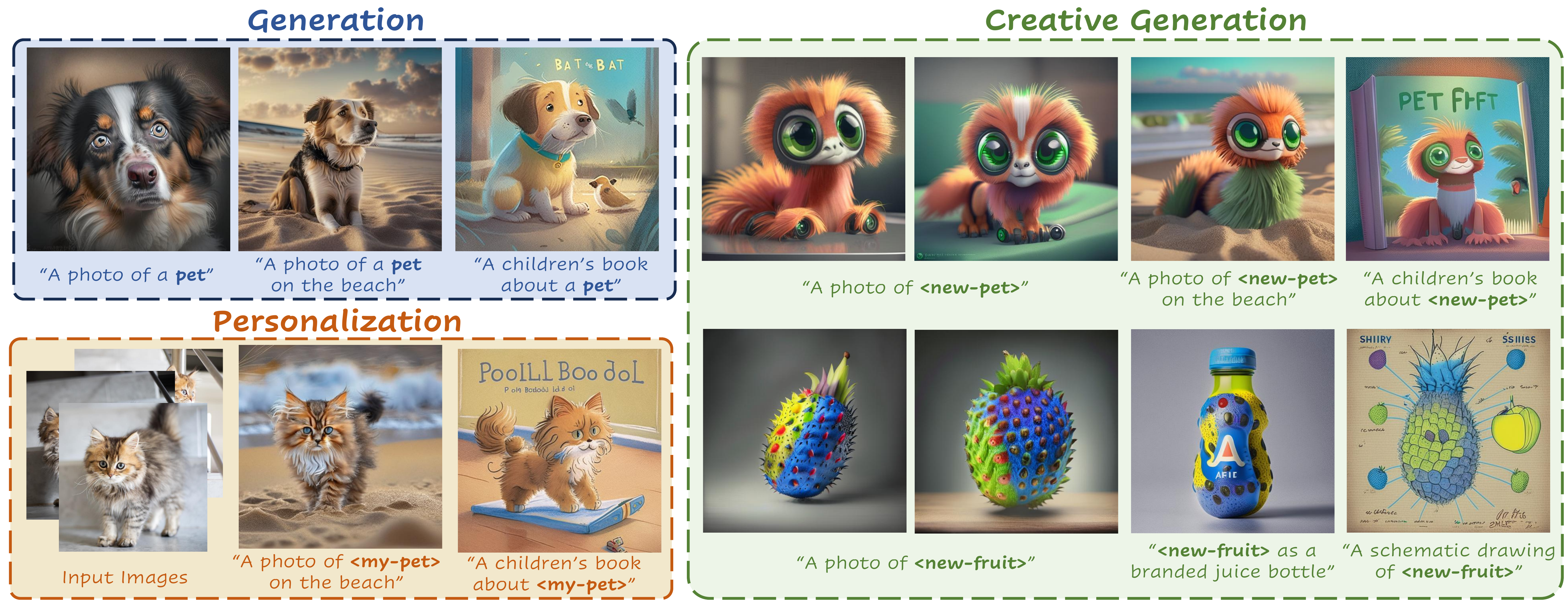}
    \vspace{-0.3cm}
    \caption{In text-guided generation (top left), an image is created given a free-form text prompt. With personalization methods (bottom left), we can learn new tokens representing a specific concept or subject. Our creative generation method (right) learns tokens that represent novel concepts belonging to a given category (e.g., ``a pet'' or ``a fruit''). The learned concepts are optimized to belong to the broad category while differing from existing members of that cateogry.
    }
    \label{fig:detailed_teaser}
\end{figure*}

In this paper, we tackle the task of \textit{creative text-to-image generation} using diffusion models. Specifically, we seek to generate novel and creative members of a given broad category. Consider, for example, the category of all ``pets''. Here, we would like to find a new concept that visually resembles a pet, but differs from any existing pet. For example, in~\Cref{fig:teaser}, we show generated concepts that semantically resemble a pet, but do not belong to a specific species. All these results were generated by only specifying the target category, resulting in a variety of possible outcomes.

Inspired by token-based personalization~\cite{cohen2022my,gal2023image}, we represent our new concept as a token in the text encoder of a pretrained generative model. However, to generate a new concept, we cannot simply apply a standard inversion scheme as we naturally do not have any images depicting the target subject. Instead, we turn to the CLIP vision-language model~\cite{radford2021learning} to help guide our optimization process. In essence, we divide our constraints into a set of positive and negative constraints. The positive constraint is introduced to encourage the generation of images that still match the broad category. Conversely, the negative constraints represent existing members of the category we wish to shift away from. Considering our previous pet example, the positive constraint is defined by the word ``pet'' while the negative constraints may consist of words such as ``cat'' and ``dog'', indicating that we wish to generate a pet that is not a cat nor a dog. Applying these constraints together should ideally encourage the learned concept to reside inside the category, but differ from the specified members.

While conceptually simple, it is not clear how to apply our CLIP-based optimization in practice in the context of diffusion models.
First, applying a CLIP loss during the diffusion denoising process requires an approximation of the output image, which was shown to be unstable without applying dedicated augmentations~\cite{avrahami2022blended_latent}, or a dedicated noise-aware CLIP model~\cite{nichol2021glide}. Second, we do not have a set of reference images that can be directly denoised during the optimization process, further complicating the process.
A key understanding in our approach is that our constraints can be better represented when used with a Diffusion Prior model~\cite{ramesh2022hierarchical}. Specifically, we show that the output space of the Diffusion Prior serves as a more suitable target space for our optimization task. As such, we optimize our learned token by applying our CLIP constraints over the outputs of the Diffusion Prior, resulting in a set of \textit{``prior constraints''}. 

While we now have a working optimization framework, another challenge remains. For our negative constraints, we should ideally specify all existing members of the given category (e.g., all types of pets). However, doing so is cumbersome and not always practical. Instead, we build upon recent question-answering VLMs~\cite{li2023blip} to iteratively suggest additional category members. This is achieved by dividing the optimization problem into segments. After each segment, we generate an image using our current concept token and then query the VLM to textually describe what member of the given category is depicted in the image. This technique allows us to ``project'' the current concept into the space of existing category members, as each member already has a unique word describing it. The new word is then added to our set of negative constraints, allowing us to gradually shift away from a growing set of category members, resulting in more creative generations.

Finally, we show that our proposed \textit{prior constraints} can also be used to mix up generated concepts and create new hybrids by using a set of positive constraints derived from the generated concepts. 
This allows us to extend and evolve the newly generated concepts.
The flexibility of our prior constraints and iterative optimization scheme is demonstrated using both quantitative and qualitative evaluation, showing its effectiveness for creative generation.

\section{Related Works}

\paragraph{Text-Guided Synthesis.} 
Recently, large-scale text-to-image diffusion models~\cite{ho2020denoising,nichol2021improved,dhariwal2021diffusion} have achieved an unprecedented ability to generate high-quality imagery guided by a text prompt~\cite{ramesh2022hierarchical,nichol2021glide,rombach2022high,saharia2022photorealistic,balaji2023ediffi,kandinsky2}. Leveraging these powerful generative models, many have attempted to utilize such models for downstream editing tasks~\cite{meng2022sdedit,hertz2023prompttoprompt,kawar2023imagic,Tumanyan_2023_CVPR,parmar2023zero,couairon2023diffedit}. Most text-guided generation techniques condition the model directly on embeddings extracting from a pretrained text encoder~\cite{hertz2023prompttoprompt,chefer2023attendandexcite,poole2023dreamfusion,avrahami2022blended_latent,brooks2022instructpix2pix}. In this work, we utilize a Latent Diffusion Model~\cite{rombach2022high} paired with a \textit{Diffusion Prior} model~\cite{ramesh2022hierarchical,kandinsky2}.

\paragraph{Diffusion Prior.} 
A \textit{Diffusion Prior} model, introduced in \citet{ramesh2022hierarchical}, is tasked with mapping an input text embedding to its corresponding image embedding in CLIP's~\cite{radford2021learning} latent space. A decoder is then trained to generate a corresponding image, conditioned on the CLIP image embedding. In \citet{ramesh2022hierarchical} the authors demonstrate that applying a Prior and conditioning over the resulting image embeddings attains improved diversity while enabling image variations and interpolations.
Several works have adopted the use of a Prior for text-guided video synthesis~\cite{singer2023makeavideo,esser2023structure} and 3D generation and texturing~\cite{xu2023dream3d,mohammad2022clip}. The use of Diffusion Prior for text-guided synthesis is further analyzed in~\cite{aggarwal2023controlled,zhou2023shifted}.

\begin{figure*}
    \centering
    \includegraphics[width=0.95\textwidth]{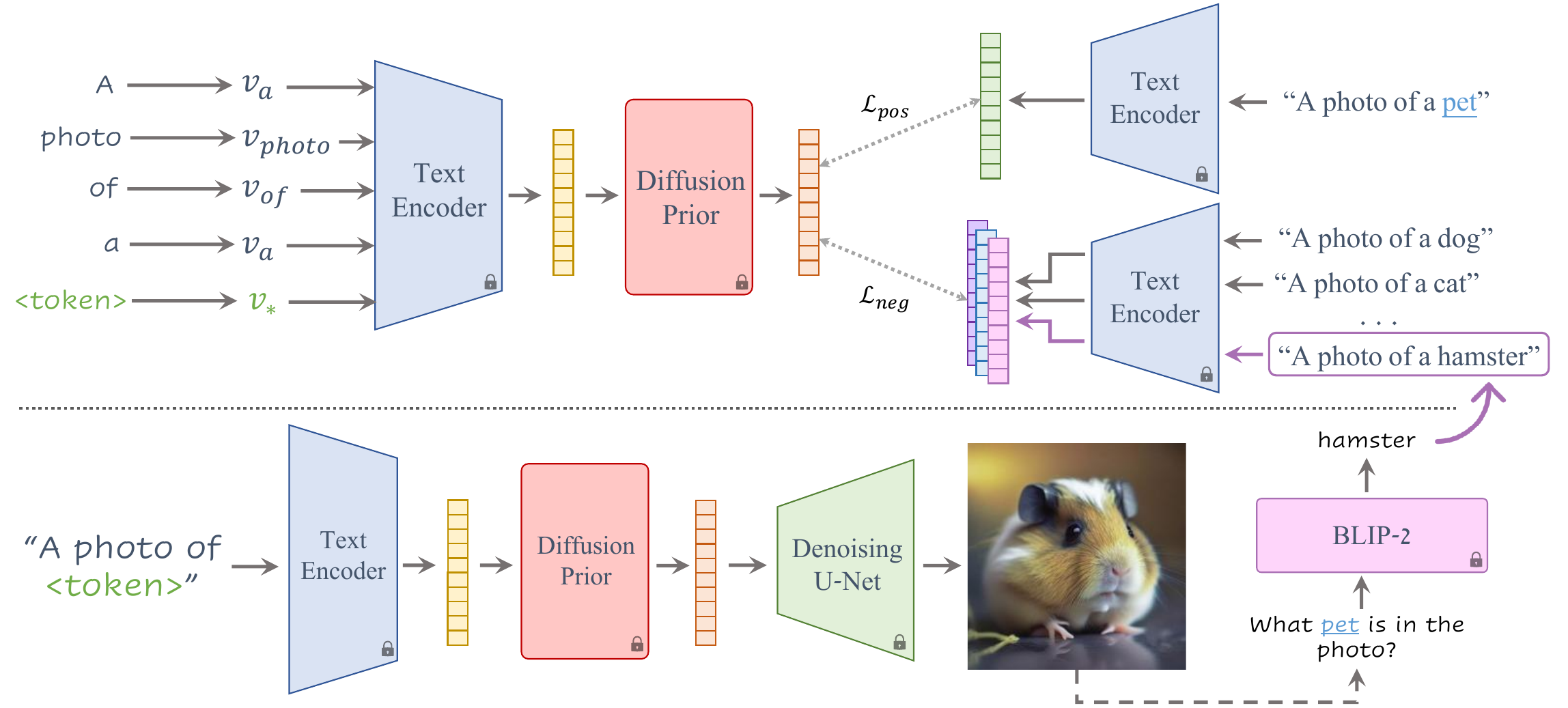}
    \\[-0.2cm]
    \caption{ConceptLab overview.
    We optimize a single embedding $v_*$ representing the novel concept we wish to generate (e.g., a new type of ``pet''). 
    To do so, we compute a set of losses encouraging the learned embedding to be similar to that of a given category while being different from a set of existing members (e.g., a ``dog'' or a ``cat''). 
    To gradually generate more unique creations, during training, we query a pretrained BLIP-2 VQA model~\cite{li2023blip} to expand the set of negative constraints based on the currently generated novel concept (e.g., we add the token ``hamster'' to shift our embedding from generating images resembling a ``hamster''). 
    }
    \label{fig:method}
\end{figure*}

\paragraph{Personalization.}
In the task of personalization~\cite{cohen2022my,gal2023image}, we aim to inject new user-specific concepts into a pretrained generative model. In the context of text-guided synthesis, doing so should allow for the generation of novel images depicting the target subject or artistic style using an input text prompt. To teach the generative model new concepts, current personalization techniques either optimize a set of text embeddings~\cite{gal2023image,voynov2023p,alaluf2023neural}, fine-tune the denoising network~\cite{ruiz2022dreambooth,kumari2022customdiffusion,tewel2023keylocked}, or train an encoder to map a concept to its textual representation~\cite{gal2023encoderbased,shi2023instantbooth,wei2023elite}. Deviating from existing personalization literature, we do not aim to teach the generative model a new subject or concept. Instead, we focus on the task of \textit{Creative Generation} and generate novel concepts, see~\Cref{fig:detailed_teaser}. 

\paragraph{Creative Generation.}
A long-standing question in computer graphics centers around whether computers can truly generate creative art~\cite{hertzmann2018can}. Naturally, generating creative content can be tackled in many different ways. \citet{xu2012fit} propose a set-evolution method for creative 3D shape modeling which aims to offer the user creative shapes that fit his preferences while still offering diversity. \citet{elgammal2017can} explore creative generation in the context of GANs~\cite{goodfellow2020generative} and learn new styles by maximizing the deviation from existing artistic styles using discriminators. \citet{sbai2018design} introduce a novel loss encouraging deviation from existing styles found in the training set. 

Some works also approach the creative generation task as a composition task, learning and fusing fine-level components into a complete creation. This has been demonstrated across various creative domains including sketching~\cite{ge2021creative} and 3D Modeling~\cite{ranaweera2016exquimo}. Recently \citet{vinker2023concept} have shown that one can decompose personalized concepts into their different visual aspects which can then be joined together in novel and creative ways.
We choose to approach creative generation by finding novel concepts that are optimized to match a given category while differing from existing concepts in that category. This allows us to generate novel and diverse concepts from that category without directly describing their look.

\section{Preliminaries}
Our creative generation scheme is built on top of the Kandinsky 2 model~\cite{kandinsky2}. This model combines the idea of a Latent Diffusion Model proposed in \cite{rombach2022high} with a Diffusion Prior model~\cite{ramesh2022hierarchical} allowing us to introduce constraints over the Diffusion Prior outputs.

\paragraph{Latent Diffusion Models.}
In a Latent Diffusion Model (LDM), the diffusion process is performed within the latent space of an autoencoder. First, an encoder $\mathcal{E}$ is trained to map a given image $x\in\mathcal{X}$ into a latent code $z = \mathcal{E}(x)$ while a decoder $\mathcal{D}$ is simultaneously tasked with reconstructing the original input image such that $\mathcal{D}(\mathcal{E}(x)) \approx x$.
Given the autoencoder, a denoising diffusion probabilistic model (DDPM)~\cite{sohl2015deep,ho2020denoising} is trained to produce latent codes within this learned latent space. During the denoising process, the diffusion model can be conditioned on an additional input vector. 
The DDPM model is trained to minimize the objective given by:
\begin{equation}~\label{eq:ldm}
    \mathcal{L} = \mathbb{E}_{z,y,\varepsilon,t} \left [ || \varepsilon - \varepsilon_\theta(z_t, t, c) ||_2^2 \right ].
\end{equation}
The denoising network $\varepsilon_\theta$ is tasked with correctly removing the noise $\varepsilon$ added to the latent code $z_t$, given $z_t$, the current timestep $t$, and the conditioning vector $c$. 

\paragraph{Diffusion Prior.}
Diffusion models are commonly trained with the conditioning vector $c$ directly derived from the CLIP~\cite{radford2021learning} text encoding of a given text prompt, $y$. In \citet{ramesh2022hierarchical}, it was proposed to decompose the generative text-to-image problem into two steps. First, an image embedding is predicted from a given text prompt, using a Diffusion Prior model. Next, the image embedding is fed into a diffusion decoder trained to generate an image conditioned on the image embedding. 

Training is typically done in two independent steps. The diffusion decoder is trained using the objective defined in~\Cref{eq:ldm} with an image embedding as the conditioning $c$.
The Diffusion Prior model, $P_\theta$, is then tasked with directly predicting the denoised image embedding $e$ from a noised embedding $e_t$:
\begin{equation}\label{eq:prior}
    \mathcal{L}_{prior} = \mathbb{E}_{e,y,t} \left [ || e - P_\theta (e_t, t, y) ||_2^2 \right ] .
\end{equation}
Once the two models are trained, each on its objective, they can be put together to create a complete text-to-image pipeline. This two-stage approach was shown to improve image diversity, but more importantly from our context, it provides direct access to an intermediate CLIP image embedding and allows introducing constraints directly in that space. We show the output space of the Diffusion Prior to be more effective than applying a constraint on a standard diffusion model or directly on the CLIP text embeddings.

\section{Method}
At its core, our method, dubbed ConceptLab, aims to tackle the creative generation task where we wish to learn a token representing a novel, never-before-seen concept belonging to a general category that differs from any existing concepts within that category. Similar to Textual Inversion~\cite{gal2023image}, we do so by optimizing a new embedding vector $v_*$ representing our novel concept in the text conditioning space of a pretrained text-to-image model.
As we seek to generate novel concepts that do not exist, optimizing this representation using a reconstruction-based objective is not possible. Instead, we impose a set of constraints over our learned representation where the embedding $v_*$ is optimized to be similar to a given broad category while differing from existing members of that category. As shall be discussed, we choose to apply this optimization scheme using a set of ``prior constraints'' (see~\Cref{sec:diffusion_prior}). During training, we gradually expand the set of constraints using VLM-Guidance (see~\Cref{sec:live_negatives}), encouraging the creation of more unique concepts over time. Our complete training scheme is illustrated in~\Cref{fig:method}. At inference, compositions of our novel concept can be generated by adding the optimized token to an input prompt, see~\Cref{fig:teaser,fig:our_results,fig:creative_art}.

\subsection{Diffusion Prior Constraints}
\label{sec:diffusion_prior}

\paragraph{The Constraints.} We define our prior constraints as a set of losses applied over the output space of a Diffusion Prior model. These constraints are divided into a set of positive constraints $\mathcal{C}_{pos}$ and negative constraints $\mathcal{C}_{neg}$, where each constraint is defined using textual tokens. For example, to generate a new member of the ``pet'' category, our positive constraints could be simple defined as $\mathcal{C}_{pos}=\{\text{pet}\}$ with $\mathcal{C}_{neg}=\{\text{cat},\text{dog},\dots,\text{hamster}\}$ as the negative constraints.

\paragraph{The Objective.} Given our two sets of constraints, we next define a measurement of similarity between $v_*$ and each constraint. We first incorporate $v_*$ and each constraining word $c$ into the same randomly sampled prompt template $y$ (e.g., ``A photo of a \{\}'', ``An oil painting of \{\}''). Each such sentence can now be encoded into a CLIP text embedding, an operation we denote as $E_\text{y}(c)$, and defines a textual constraint. 
Given the textual constraints, a simple approach for defining the similarity to $v_*$ would be to compute the cosine similarity between $E_y(v_*)$ and each textual constraint $E_y(c)$. We instead show that it is preferable to pass $E_y(v_*)$ through the Diffusion Prior model before computing the similarity measure.
Intuitively, passing a text prompt through the Diffusion Prior results in a specific instance of the prompt. For example, applying the prior on ``A photo of a dog'' would result in a specific image of a specific dog breed. By passing $E_\text{y}(v_*)$ through the prior we encourage all realizations of $v_*$ to align with the textual constraints, resulting in more consistent generations. 
Conversely, we choose \textit{not} to pass the positive and negative constraints through the Diffusion Prior. This is motivated by the intuition that we want to ideally keep the constraints themselves as broad as possible. That is, instead of applying the constraints over a specific image of a ``cat'' or ``dog'', we wish to shift away from the set of all possible ``cats'' and ``dogs''.

Thus our loss objective is defined as: 
\begin{align}~\label{eq:objective}
\begin{split}
    \mathcal{S}(\mathcal{C},v_*)  = \mathbb{E}_{c\sim\mathcal{C}} \left [ \langle E_\text{y}(c), P(E_\text{y}(v_*)) \rangle \right ]  \\
    \mathcal{L}  = \mathcal{S}(\mathcal{C}_{neg},v_*)+\lambda(1-\mathcal{S}(\mathcal{C}_{pos},v_*))
\end{split}
\end{align}
In words, we encourage every sampled image embedding $P(E_\text{y}(v_*))$ generated from our learned embedding $v_*$ to distance itself from the text constraints defined by $C_{neg}$ while staying close to those of $C_{pos}$, with $\lambda$ allowing us to control the balance between the two.  

\paragraph{Regularizations.} When the set of constraints becomes large, the penalty for collapsing to a specific member of the constraint becomes increasingly more negligible. To avoid such a collapse, we use an additional objective that measures the \textit{maximal} similarity to the negative constraints:
\begin{align}~\label{eq:objective_2}
\begin{split}
    \mathcal{S}_{max}(\mathcal{C},v_*)  = \max_{c\sim\mathcal{C}} \left ( \langle E_\text{y}(c), P(E_\text{y}(v_*)) \rangle \right ).
\end{split}
\end{align}
This similarity measure is incorporated into~\Cref{eq:objective}, by averaging it with $\mathcal{S}(\mathcal{C},v_*)$, ensuring that the constraint that is closest to $v_*$ receives a greater penalty.

Finally, we also restrict the similarity measure between two predetermined similarity values to avoid pathological solutions. For example, we empirically find that without such a restriction the model starts to generate text in the image that matches the target category, as a way to obtain high similarity without actually generating the desired concept.

\paragraph{Using the Constraints.}
In the context of creative generation, we set the positive constraints, $C_{pos}$, to contain a single broad category, e.g., \{"pet"\}, and set the negative constraints either manually, or automatically through our \textit{adaptive negatives} scheme, introduced below.
An additional application enabled through our constraints is that of \textit{concept mixing}, where we wish to fuse existing concepts into a single creation. To this end, we can define a set of positive constraints with no negative constraints, see~\Cref{fig:mixing_real_concepts}.

\subsection{Adaptive Negatives with VLM-Guidance}
\label{sec:live_negatives}

Ideally, we would like to apply a large set of negative constraints in order to encourage the generation of truly unique creations. Yet, manually defining a large set of negative constraints is both cumbersome and may not accurately represent the most relevant members of the broad category.
To this end, we propose an adaptive scheme to gradually expand the set of negative constraints during training using guidance from a VLM. 
As illustrated at the bottom of~\Cref{fig:method}, at regular intervals during the optimization process (e.g., 250 steps) we generate an image using our current representation. We then query a pretrained BLIP-2 VLM~\cite{li2023blip} and ask the model to identify which member of the broad category is currently present in the image. 
We then add the resulting instance to the set of negative constraints for the rest of the training. Note that we always incorporate the target category (e.g., ``pet'') as part of the question (e.g., ``What kind of \textit{pet} is in this photo'') to encourage the VLM to respond with members of that category. 
This adaptive scheme not only shifts the learned concepts away from existing members but also results in diverse creations across different seeds as each training seed may add a different set of negative classes or change the order in which they are added, see~\Cref{fig:creative_results_original_token}. While it is possible to use a Large Language Model (LLM) to automatically generate a list of negative constraints, we found that the optimization yielded better results when constraints were incrementally added based on the specific concepts that emerged during the optimization process.
\begin{figure}
    \centering
    \setlength{\tabcolsep}{1pt}
    \renewcommand{\arraystretch}{0.7}
    {\footnotesize
    \begin{tabular}{c c c c c c}

        &
        \textcolor{red}{\textendash \mbox{} cat} &
        \textcolor{red}{\textendash \mbox{} guinea pig} &
        \textcolor{red}{\textendash \mbox{} parrot} &
        &
        Result \\ 

        \includegraphics[width=0.085\textwidth]{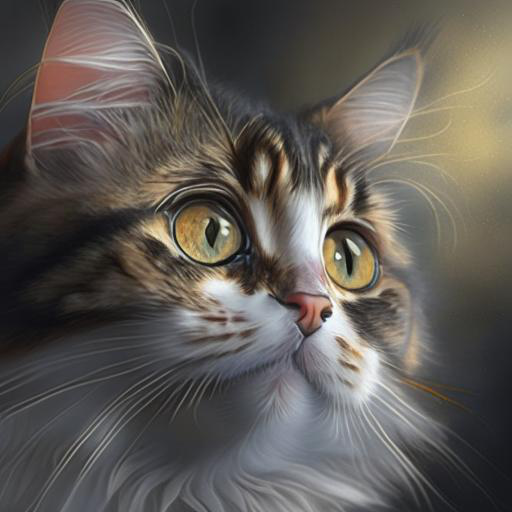} &
        \includegraphics[width=0.085\textwidth]{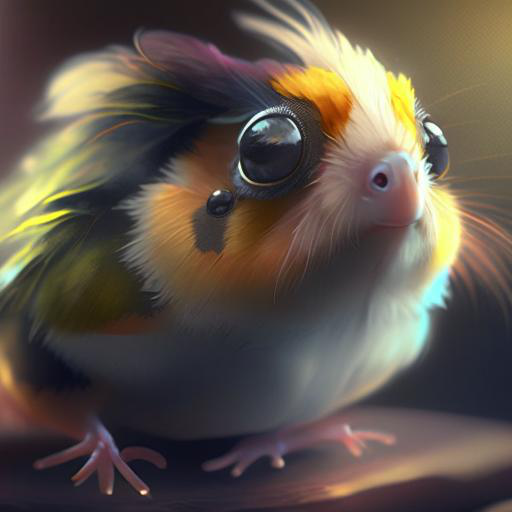} &
        \includegraphics[width=0.085\textwidth]{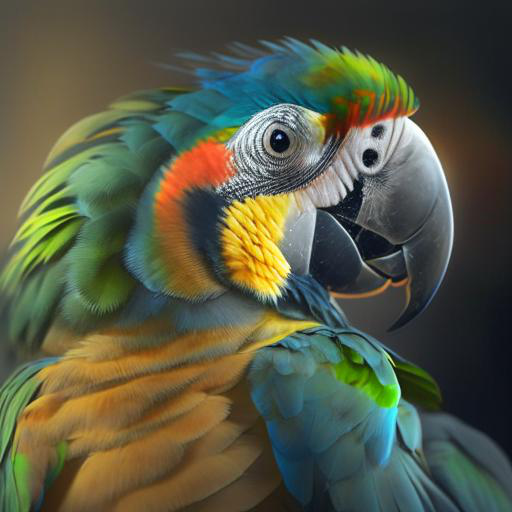} &
        \includegraphics[width=0.085\textwidth]{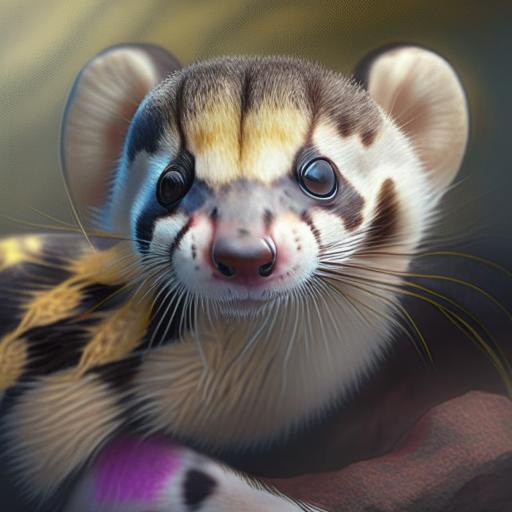} &
        \raisebox{0.3in}{...} &
        \includegraphics[width=0.085\textwidth]{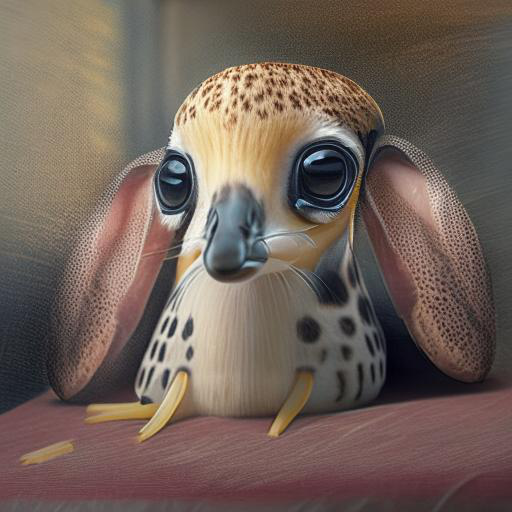} \\

        &
        \textcolor{red}{\begin{tabular}{c} \textendash \mbox{} oil \\ painting
        \end{tabular}} &
        \textcolor{red}{\begin{tabular}{c} \textendash \mbox{} colorful \\ abstract
        \end{tabular}} &
        \textcolor{red}{\begin{tabular}{c} \textendash \mbox{} black \\ and white
        \end{tabular}} &
        &
        Result \\

        \includegraphics[width=0.085\textwidth]{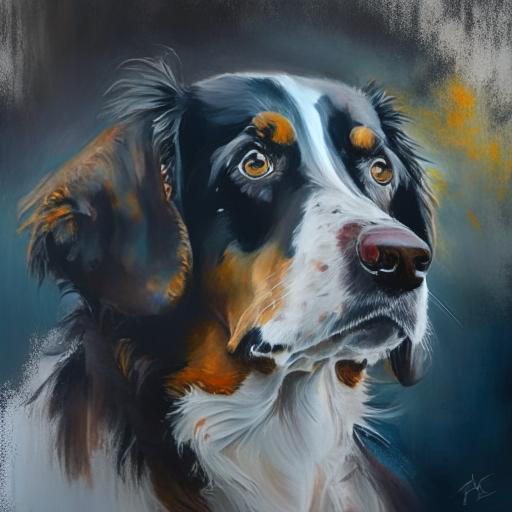} &
        \includegraphics[width=0.085\textwidth]{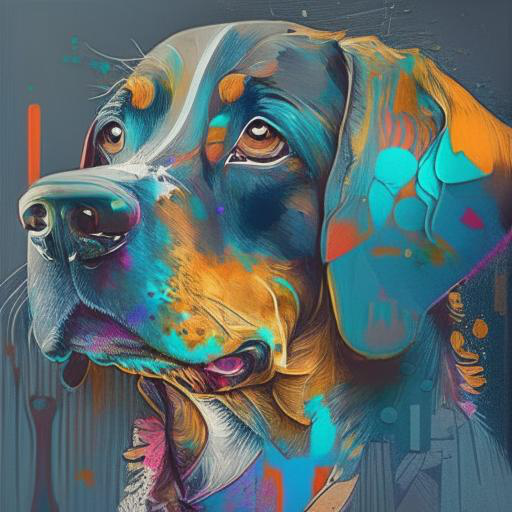} &
        \includegraphics[width=0.085\textwidth]{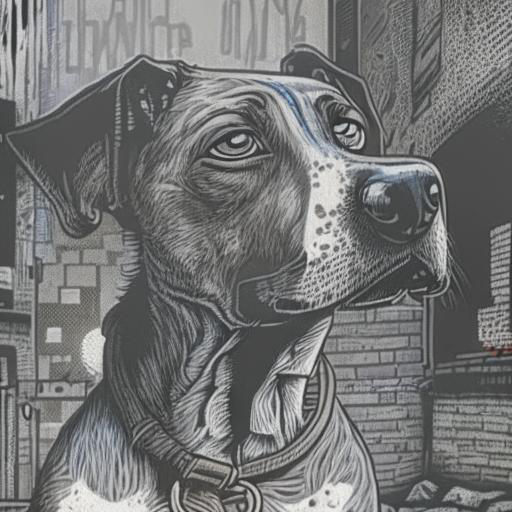} &
        \includegraphics[width=0.085\textwidth]{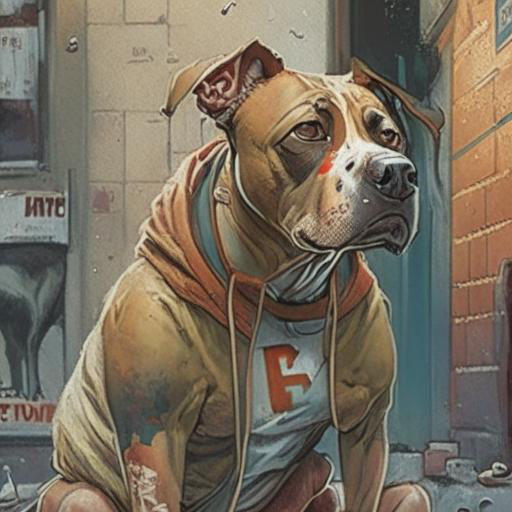} &
        \raisebox{0.3in}{...} &
        \includegraphics[width=0.085\textwidth]{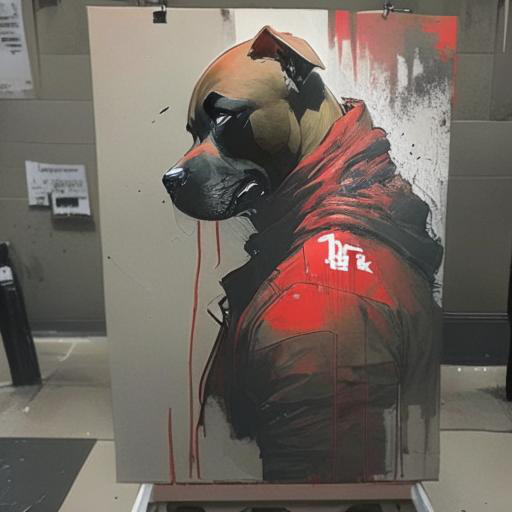} \\

        \multicolumn{6}{c}{\raisebox{0.05cm}{\limitarrowmain{}}} \\

    \end{tabular}
    
    }
    \vspace{-0.4cm}
    \caption{
    During training, we use BLIP-2 to infer the closest word to our current concept, which is then added to our constraints.
    }
    
    \label{fig:intro_live_negatives}
\end{figure}

\subsection{Evolutionary Generation}
\label{sec:evolution}

Building on our prior constraints, we show that one can also fuse generated concepts into a new concept. 
To perform \textit{concept mixing} over a given set of concepts we first generate a set of images from each concept, creating a set of image constraints, $C_{im}$. Each image is then passed through a CLIP image encoder, $E_{im}(c)$ to create a set of image embeddings. We then apply a modified loss that pushes a learnable concept $v_{mix}$ closer to the given embeddings,
\begin{align}~\label{eq:mix}
\begin{split}
    \mathcal{L_\text{mix}}  =  1 - \mathbb{E}_{c\sim\mathcal{C}_{im}} \left [ \langle E_{im}(c), P(E_\text{y}(v_{mix})) \rangle \right ].
\end{split}
\end{align}
This objective can be applied over either generated concepts or real images and can also be iteratively applied to create hierarchical generations of creative creatures. An optional weight term can additionally be applied to better control the effect of each concept on the generated output.

\section{Implementation Details}
\label{sec:additional_details}
We operate over the official implementation of the Kandinsky 2.1 text-to-image model~\cite{kandinsky2}. The Kandinsky model uses the CLIP ViT-L/14 model~\cite{radford2021learning}, alongside an extended multilingual CLIP ViT-L/14 text encoder, introduced to allow multilingual generation.
We use the extended text encoder for our textual constraints as we found it to be empirically more effective than the standard one.
Training is performed on a single GPU for up to $2500$ training steps using a batch size of $1$ and a fixed learning rate of $0.0001$. Each optimization step takes about 0.2 seconds, where a BLIP-guidance step takes about 8 seconds. We manually stop the optimization process when BLIP is unable to correctly classify the generated concept.
Unless otherwise noted, we initialize our learned token embedding using the token of our positive concept (e.g., ``pet'').
To balance the positive and negative constraints in~\Cref{eq:objective}, we set $\lambda=1$.
When using our adaptive negatives technique, we query the BLIP model every $250$ training steps, which was empirically determined to give the optimization process a sufficient number of iterations to alter the generated result.

\begin{figure}
    \centering
    \setlength{\tabcolsep}{1pt}
    {\footnotesize
    \begin{tabular}{c c c c c}

        \raisebox{0.4in}{\multirow{1}{*}{
            \rotatebox{90}{\begin{tabular}{c} \textcolor{mygreen}{\begin{tabular}{c} \\ + plant \end{tabular}}  \end{tabular}}
        }} &
        \includegraphics[width=0.105\textwidth]{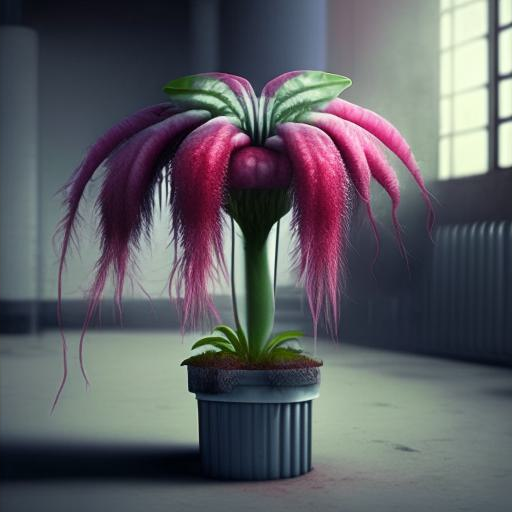} & 
        \includegraphics[width=0.105\textwidth]{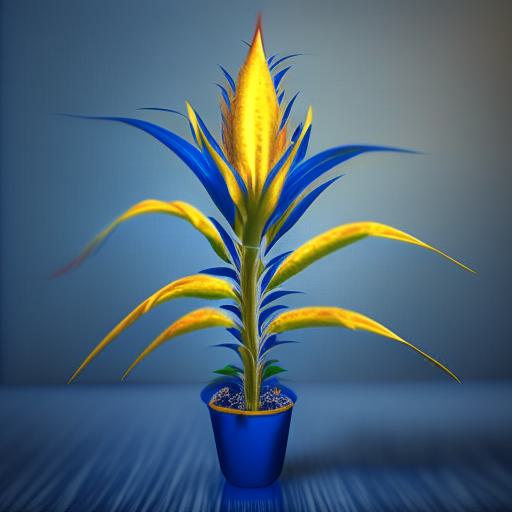} & 
        \includegraphics[width=0.105\textwidth]{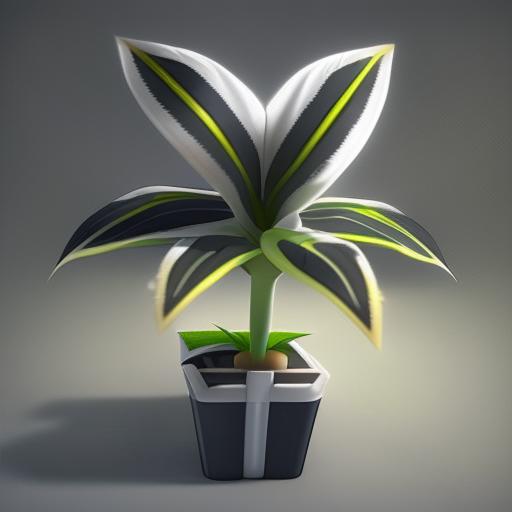} & 
        \includegraphics[width=0.105\textwidth]{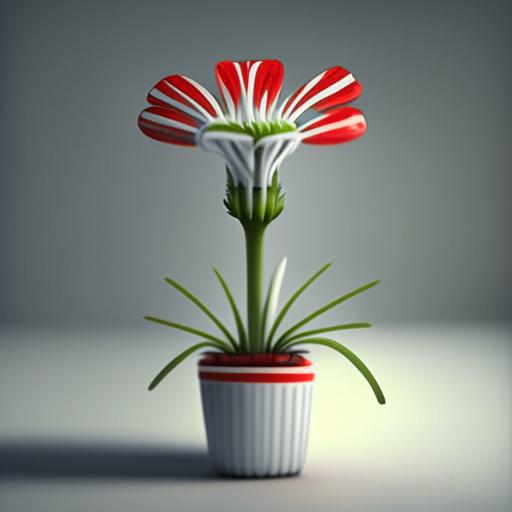} \\

        & 
        \multicolumn{4}{c}{\textcolor{red}{\begin{tabular}{c} \textendash \mbox{} bonsai tree, green leaf plant, cactus, desert flower
        \end{tabular}}} \\

        \raisebox{0.475in}{\multirow{1}{*}{
            \rotatebox{90}{\begin{tabular}{c} \textcolor{mygreen}{\begin{tabular}{c} \\ + fruit \end{tabular}}  \end{tabular}}
        }} &
        \includegraphics[width=0.105\textwidth]{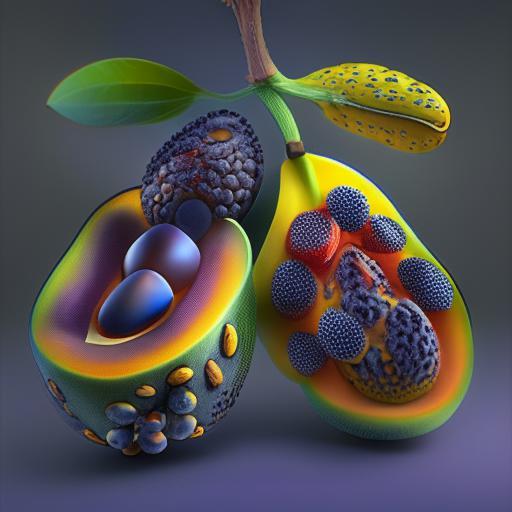} & 
        \includegraphics[width=0.105\textwidth]{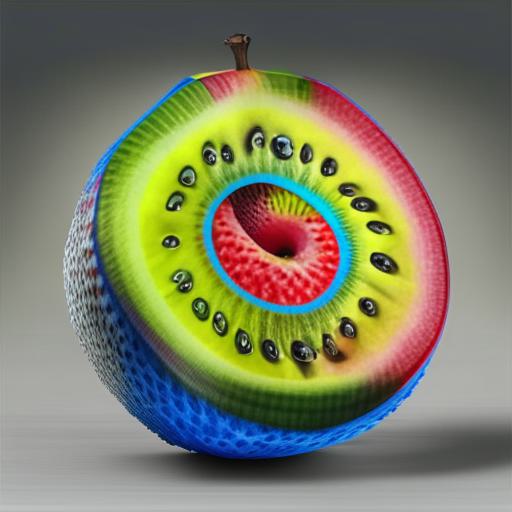} & 
        \includegraphics[width=0.105\textwidth]{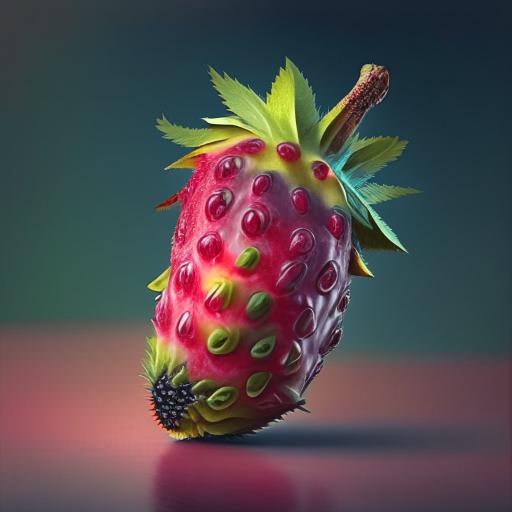} & 
        \includegraphics[width=0.105\textwidth]{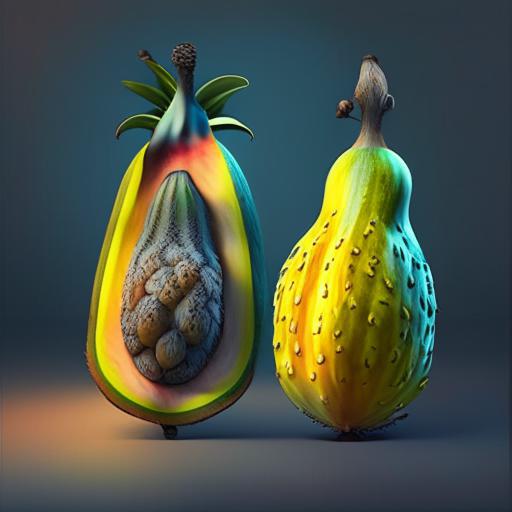} \\

        \raisebox{0.475in}{\multirow{1}{*}{
            \rotatebox{90}{\begin{tabular}{c} \textcolor{mygreen}{\begin{tabular}{c} \\ + building \end{tabular}}  \end{tabular}}
        }} &
        \includegraphics[width=0.105\textwidth]{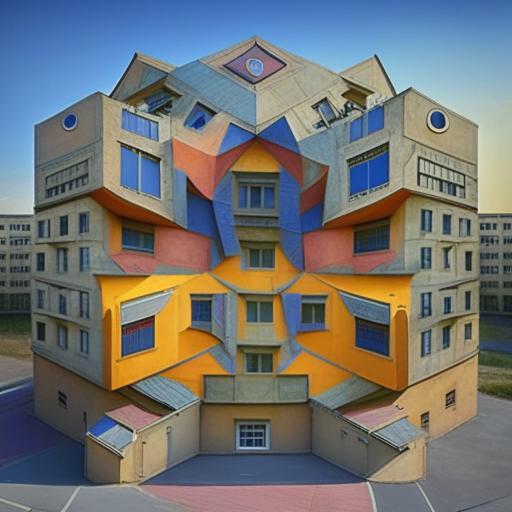} & 
        \includegraphics[width=0.105\textwidth]{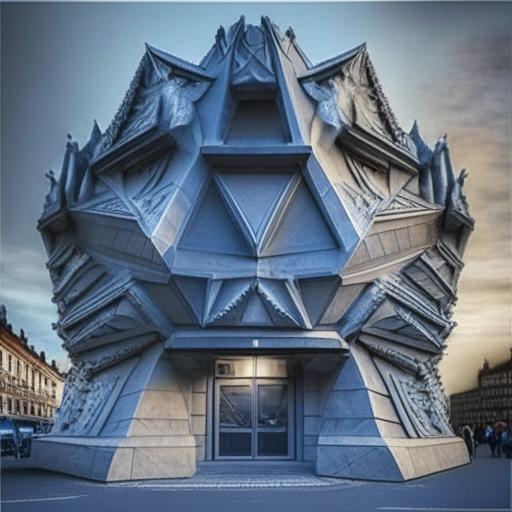} & 
        \includegraphics[width=0.105\textwidth]{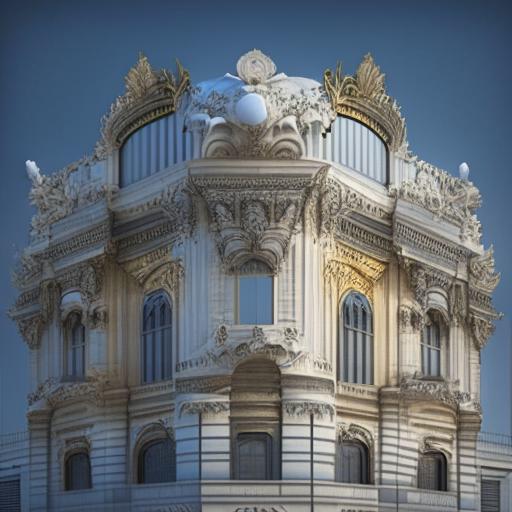} & 
        \includegraphics[width=0.105\textwidth]{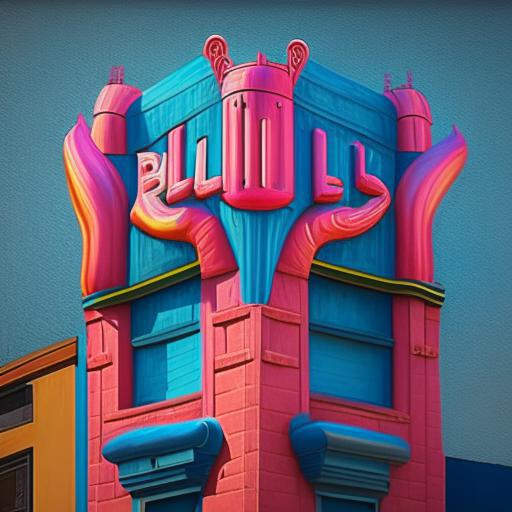} \\

        \raisebox{0.5in}{\multirow{1}{*}{
            \rotatebox{90}{\begin{tabular}{c} \textcolor{mygreen}{\begin{tabular}{c} + large \\ mammal \end{tabular}} \end{tabular}}
        }} &
        \includegraphics[width=0.105\textwidth]{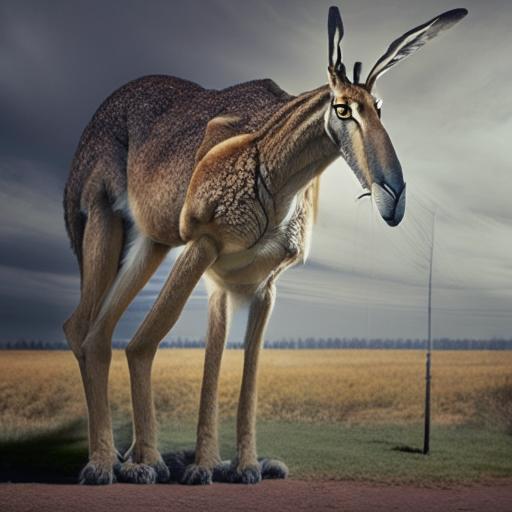} & 
        \includegraphics[width=0.105\textwidth]{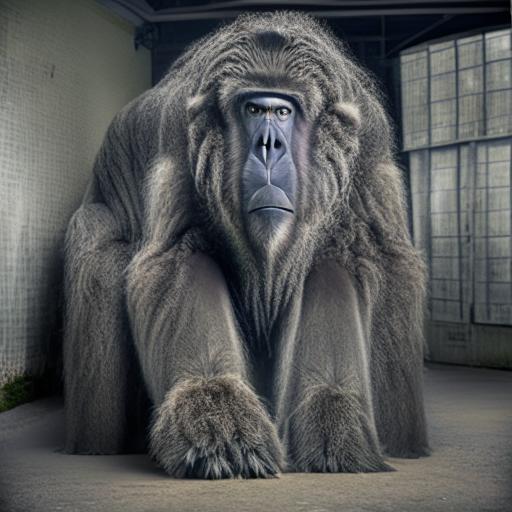} & 
        \includegraphics[width=0.105\textwidth]{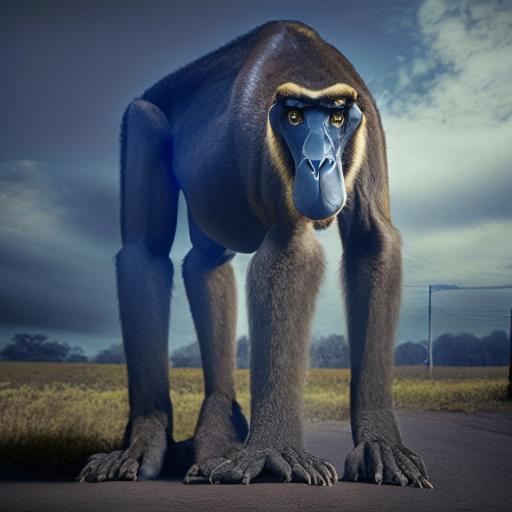} & 
        \includegraphics[width=0.105\textwidth]{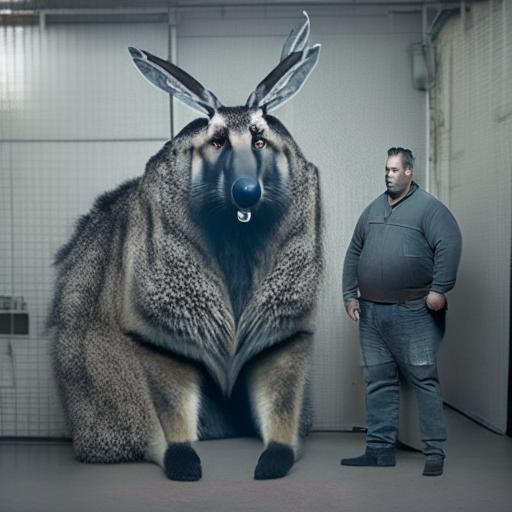} \\

    \end{tabular}
    }
    \caption{
    Creative generation results obtained across various  categories using adaptive negatives with different training seeds. 
    }
    \label{fig:creative_results_original_token}
\end{figure}

\begin{figure*}
    \centering
    \setlength{\tabcolsep}{0.1pt}
    {\footnotesize
    \begin{tabular}{c c@{\hspace{0.25cm}} c@{\hspace{0.25cm}} c@{\hspace{0.25cm}} c@{\hspace{0.25cm}} c@{\hspace{0.25cm}} c@{\hspace{0.25cm}}}

    \raisebox{0.65in}{\multirow{2}{*}{
            \rotatebox{90}{\begin{tabular}{c} \textcolor{mygreen}{+ super hero}  \end{tabular}}
        }} &
        \includegraphics[width=0.15\textwidth]{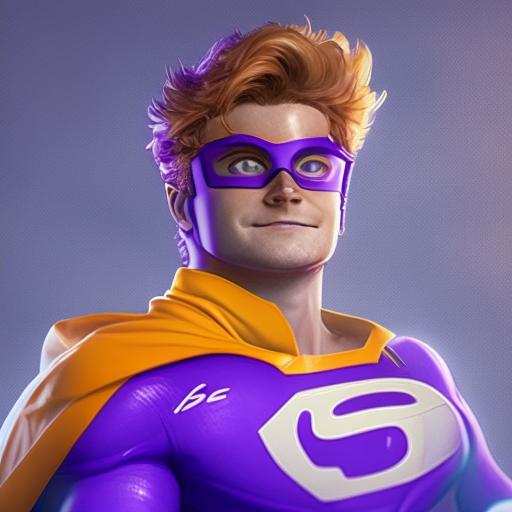}
        \includegraphics[width=0.15\textwidth]{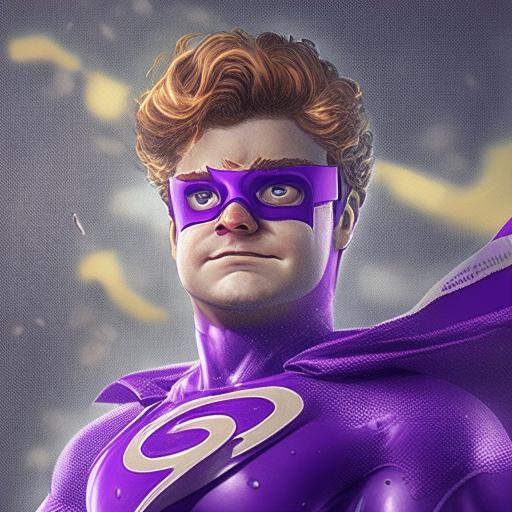} &
        \includegraphics[width=0.15\textwidth]{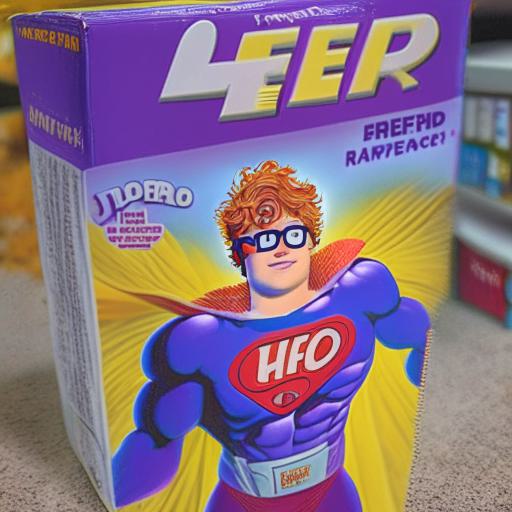} &
        \includegraphics[width=0.15\textwidth]{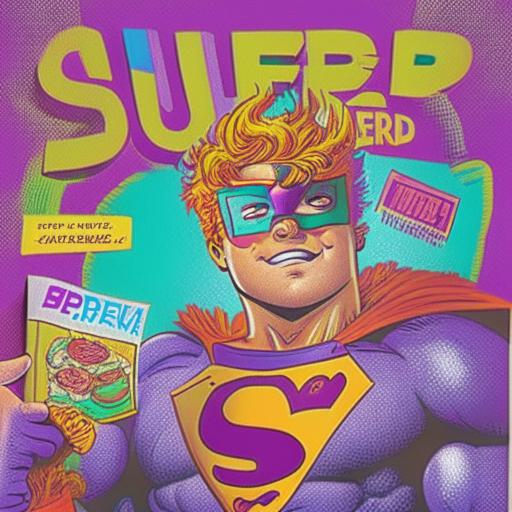} &
        \includegraphics[width=0.15\textwidth]{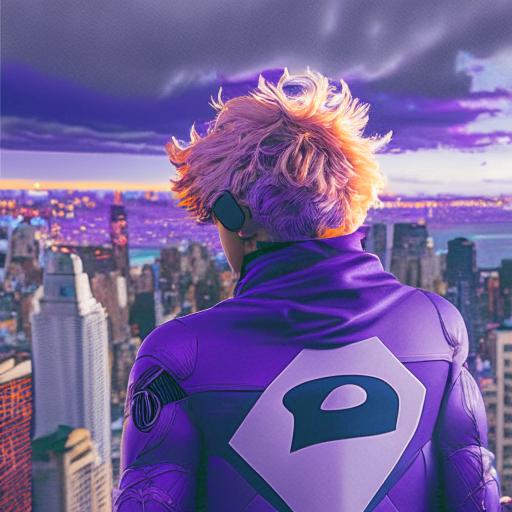} &
        \includegraphics[width=0.15\textwidth]{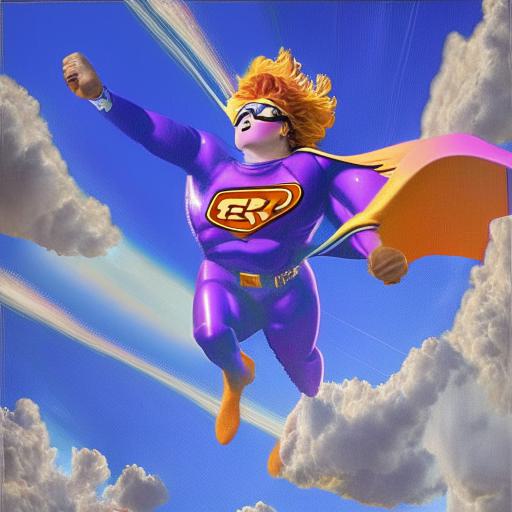} \\

        &
        \begin{tabular}{c} ``Professional high-quality photo of \\ a $S_*$. photorealistic, 4k, HQ'' \end{tabular} &
        \begin{tabular}{c} ``$S_*$ on a cereal box'' \end{tabular} &
        \begin{tabular}{c} ``$S_*$ in a comic book'' \end{tabular} &
        \begin{tabular}{c} ``A photo of $S_*$ \\  overlooking the city'' \end{tabular} &
        \begin{tabular}{c} ``A photo of $S_*$ \\ flying in the sky'' \end{tabular} \\

        \raisebox{0.45in}{\multirow{2}{*}{
            \rotatebox{90}{\begin{tabular}{c} \textcolor{mygreen}{+ pet}  \end{tabular}}
        }} &
        \includegraphics[width=0.15\textwidth]{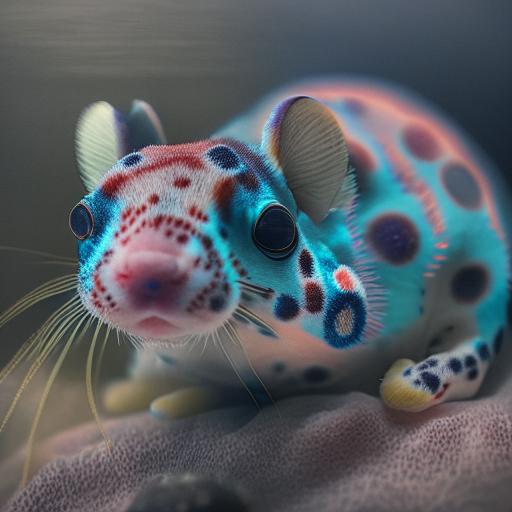}
        \includegraphics[width=0.15\textwidth]{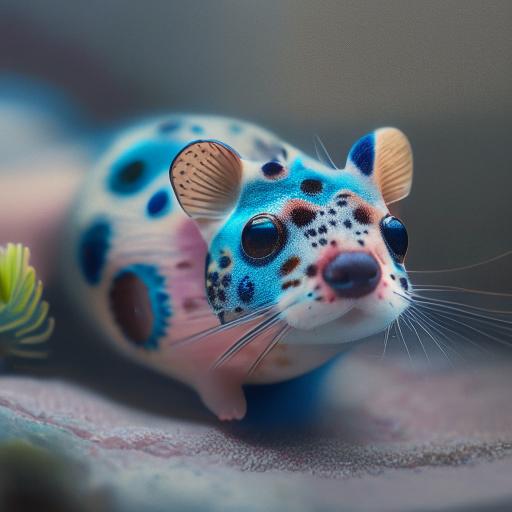} &
        \includegraphics[width=0.15\textwidth]{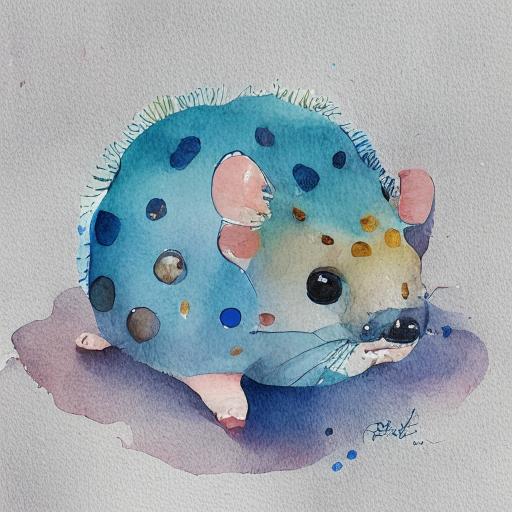} &
        \includegraphics[width=0.15\textwidth]{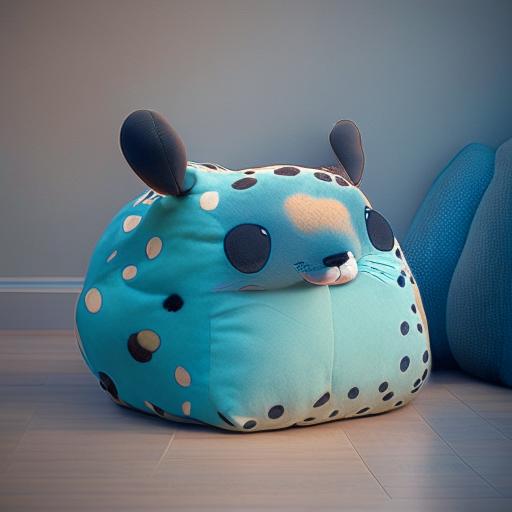} &
        \includegraphics[width=0.15\textwidth]{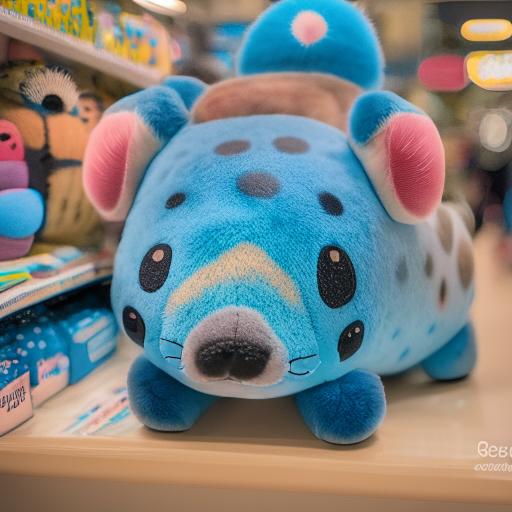} &
        \includegraphics[width=0.15\textwidth]{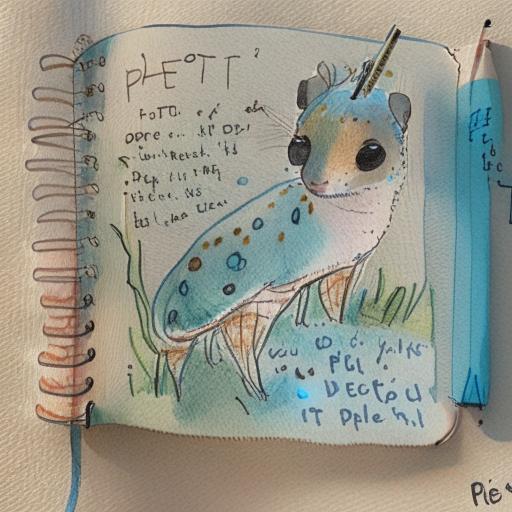} \\

        &
        \begin{tabular}{c} ``Professional high-quality photo of \\ a $S_*$. photorealistic, 4k, HQ'' \end{tabular} &
        \begin{tabular}{c} ``A watercolor \\ painting of $S_*$'' \end{tabular} &
        \begin{tabular}{c} ``$S_*$ as a bean \\ bag chair'' \end{tabular} &
        \begin{tabular}{c} ``A plush toy $S_*$ \\ in a toy store'' \end{tabular} &
        \begin{tabular}{c} ``An illustrated notebook \\  page about  $S_*$'' \end{tabular} \\

        \raisebox{0.65in}{\multirow{2}{*}{
            \rotatebox{90}{\begin{tabular}{c} \textcolor{mygreen}{+ building}  \end{tabular}}
        }} &
        \includegraphics[width=0.15\textwidth]{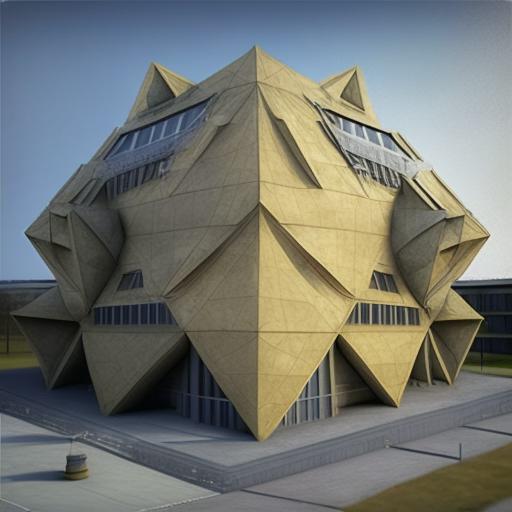}
        \includegraphics[width=0.15\textwidth]{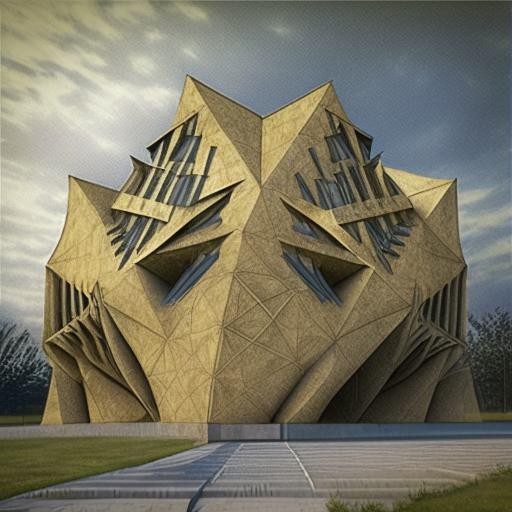} &
        \includegraphics[width=0.15\textwidth]{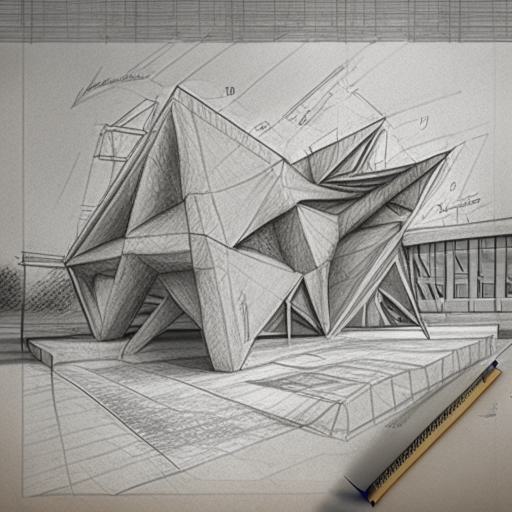} &
        \includegraphics[width=0.15\textwidth]{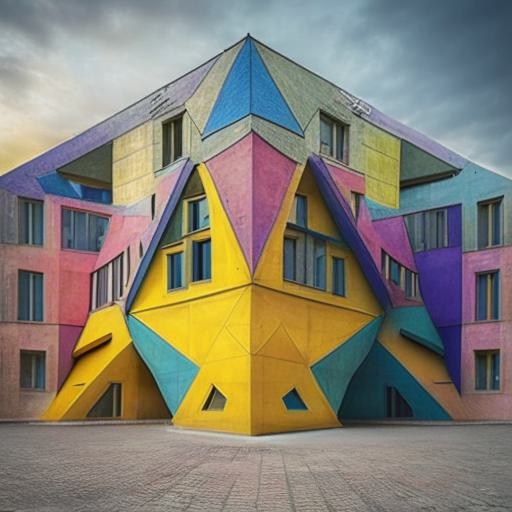} &
        \includegraphics[width=0.15\textwidth]{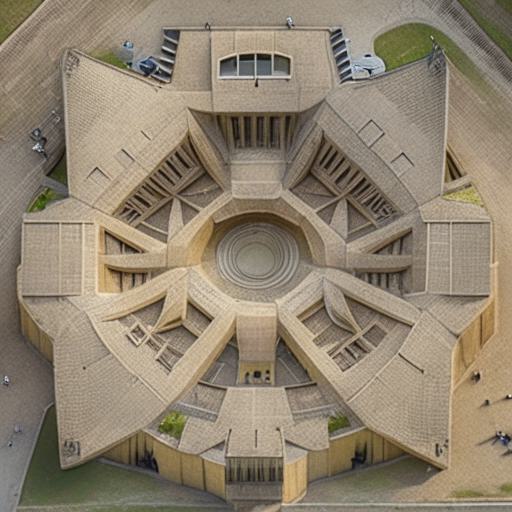} &
        \includegraphics[width=0.15\textwidth]{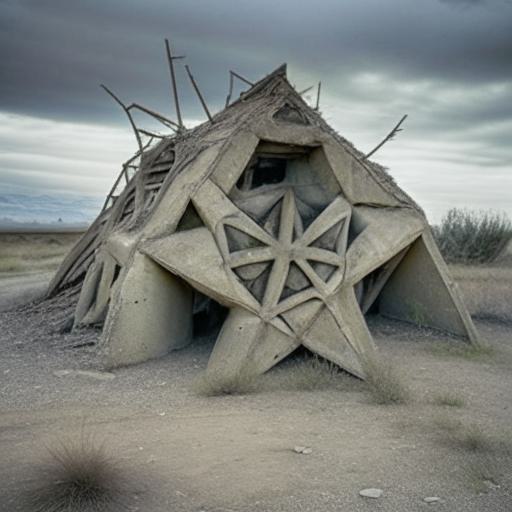} \\

        &
        \begin{tabular}{c} ``Professional high-quality photo of \\ a $S_*$. photorealistic, 4k, HQ'' \end{tabular} &
        \begin{tabular}{c} ``A high-detailed pencil \\ sketch of $S_*$'' \end{tabular} &
        \begin{tabular}{c} ``A colorful building with \\ the architecture of $S_*$'' \end{tabular} &
        \begin{tabular}{c} ``A bird's eye \\ view of $S_*$'' \end{tabular} &
        \begin{tabular}{c} ``An abandoned $S_*$ \\ in a ghost town \end{tabular} \\ 

        \raisebox{0.55in}{\multirow{2}{*}{
            \rotatebox{90}{\begin{tabular}{c} \textcolor{mygreen}{+ reptile}  \end{tabular}}
        }} &
        \includegraphics[width=0.15\textwidth]{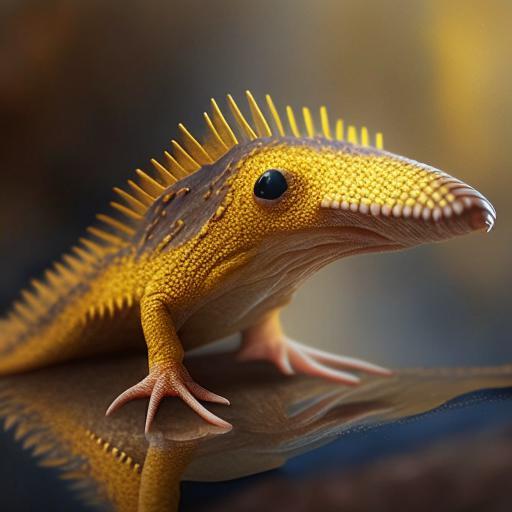}
        \includegraphics[width=0.15\textwidth]{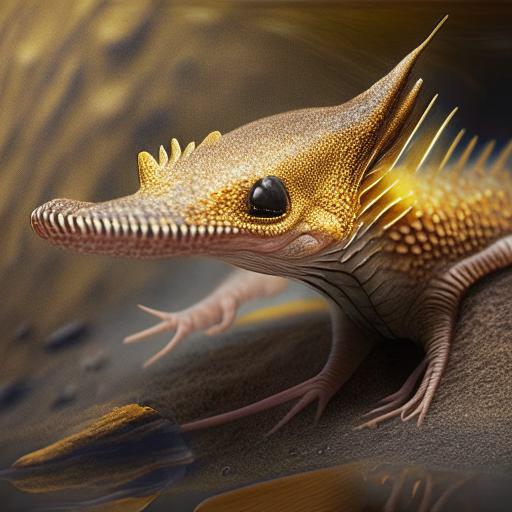} &
        \includegraphics[width=0.15\textwidth]{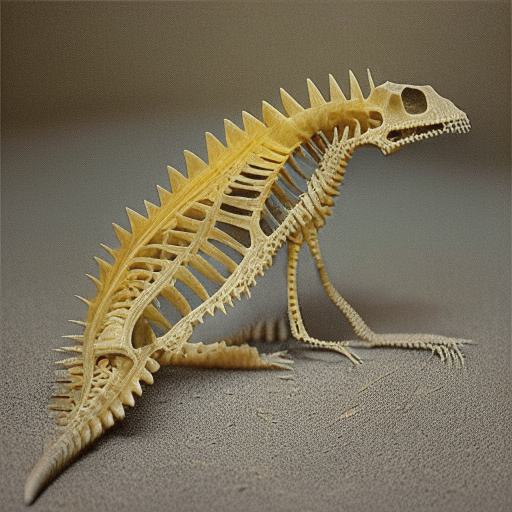} &
        \includegraphics[width=0.15\textwidth]{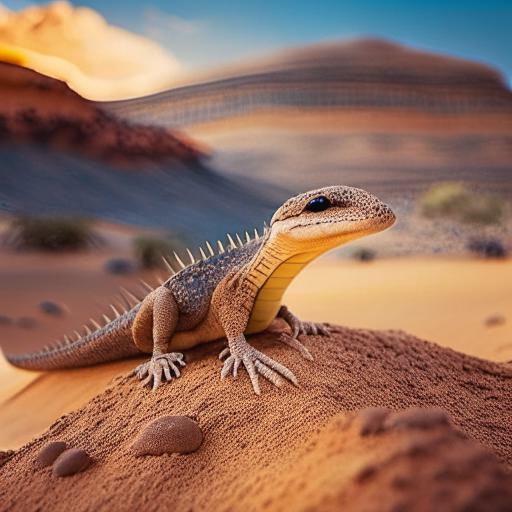} &
        \includegraphics[width=0.15\textwidth]{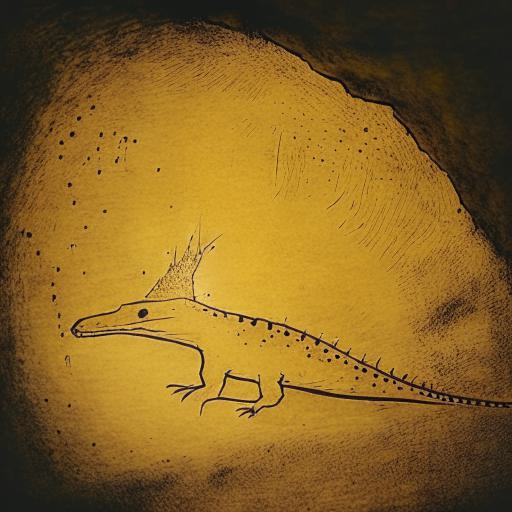} &
        \includegraphics[width=0.15\textwidth]{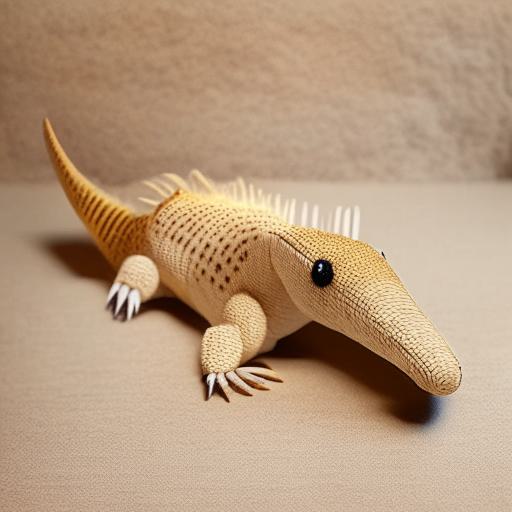} \\ 
        
        &
        \begin{tabular}{c} ``Professional high-quality photo of \\ a $S_*$. photorealistic, 4k, HQ'' \end{tabular} &
        \begin{tabular}{c} ``A skeleton of a $S_*$'' \end{tabular} &
        \begin{tabular}{c} ``A photo of a $S_*$ on \\ the rock in the desert'' \end{tabular} &
        \begin{tabular}{c} ``A cave drawing of \\ a $S_*$'' \end{tabular} &
        \begin{tabular}{c} ``A plush of a $S_*$'' \end{tabular} \\

        \raisebox{0.5in}{\multirow{2}{*}{
            \rotatebox{90}{\begin{tabular}{c} \textcolor{mygreen}{+ fruit}  \end{tabular}}
        }} &
        \includegraphics[width=0.15\textwidth]{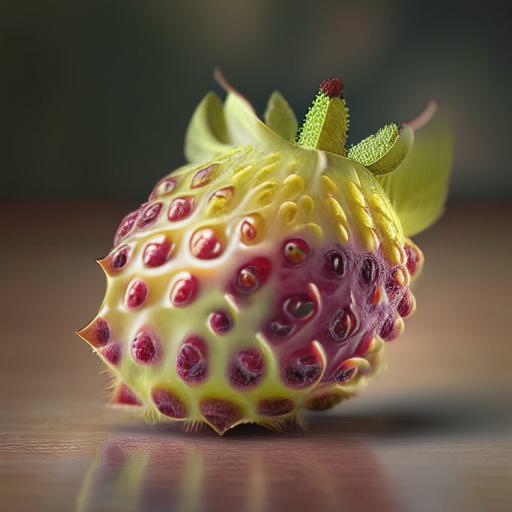}
        \includegraphics[width=0.15\textwidth]{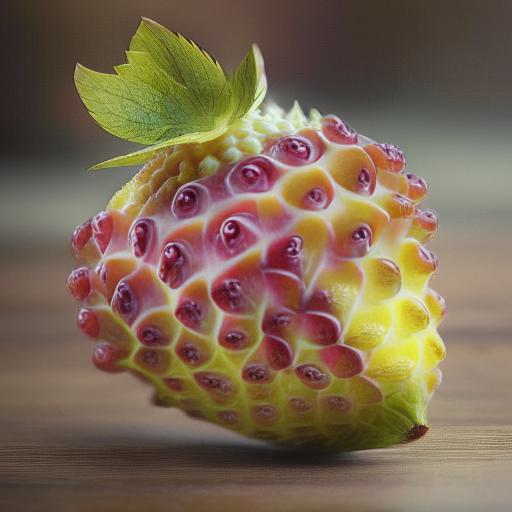} &
        \includegraphics[width=0.15\textwidth]{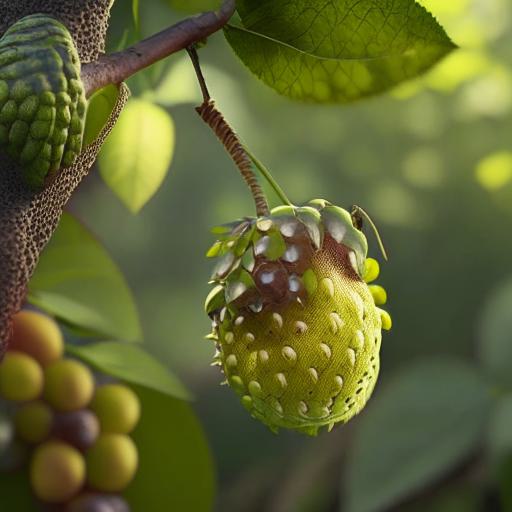} &
        \includegraphics[width=0.15\textwidth]{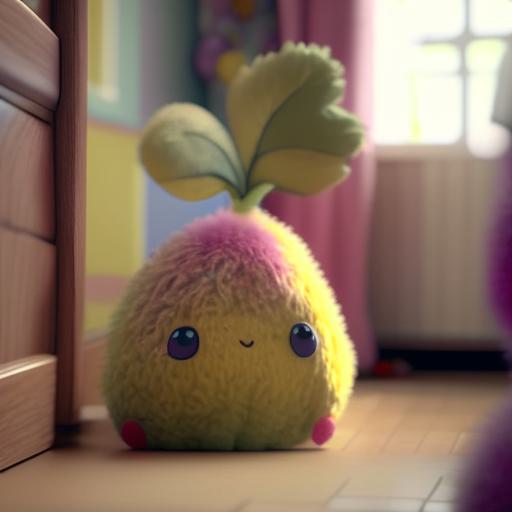} &
        \includegraphics[width=0.15\textwidth]{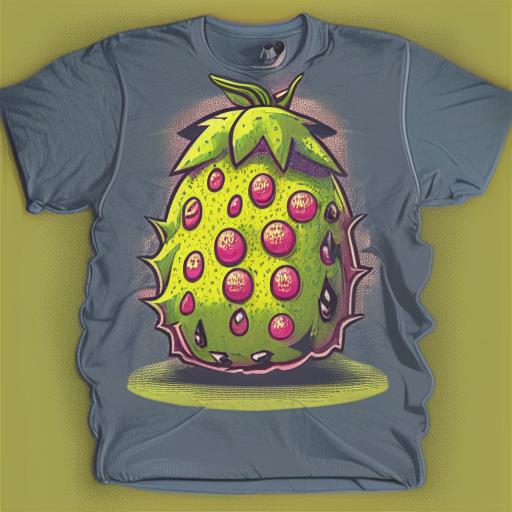} &
        \includegraphics[width=0.15\textwidth]{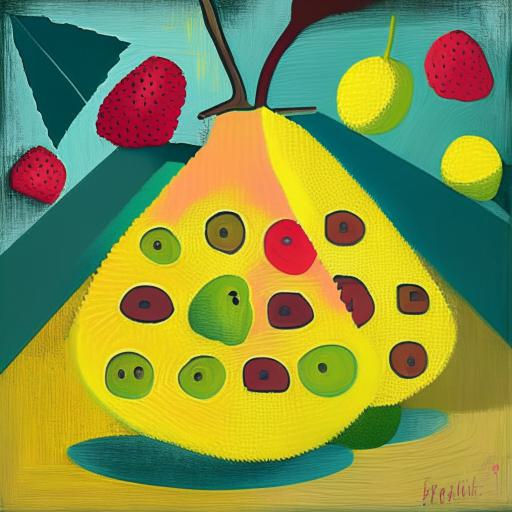} \\

        &
        \begin{tabular}{c} ``Professional high-quality photo of \\ a $S_*$. photorealistic, 4k, HQ'' \end{tabular} &
        \begin{tabular}{c} ``A photo of a $S_*$ \\ hanging on a tree'' \end{tabular} &
        \begin{tabular}{c} ``A plush $S_*$ \\ in a playroom'' \end{tabular} &
        \begin{tabular}{c} ``A shirt featuring \\ a $S_*$ design'' \end{tabular} &
        \begin{tabular}{c} ``An abstract Picasso \\ painting of $S_*$'' \end{tabular} \\

        \raisebox{0.55in}{\multirow{2}{*}{
            \rotatebox{90}{\begin{tabular}{c} \textcolor{mygreen}{+ mammal}  \end{tabular}}
        }} &
        \includegraphics[width=0.15\textwidth]{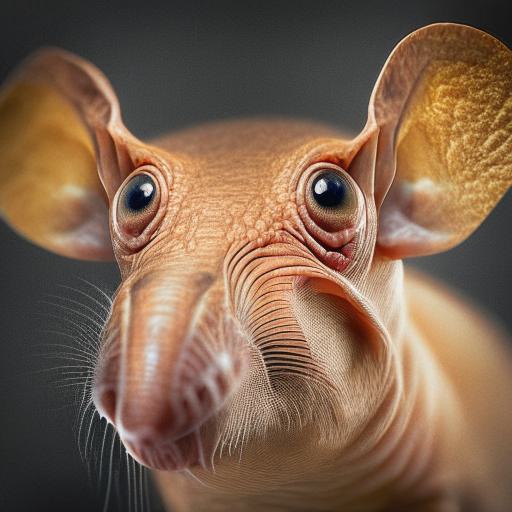}
        \includegraphics[width=0.15\textwidth]{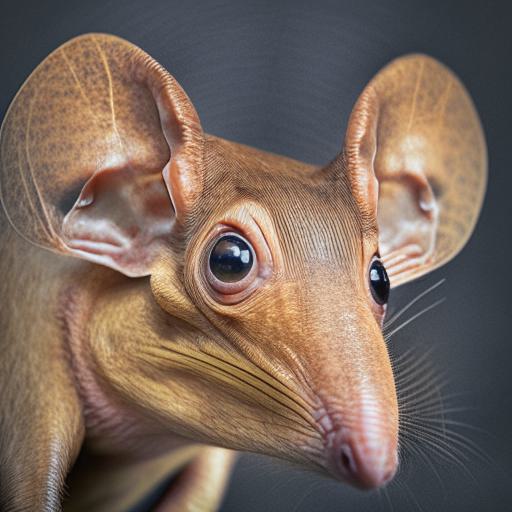} &
        \includegraphics[width=0.15\textwidth]{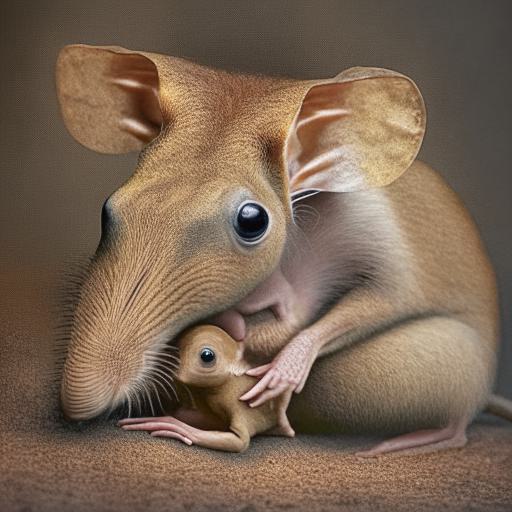} &
        \includegraphics[width=0.15\textwidth]{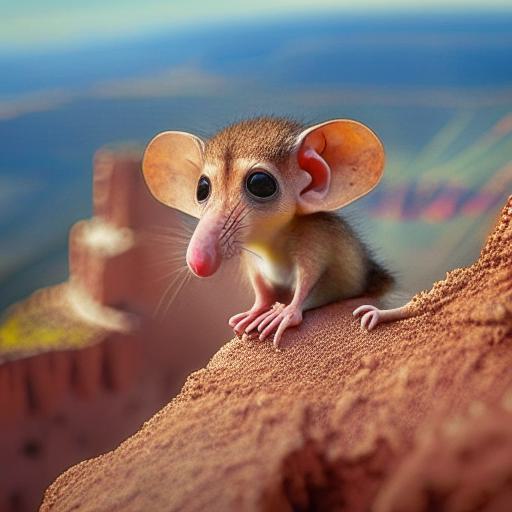} &
        \includegraphics[width=0.15\textwidth]{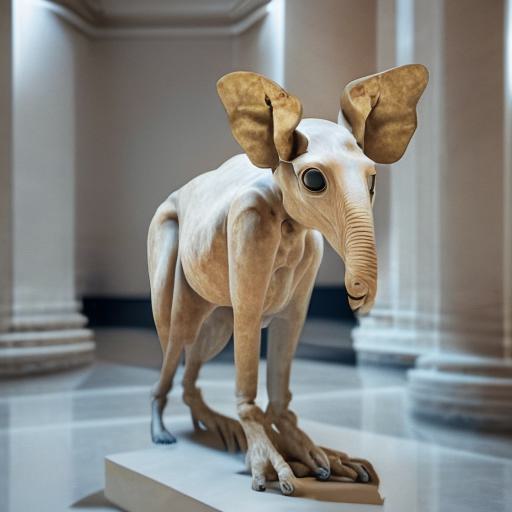} &
        \includegraphics[width=0.15\textwidth]{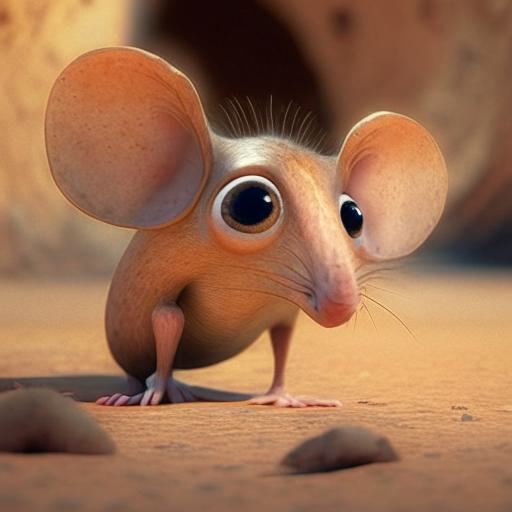} \\

        &
        \begin{tabular}{c} ``Professional high-quality photo of \\ a $S_*$. photorealistic, 4k, HQ'' \end{tabular} &
        \begin{tabular}{c} ``A mother $S_*$ \\ with her baby $S_*$'' \end{tabular} &
        \begin{tabular}{c} ``A baby  $S_*$ on a cliff, \\ overlooking a \\ national park'' \end{tabular} &
        \begin{tabular}{c} ``A marble statue of \\ $S_*$ in a museum'' \end{tabular} &
        \begin{tabular}{c} ``A $S_*$ appearing in \\ a pixar movie'' \end{tabular} \\

            \end{tabular}

    }
    \caption{Sample text-guided creative generation results obtained with ConceptLab. The positive concept used for training is shown to the left. All results are obtained using our adaptive negative technique.}
    \label{fig:our_results}
\end{figure*}
\begin{figure*}
    \centering
    \setlength{\tabcolsep}{1pt}
    \renewcommand{\arraystretch}{0.5}
    {\small
    \begin{tabular}{c c c c c c}
        \includegraphics[width=0.14\textwidth]{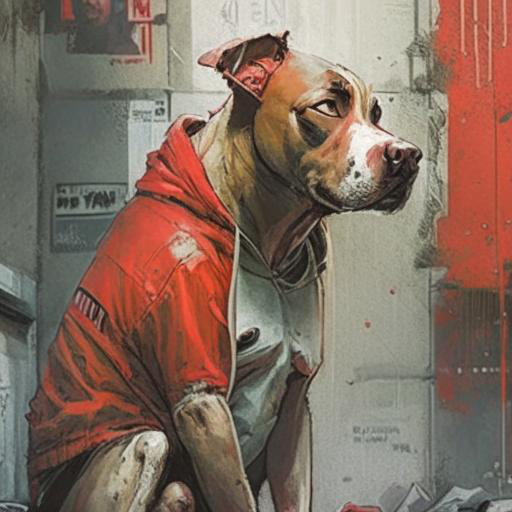} &
        \includegraphics[width=0.14\textwidth]{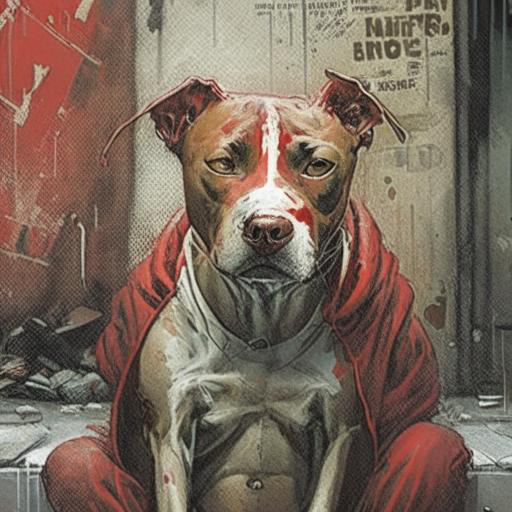} &
        \includegraphics[width=0.14\textwidth]{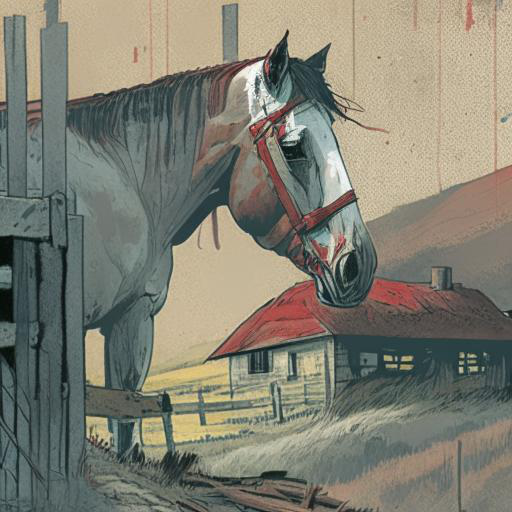} &
        \includegraphics[width=0.14\textwidth]{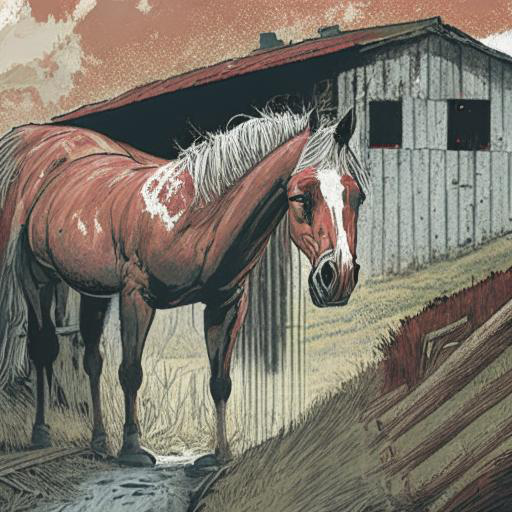} &
        \includegraphics[width=0.14\textwidth]{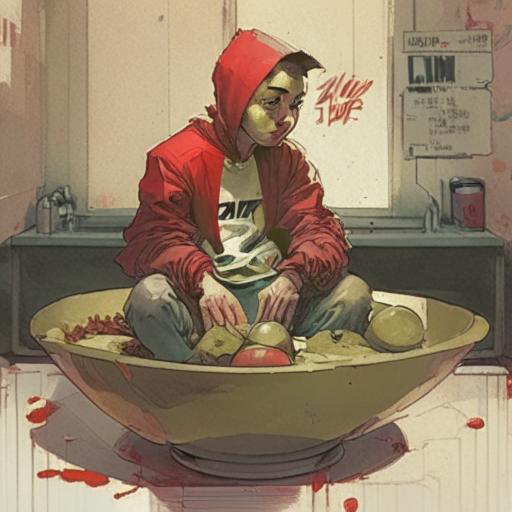} &
        \includegraphics[width=0.14\textwidth]{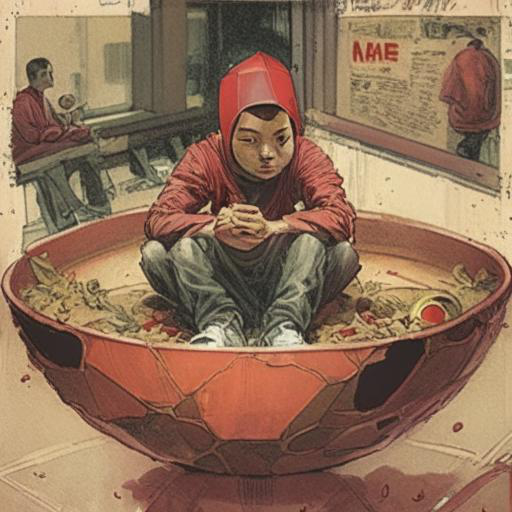} 
        \\ 
        \includegraphics[width=0.14\textwidth]{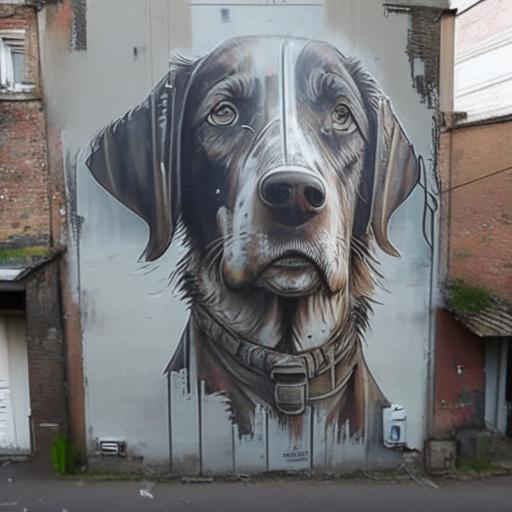} &
        \includegraphics[width=0.14\textwidth]{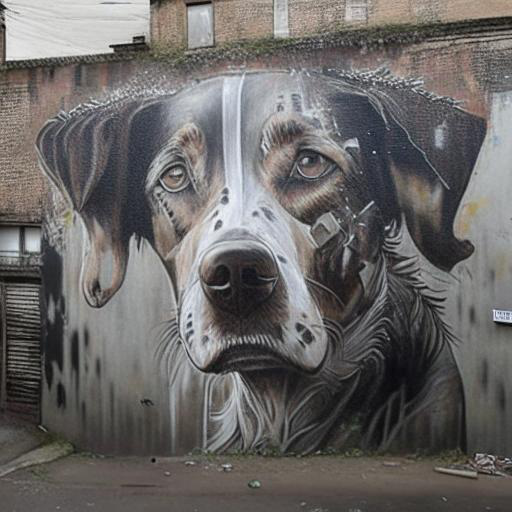} &
        \includegraphics[width=0.14\textwidth]{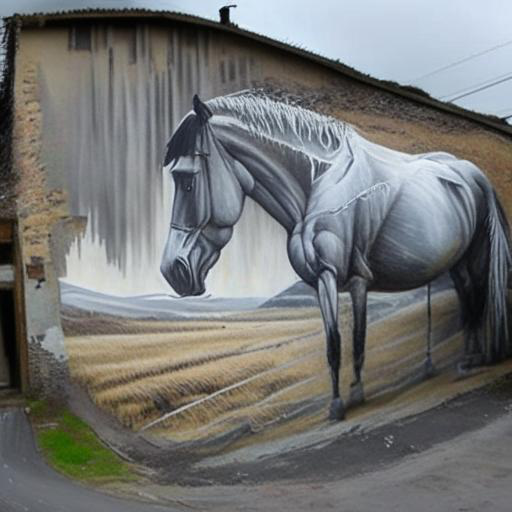} &
        \includegraphics[width=0.14\textwidth]{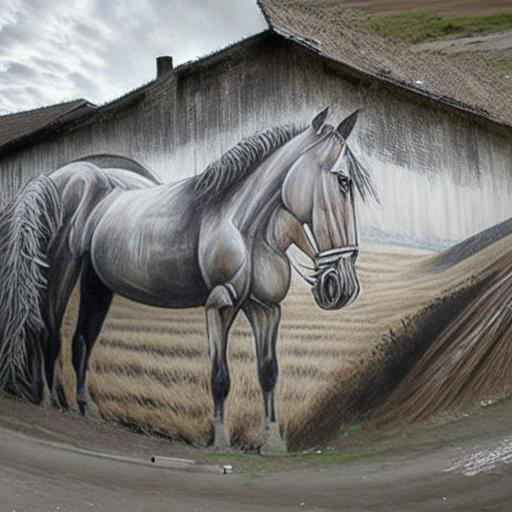} &
        \includegraphics[width=0.14\textwidth]{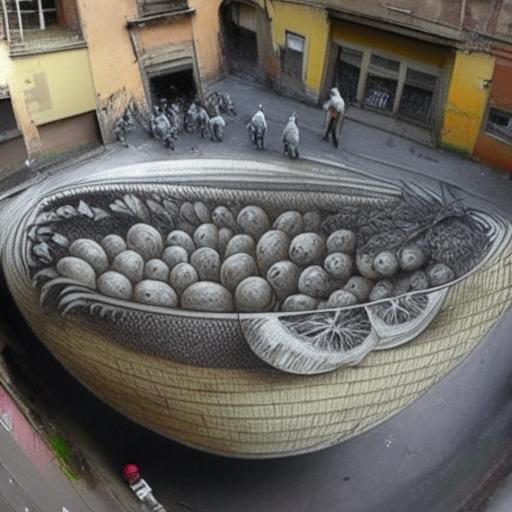} &
        \includegraphics[width=0.14\textwidth]{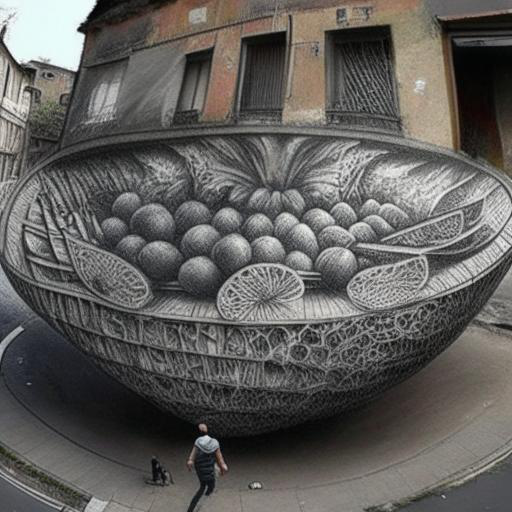} 
        \\ 

        \includegraphics[width=0.14\textwidth]{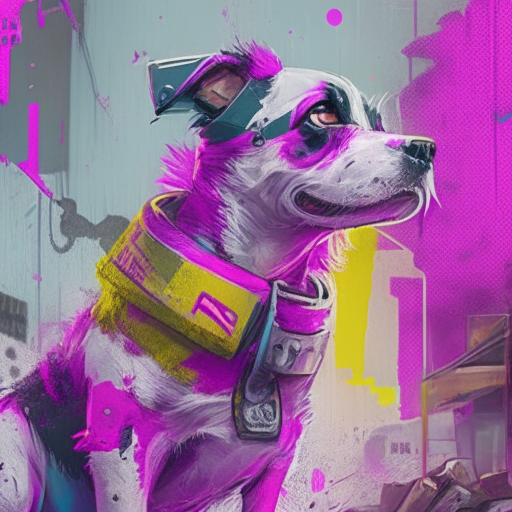} &
        \includegraphics[width=0.14\textwidth]{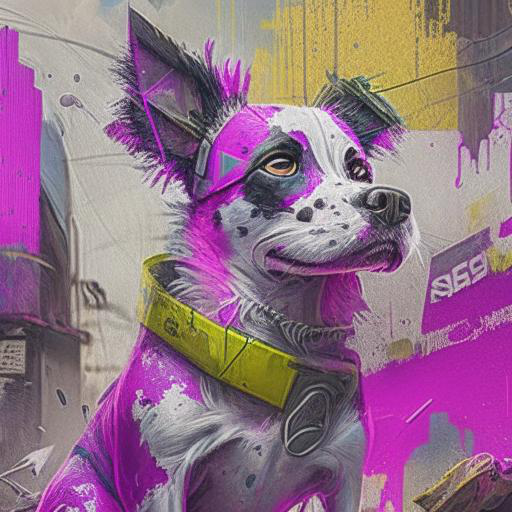} &
        \includegraphics[width=0.14\textwidth]{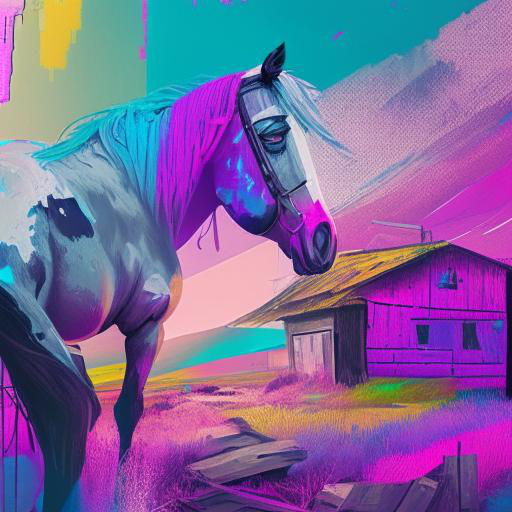} &
        \includegraphics[width=0.14\textwidth]{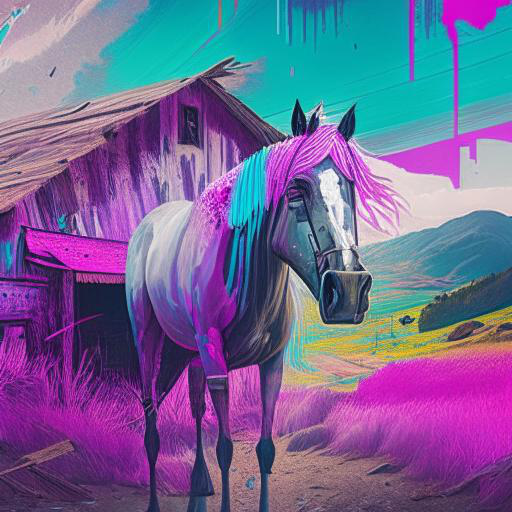} &
        \includegraphics[width=0.14\textwidth]{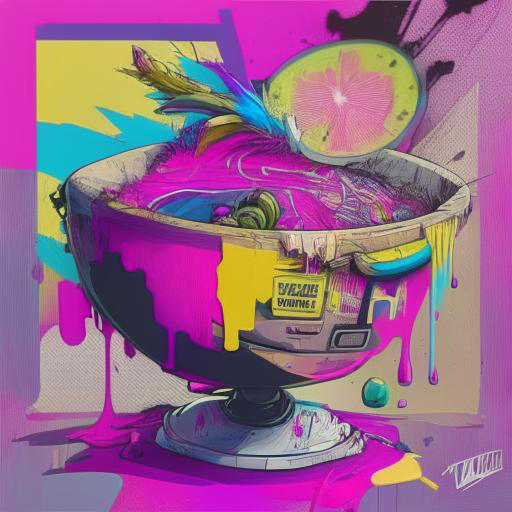} &
        \includegraphics[width=0.14\textwidth]{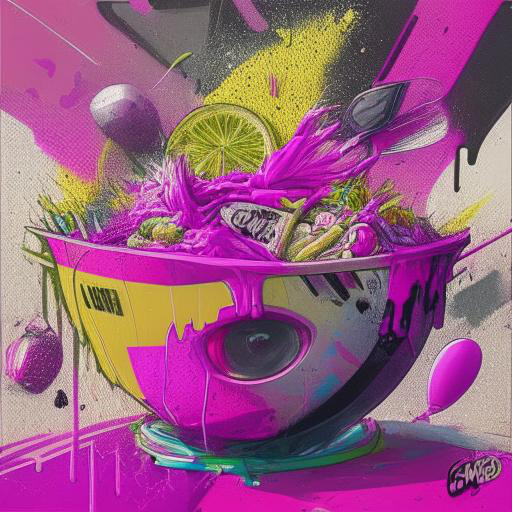} 
        \\ 
        \includegraphics[width=0.14\textwidth]{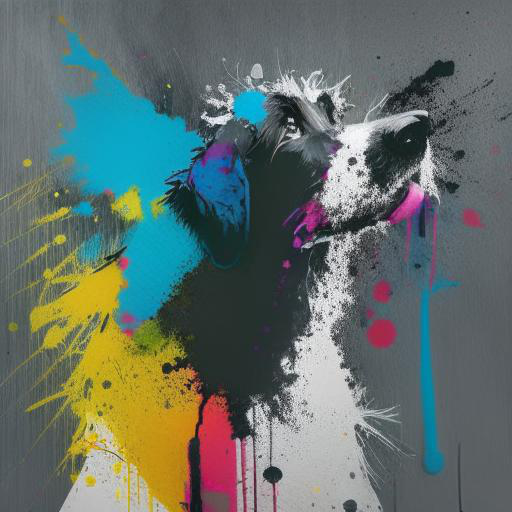} &
        \includegraphics[width=0.14\textwidth]{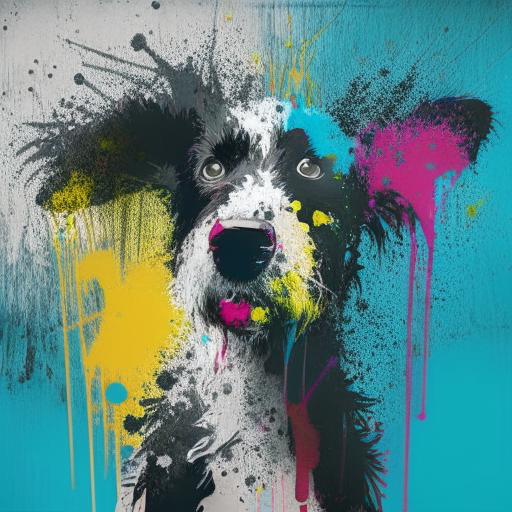} &
        \includegraphics[width=0.14\textwidth]{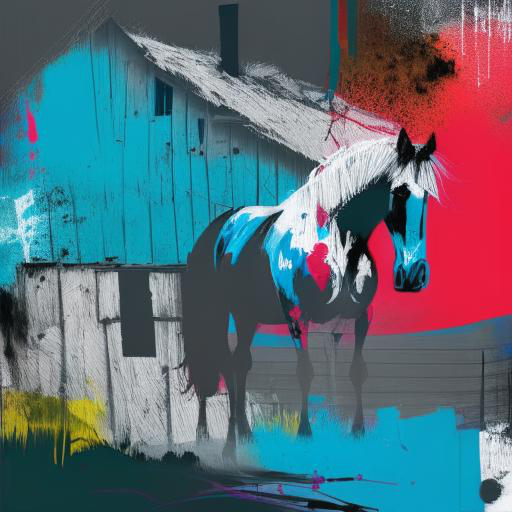} &
        \includegraphics[width=0.14\textwidth]{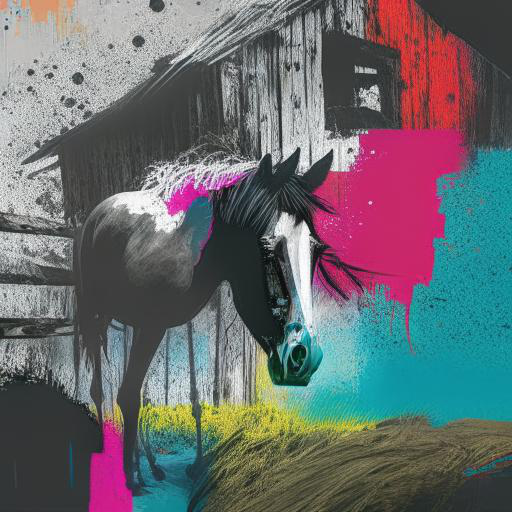} &
        \includegraphics[width=0.14\textwidth]{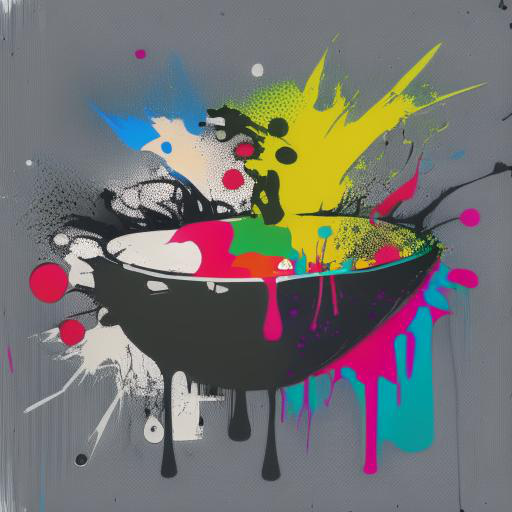} &
        \includegraphics[width=0.14\textwidth]{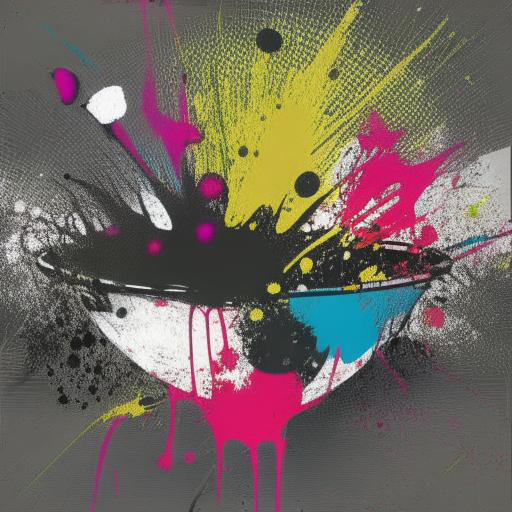} 
        \\ 
        \includegraphics[width=0.14\textwidth]{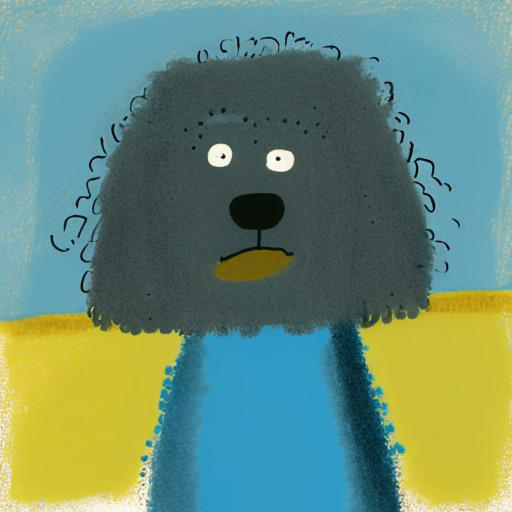} &
        \includegraphics[width=0.14\textwidth]{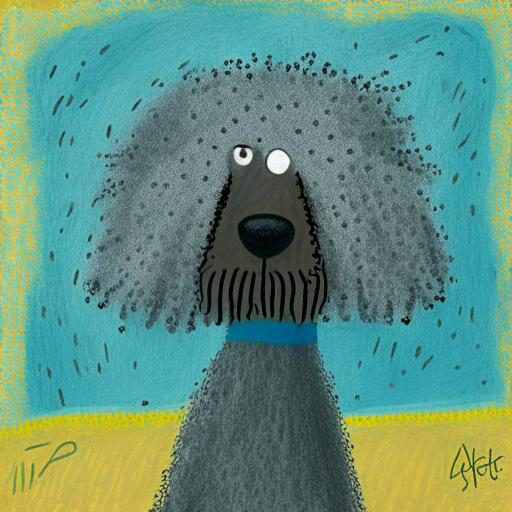} &
        \includegraphics[width=0.14\textwidth]{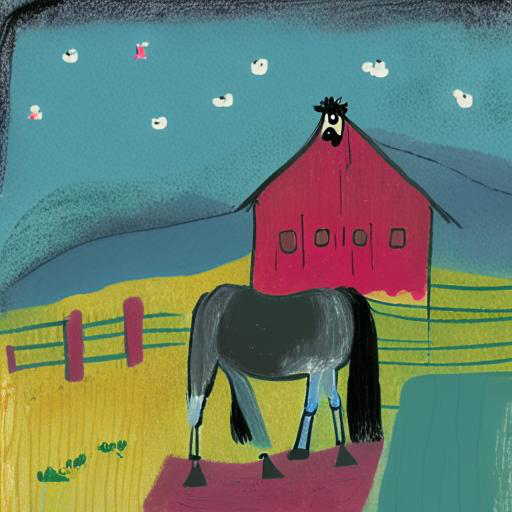} &
        \includegraphics[width=0.14\textwidth]{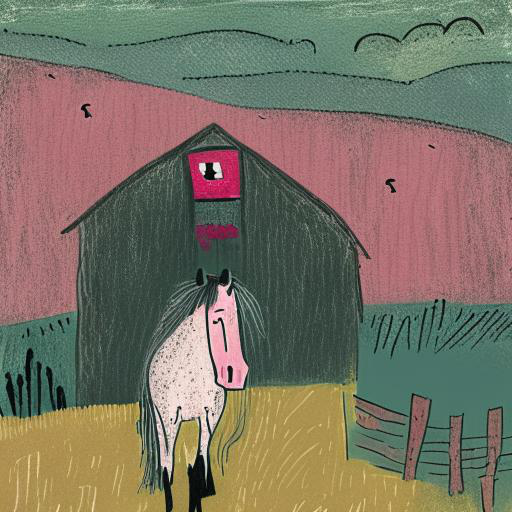} &
        \includegraphics[width=0.14\textwidth]{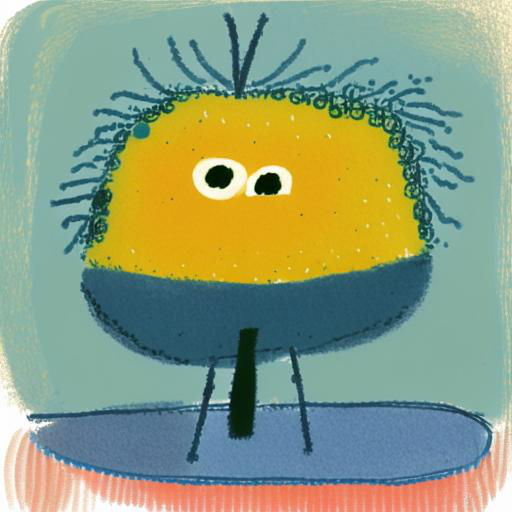} &
        \includegraphics[width=0.14\textwidth]{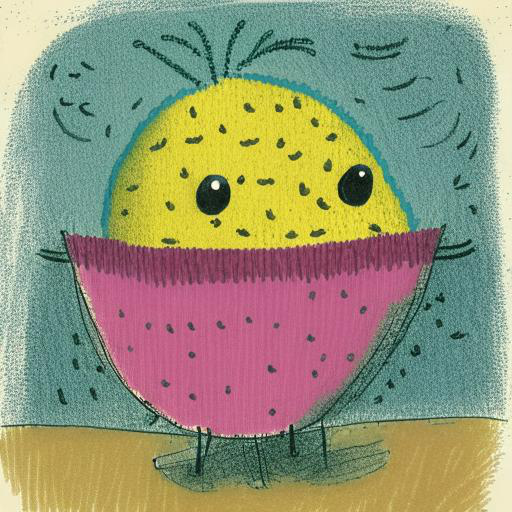} 
        \\ 
        \includegraphics[width=0.14\textwidth]{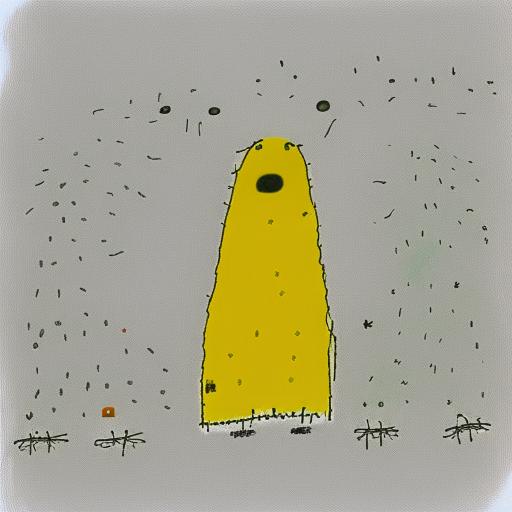} &
        \includegraphics[width=0.14\textwidth]{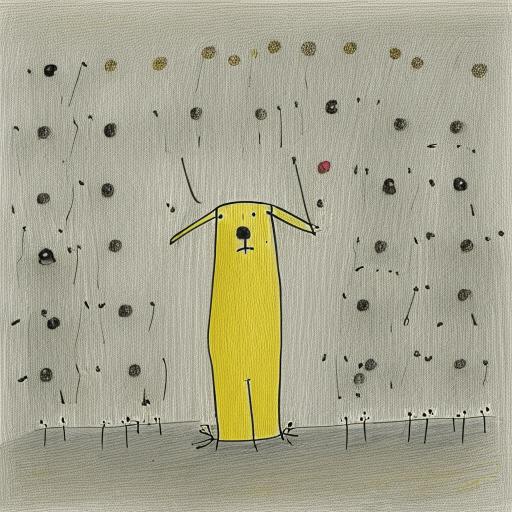} &
        \includegraphics[width=0.14\textwidth]{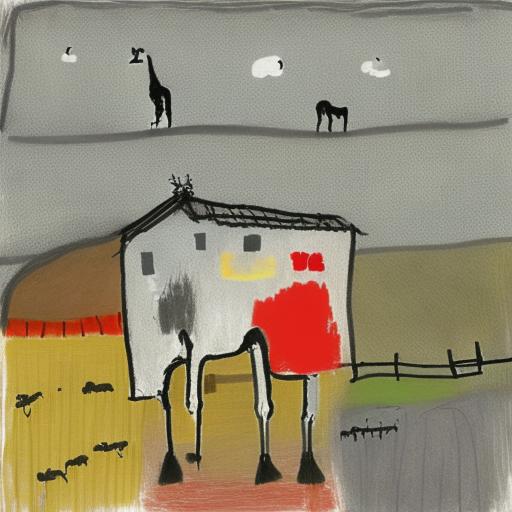} &
        \includegraphics[width=0.14\textwidth]{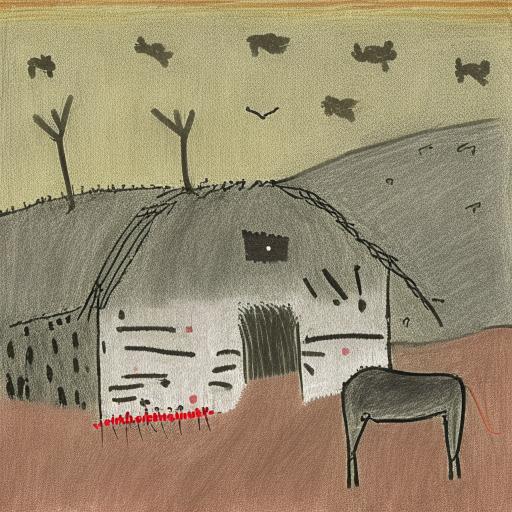} &
        \includegraphics[width=0.14\textwidth]{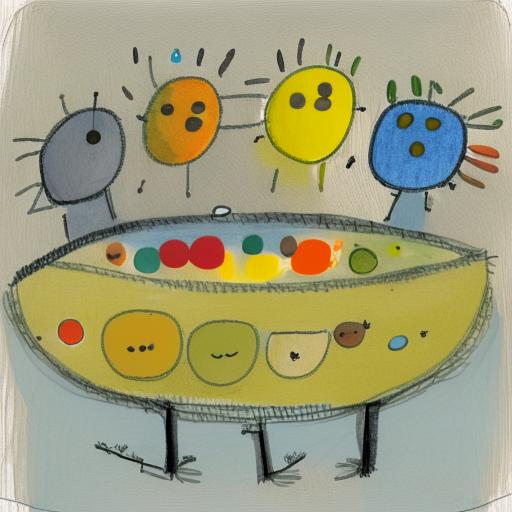} &
        \includegraphics[width=0.14\textwidth]{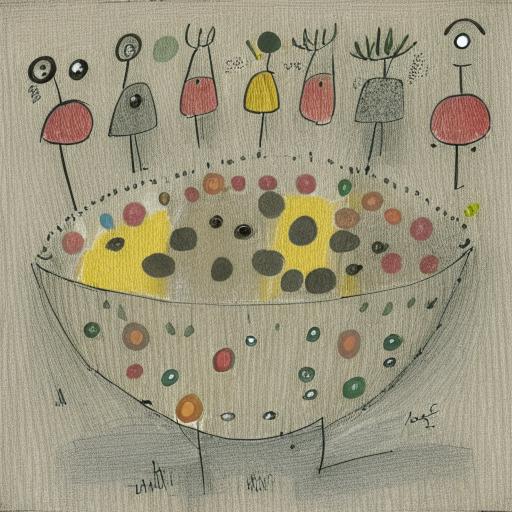} 
        \\ 
        \includegraphics[width=0.14\textwidth]{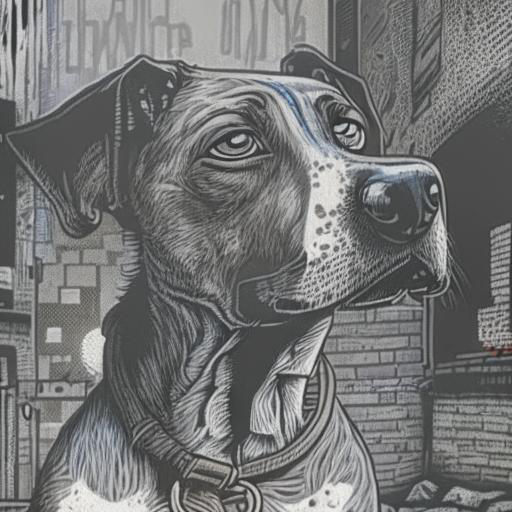} &
        \includegraphics[width=0.14\textwidth]{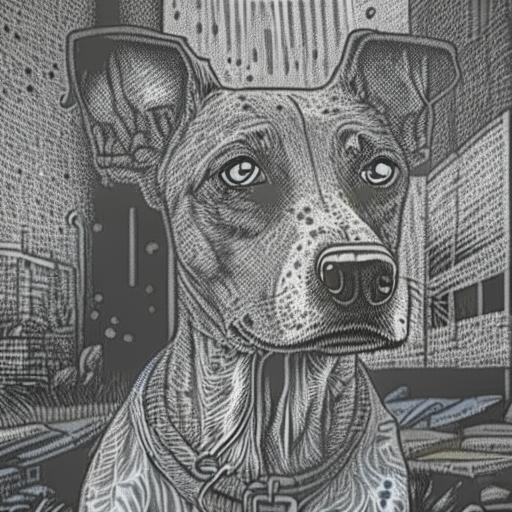} &
        \includegraphics[width=0.14\textwidth]{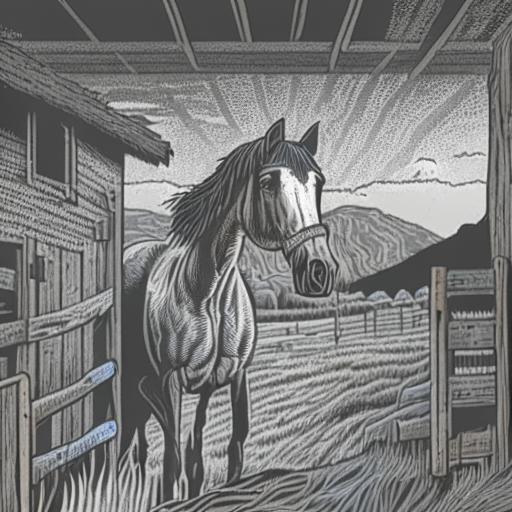} &
        \includegraphics[width=0.14\textwidth]{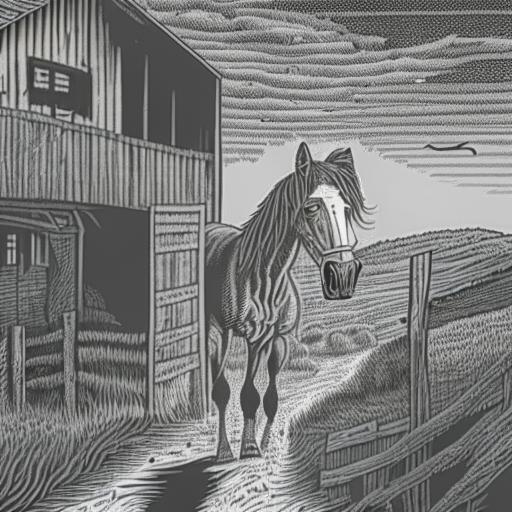} &
        \includegraphics[width=0.14\textwidth]{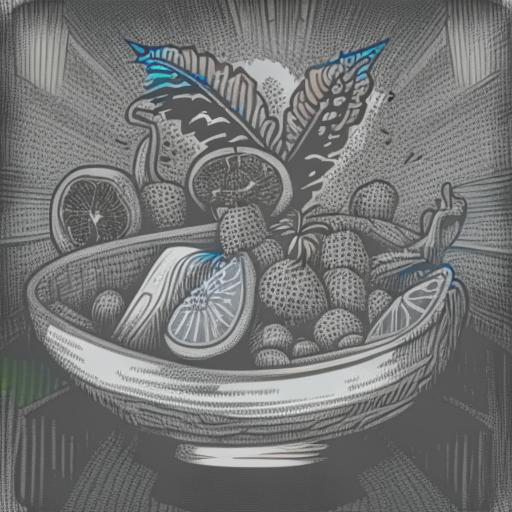} &
        \includegraphics[width=0.14\textwidth]{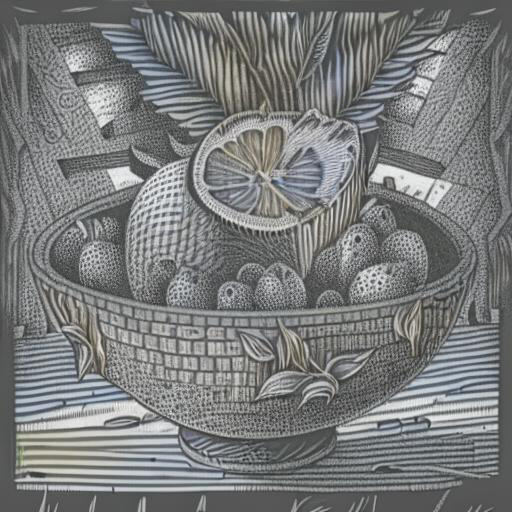} 
        \\ 
        \includegraphics[width=0.14\textwidth]{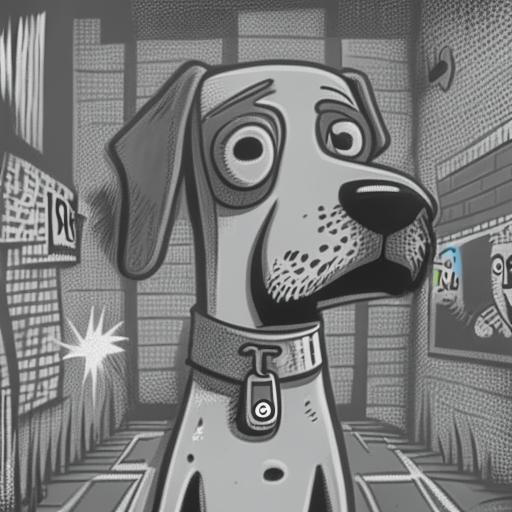} &
        \includegraphics[width=0.14\textwidth]{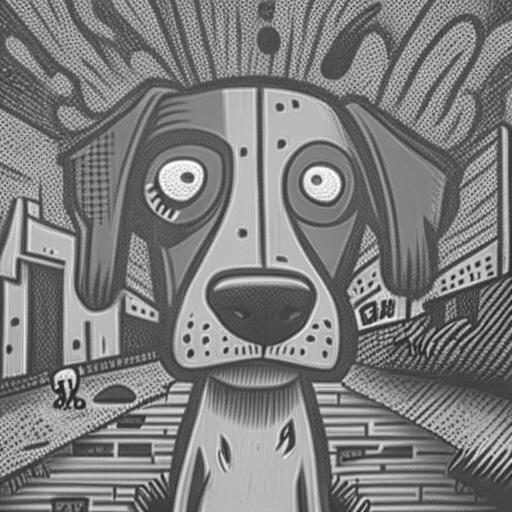} &
        \includegraphics[width=0.14\textwidth]{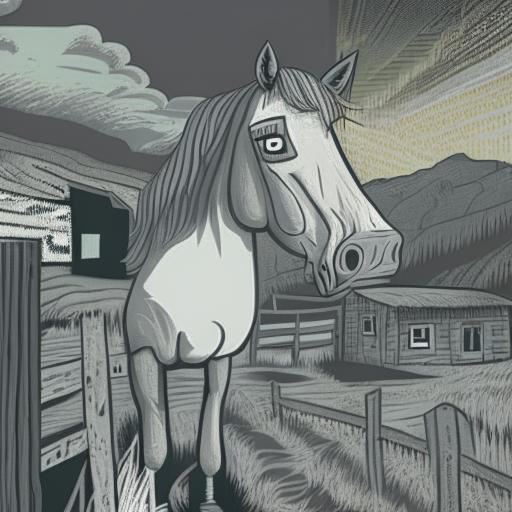} &
        \includegraphics[width=0.14\textwidth]{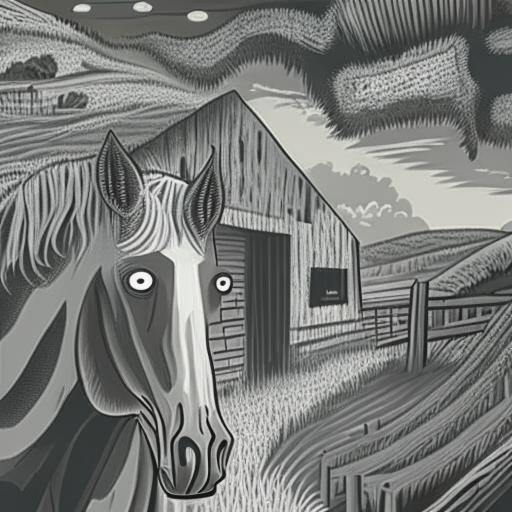} &
        \includegraphics[width=0.14\textwidth]{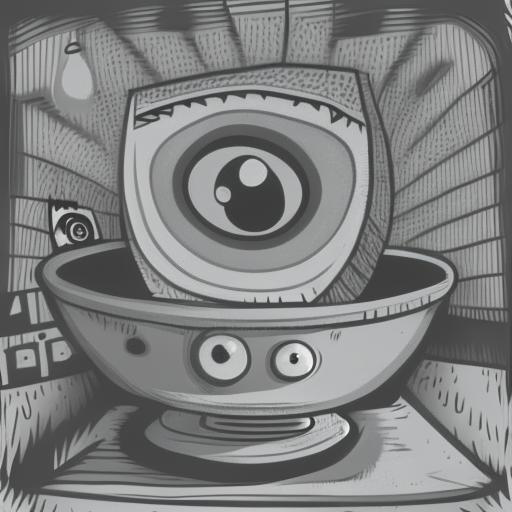} &
        \includegraphics[width=0.14\textwidth]{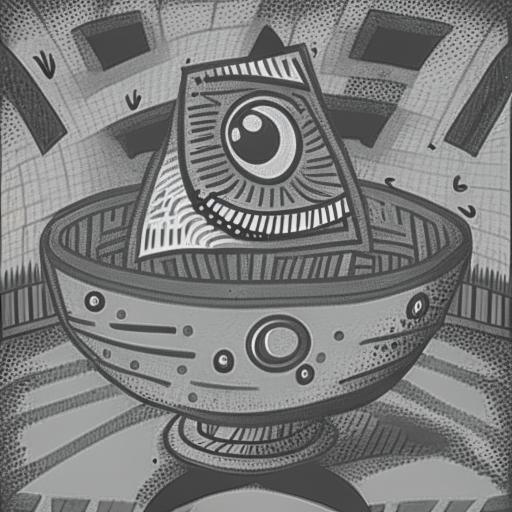} 
        \\ \\ 

        \multicolumn{2}{c}{ \begin{tabular}{c}``... a dog ...''\end{tabular}} &
        \multicolumn{2}{c}{ \begin{tabular}{c}``... a horse and a  barn in a valley ... ''\end{tabular}} &
        \multicolumn{2}{c}{\begin{tabular}{c}`` ... a bowl of fruit ... '' \end{tabular}} \\
        
    \end{tabular}
    
    }
    \caption{Styles suggested by ConceptLab using our artistic prompts with adaptive negatives. $S_*$ is always initialized as ``painting''.
    All prompts start with ``a painting of '' and end with ``in the style of $S_*$''
    }
    \label{fig:creative_art}
\end{figure*}

\begin{figure*}
    \centering
    \includegraphics[width=0.98\textwidth]{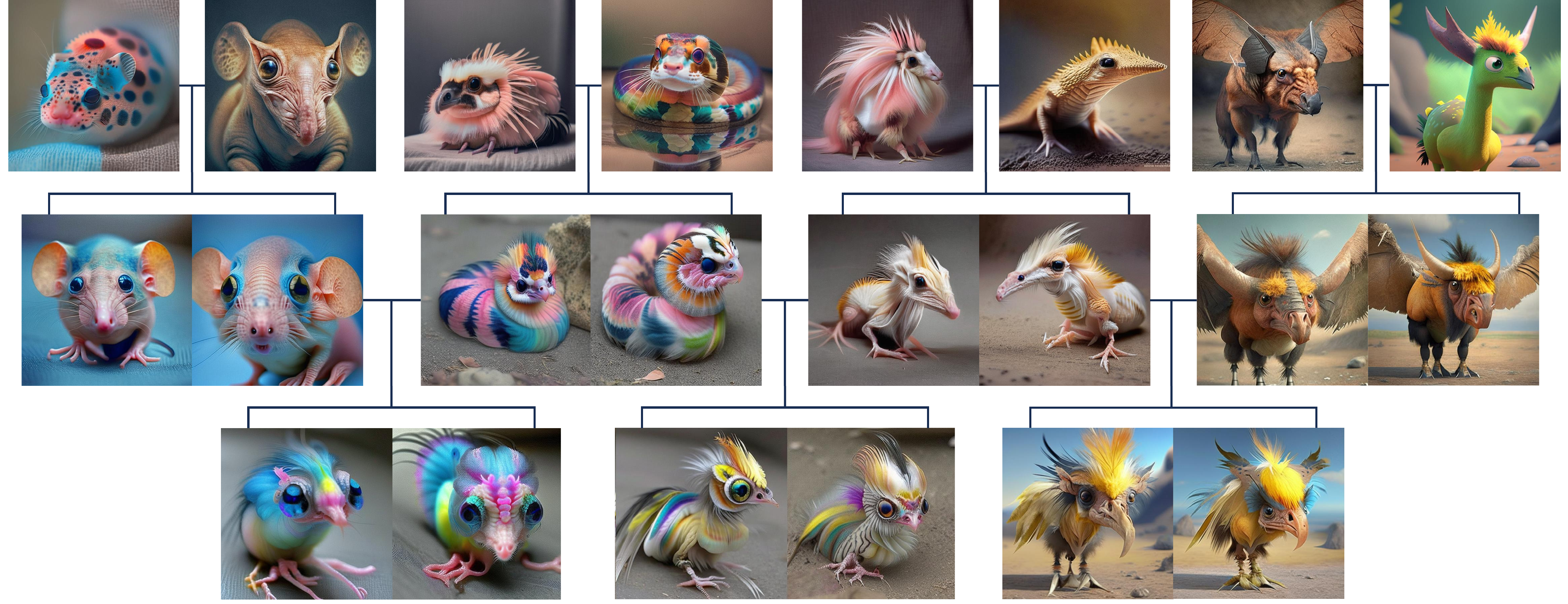}
    \caption{Evolutionary Generation. ConceptLab can be used to mix up generated concepts to iteratively learn new unique creations. In the topmost row, we show concepts learned using our adaptive negatives technique (\Cref{sec:live_negatives}) followed by concepts obtained using our evolution generation process (\Cref{sec:evolution}).}
    \label{fig:mix_tree}
\end{figure*}

\section{Experiments}

We now turn to validate the effectiveness of ConceptLab through a series of qualitative and quantitative evaluations. 

\subsection{Results}

\paragraph{Creative Generation. }
First, in~\Cref{fig:creative_results_original_token}, we demonstrate ConceptLab's ability to learn a wide range of novel creative concepts across various categories. All results are obtained using our adaptive negatives technique, highlighting our ability to generate these diverse concepts simply by varying the training seed. 

Next, as demonstrated~\Cref{fig:our_results}, ConceptLab can place these learned creative concepts in novel scenes. As shown, these generations range from background modifications and artistic styles to imagining new creations resembling the concept.
Yet, ConceptLab can go beyond generating new members of an \textit{object} category. In~\Cref{fig:creative_art} we show how ConceptLab can be used to discover new artistic styles using our adaptive negative technique. Observe how each row captures a unique style while remaining faithful to the guiding text prompt. This further highlights the advantages of our adaptive training scheme which can be applied for a variety of different categories.

\begin{figure}
    \centering
    \setlength{\tabcolsep}{0.3pt}
    \renewcommand{\arraystretch}{0.7}
    {\footnotesize
    \begin{tabular}{c c c@{\hspace{0.2cm}} c c}

        \includegraphics[width=0.09\textwidth]{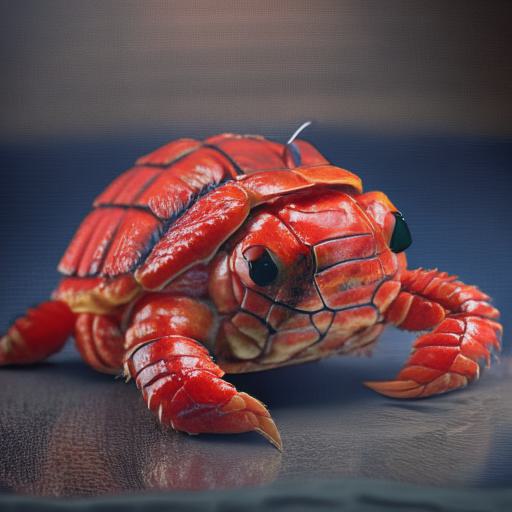} &
        \includegraphics[width=0.09\textwidth]{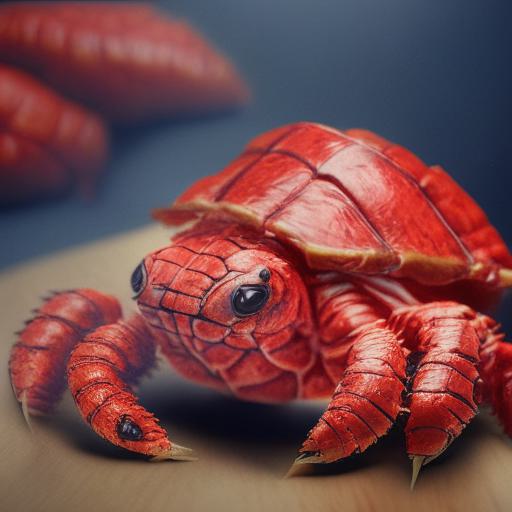} &
        \includegraphics[width=0.09\textwidth]{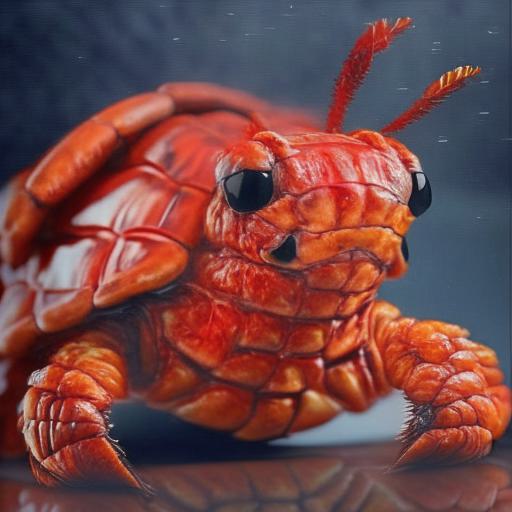} &
        \includegraphics[width=0.09\textwidth]{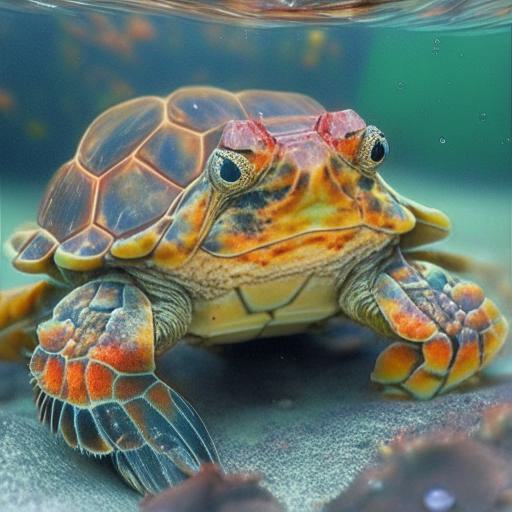} &
        \includegraphics[width=0.09\textwidth]{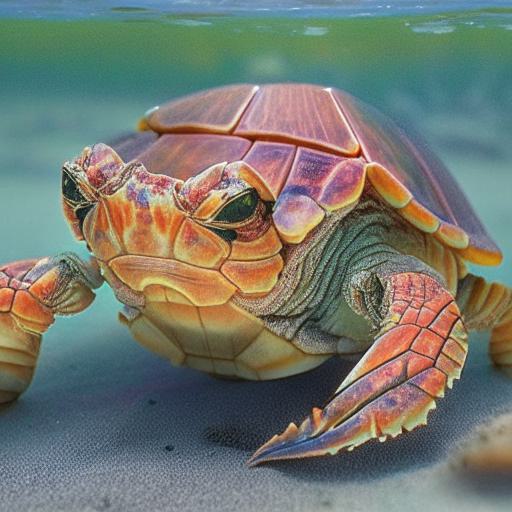} \\

        \multicolumn{3}{c}{\begin{tabular}{c} $\mathcal{C}_{pos}=\{lobster,turtle\}$ \end{tabular}} &
        \multicolumn{2}{c}{\begin{tabular}{c} ``A photo of a \textcolor{blue}{lobster} that \\ looks like a \textcolor{blue}{turtle}'' \end{tabular}} \\ \\[-0.1cm]

        \includegraphics[width=0.09\textwidth]{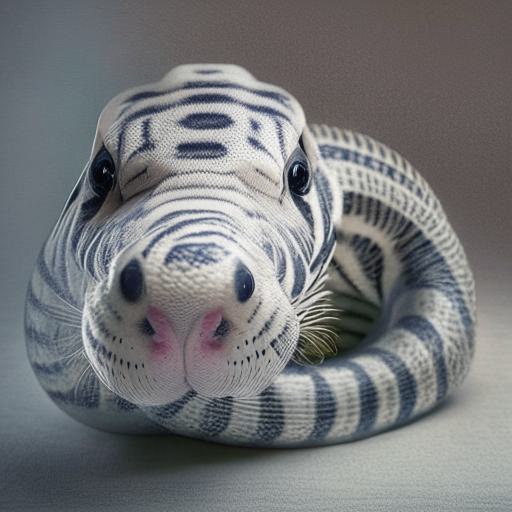} &
        \includegraphics[width=0.09\textwidth]{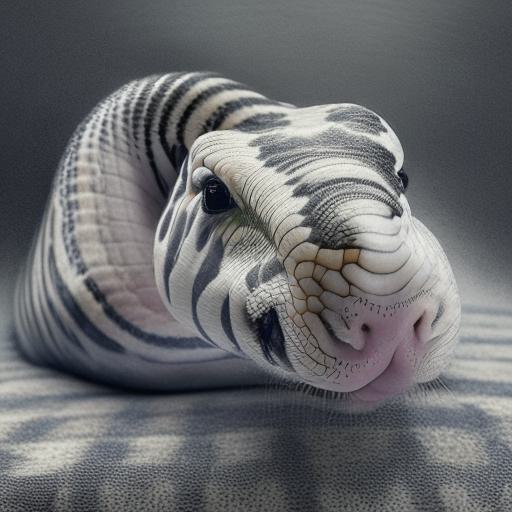} &
        \includegraphics[width=0.09\textwidth]{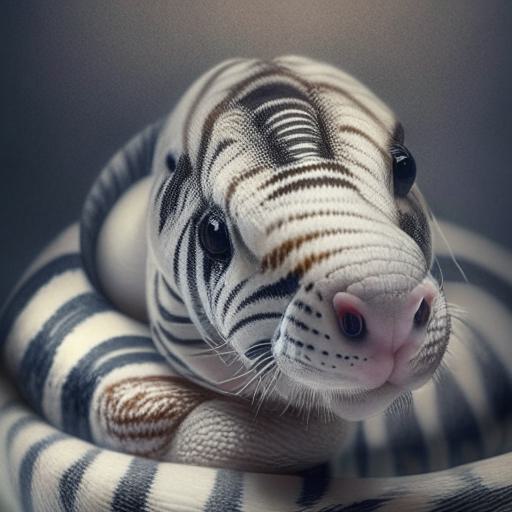} &
        \includegraphics[width=0.09\textwidth]{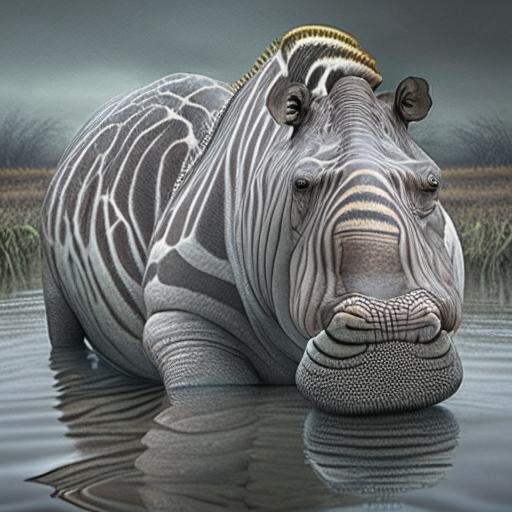} &
        \includegraphics[width=0.09\textwidth]{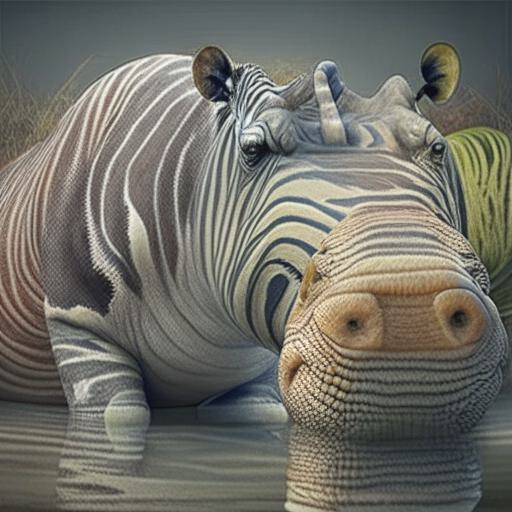} \\

        \multicolumn{3}{c}{\begin{tabular}{c} $\mathcal{C}_{pos}=\{snake,hippo, zebra\}$ \end{tabular}} &
        \multicolumn{2}{c}{\begin{tabular}{c} ``An animal that resembles  a \\ \textcolor{blue}{snake}, \textcolor{blue}{hippo}, and \textcolor{blue}{zebra}'' \end{tabular}} \\ \\[-0.1cm]

        \includegraphics[width=0.09\textwidth]{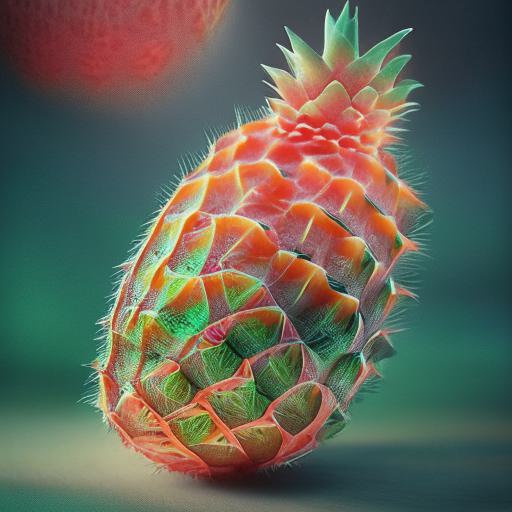} &
        \includegraphics[width=0.09\textwidth]{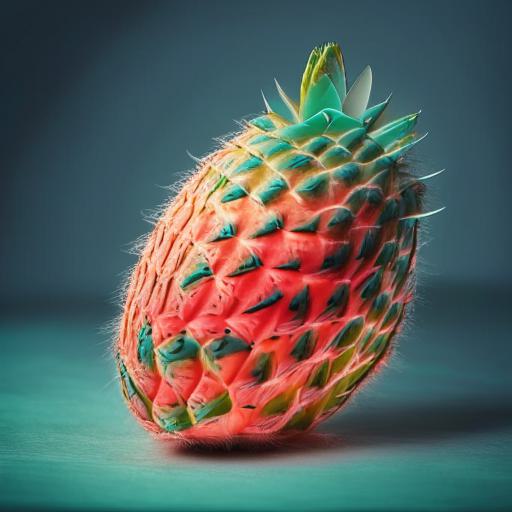} &
        \includegraphics[width=0.09\textwidth]{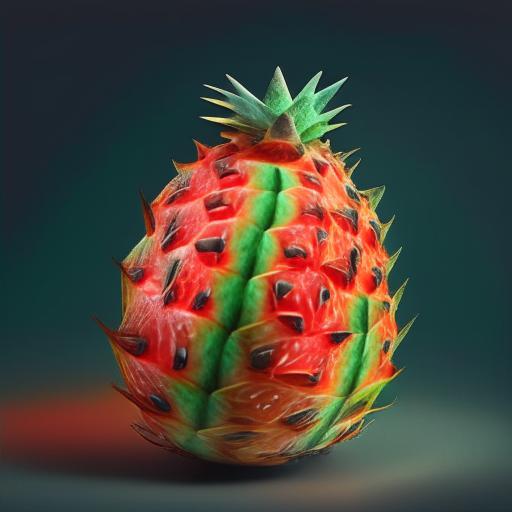} &
        \includegraphics[width=0.09\textwidth]{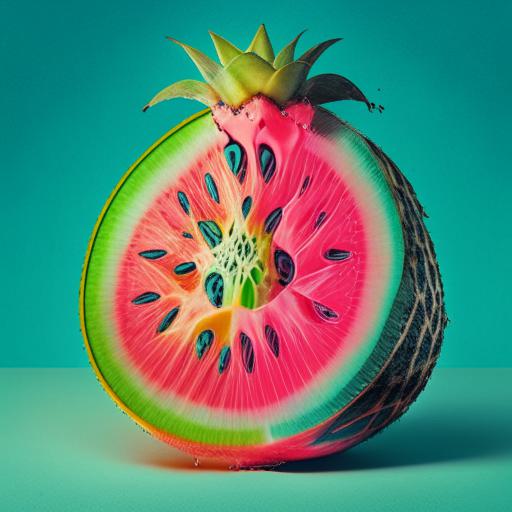} &
        \includegraphics[width=0.09\textwidth]{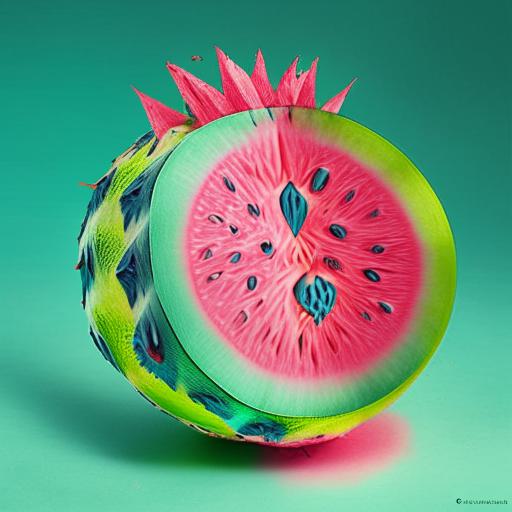} \\

        \multicolumn{3}{c}{\begin{tabular}{c} $\mathcal{C}_{pos}=\{pineapple,watermelon\}$ \end{tabular}} &
        \multicolumn{2}{c}{\begin{tabular}{c} ``A \textcolor{blue}{pineapple} with the \\ colors of a \textcolor{blue}{watermelon}'' \end{tabular}} \\ \\[-0.1cm]
       
    \end{tabular}
    }
    \vspace{-0.3cm}
    \caption{Mixing results obtained with ConceptLab. On the left, we show images generated using a concept learned by ConceptLab using positive constraints. On the right, we show results obtained with Kandinsky using curated prompts that aim to achieve a mixing result. 
    }
    \label{fig:mixing_real_concepts}
\end{figure}

\paragraph{Concept Mixing. }
In~\Cref{fig:mixing_real_concepts} we show how we can form hybrid concepts by merging unique traits across multiple real concepts using only positive constraints. Observe, for example, the first row where we are able to capture key characteristics of the lobster (e.g., its color and claws) and fuse them with those of a turtle (e.g., its shell). Moreover, in the second row, we are able to fuse three concepts, capturing the body of the snake, the texture of the zebra, and the head of the hippopotamus. To illustrate that learning such combinations of concepts is non-trivial, we attempt to achieve a similar mixture using hand-crafted prompts. As shown on the right-hand side of~\Cref{fig:mixing_real_concepts}, such prompts fail to capture key aspects of all desired concepts.

\paragraph{Evolutionary Generation. }
We next explore our ability to mix various learned concepts using our evolution generation procedure, as described in~\Cref{sec:evolution}. In~\Cref{fig:mix_tree}, we show results obtained across multiple ``generations'' of concepts learned by ConceptLab. 
For example, consider the leftmost mixing in the provided family tree. Observe how we are able to fuse the color and general shape of the left parent with the distinct ears of the right parent to obtain a plausible blue-like rat mammal. 
We can then continue this evolutionary mix-up process across multiple generations as shown in the bottom-most row. 

\subsection{Comparisons}

\paragraph{Evaluation Setup. }
While no related work tackles the exact same problem as ConceptLab, a natural baseline arises from the negative prompting technique~\cite{liu2022compositional}, which has become a prominent technique in text-to-image generation. 
In the context of creative generation, it can  potentially be used to generate novel concepts by defining a negative prompt that includes the negative constraints. We compare ConceptLab to two such baselines. Specifically, we consider both Stable Diffusion 2~\cite{rombach2022high} and Kandinsky 2.1~\cite{kandinsky2} and generate images using an input prompt of the form ``A photo of a $c_{pos}$'' where $c_{pos}$ is our positive token (e.g., ``pet'') and a negative prompt of the form ``A photo of a $c_{neg,1}$,\dots, A photo of a $c_{neg,k}$'' where $c_{neg,1},\dots,c_{neg,k}$ are our negative tokens (e.g., ``cat'', ``dog'', ``hamster''). For Kandinsky, the negative prompt is applied over the Diffusion Prior and not the Latent Diffusion, as it empirically resulted in more favorable results.

\begin{figure}
    \centering
    \setlength{\tabcolsep}{1pt}
    \renewcommand{\arraystretch}{0.5}
    {\small
    \begin{tabular}{c c@{\hspace{0.1cm}} c c@{\hspace{0.1cm}} c c@{\hspace{0.1cm}}}
        \includegraphics[width=0.075\textwidth]{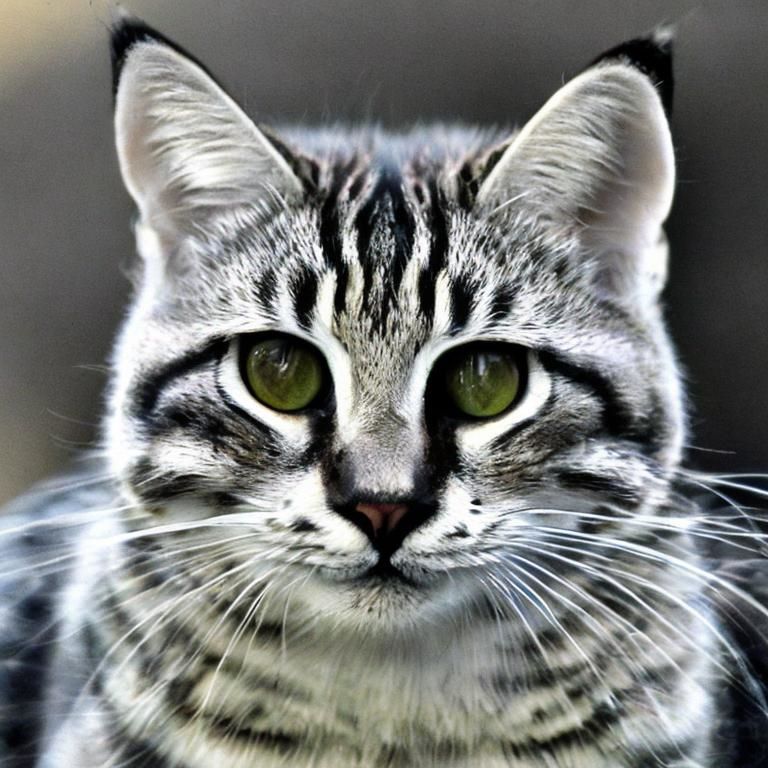} &
        \includegraphics[width=0.075\textwidth]{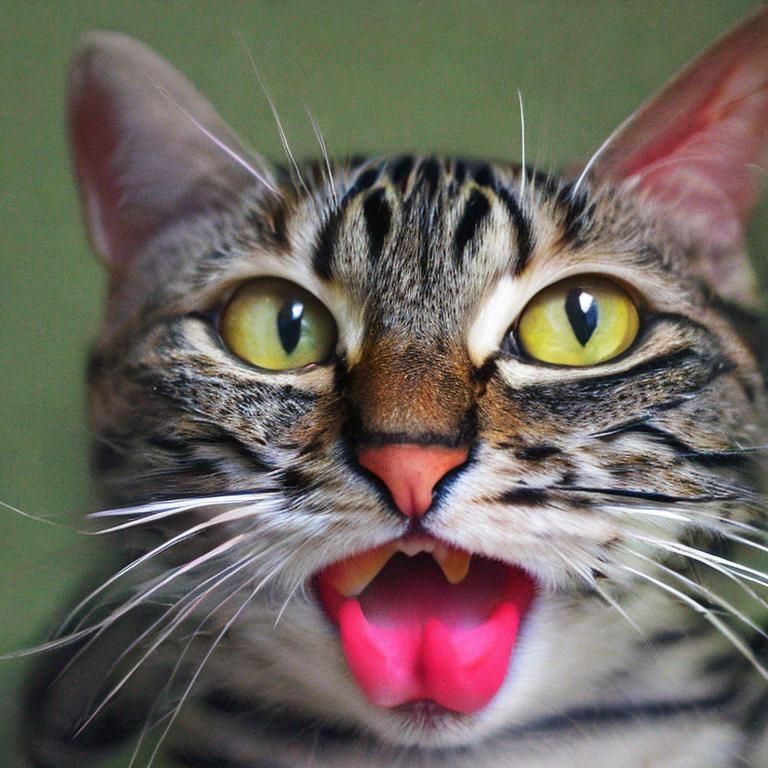} &
        \includegraphics[width=0.075\textwidth]{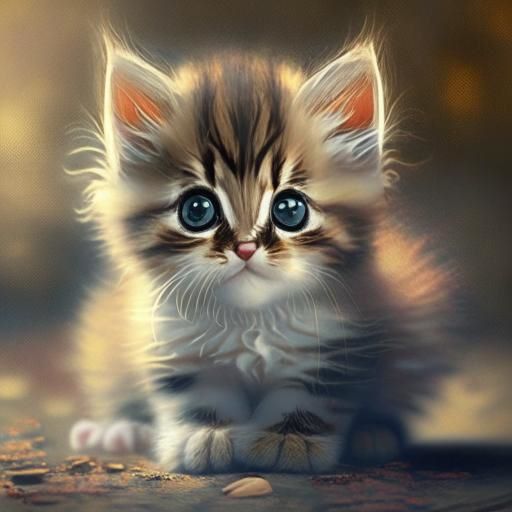} &
        \includegraphics[width=0.075\textwidth]{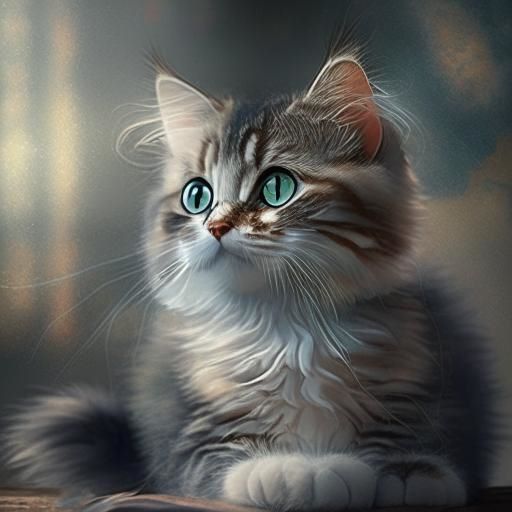} &
        \includegraphics[width=0.075\textwidth]{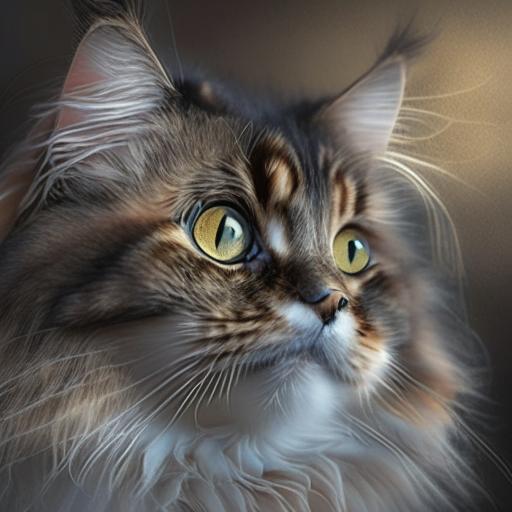} &
        \includegraphics[width=0.075\textwidth]{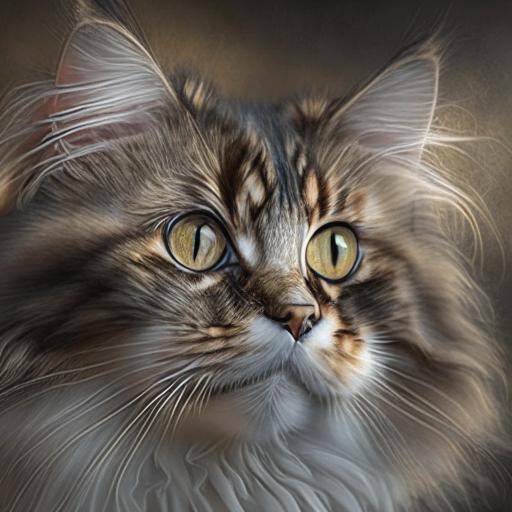} \\
        \multicolumn{6}{c}{{\begin{tabular}{c} \textcolor{mygreen}{+ pet}, \textcolor{red}{\textendash \mbox{} dog}  \end{tabular}}} \\ \\

        \includegraphics[width=0.075\textwidth]{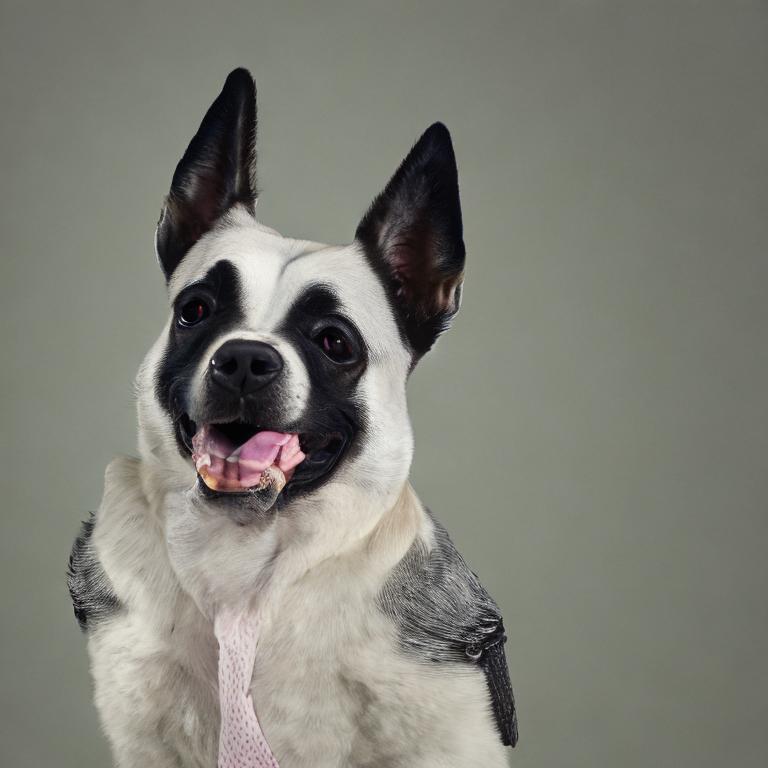} &
        \includegraphics[width=0.075\textwidth]{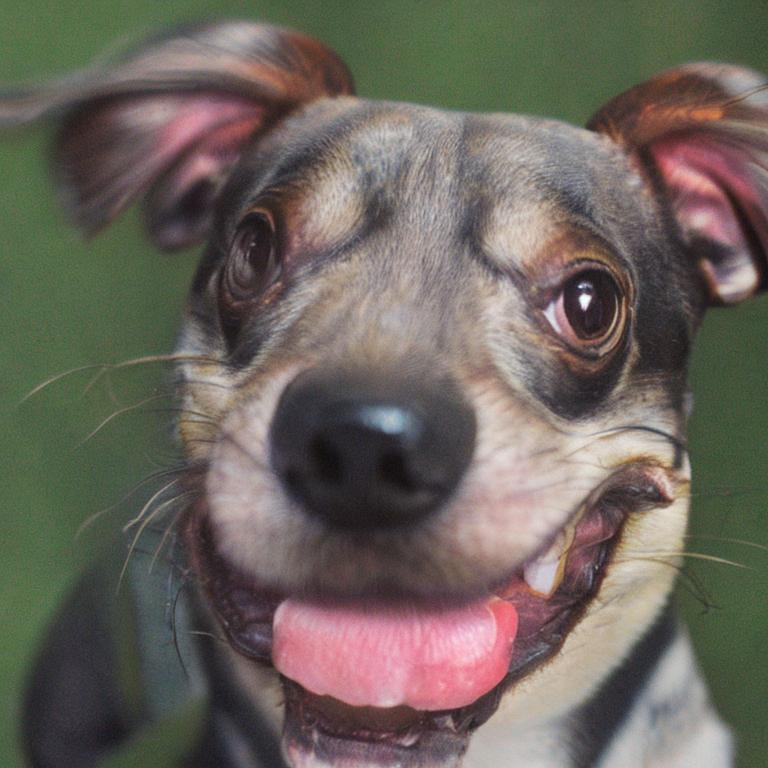} &
        \includegraphics[width=0.075\textwidth]{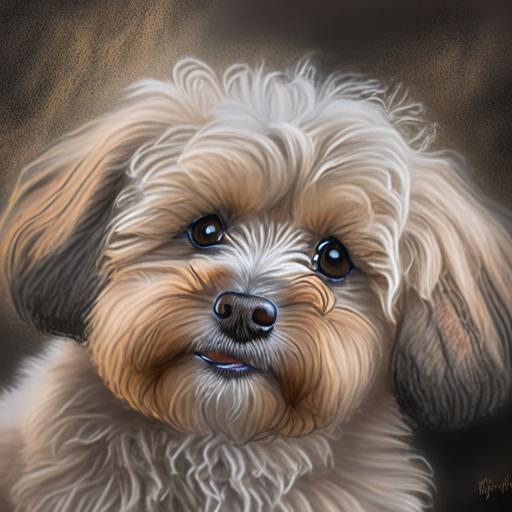} &
        \includegraphics[width=0.075\textwidth]{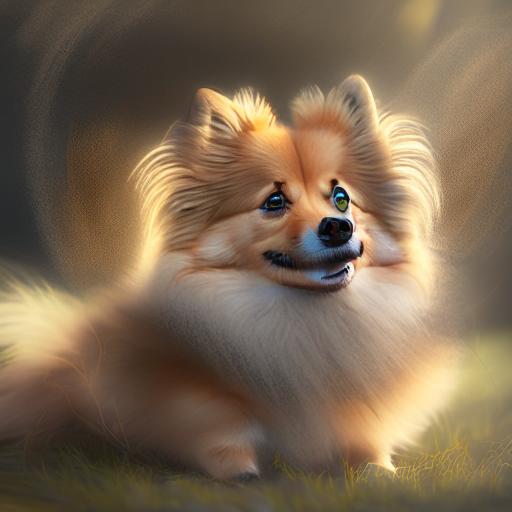} &
        \includegraphics[width=0.075\textwidth]{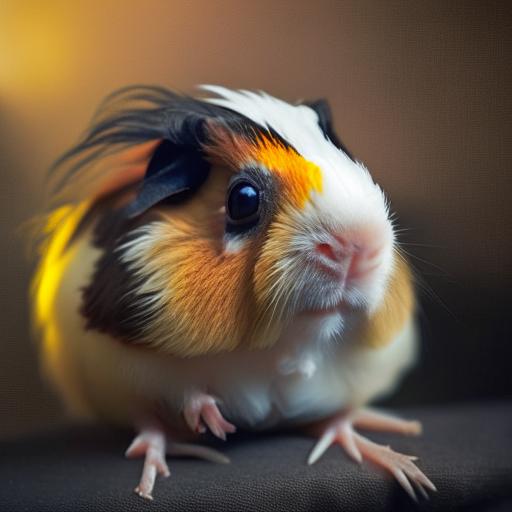} &
        \includegraphics[width=0.075\textwidth]{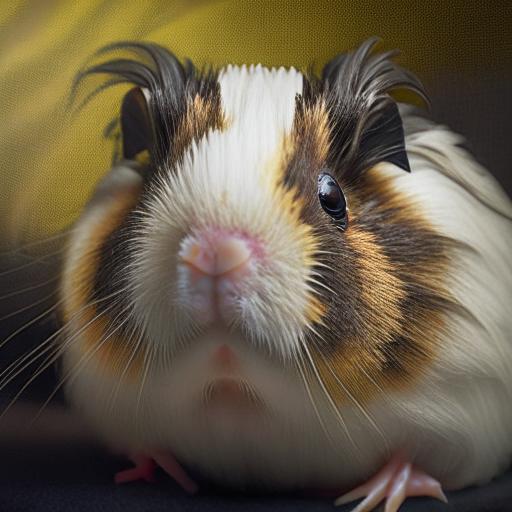} \\
        \multicolumn{6}{c}{{\begin{tabular}{c} \textcolor{mygreen}{+ pet}, \textcolor{red}{\textendash \mbox{} dog, cat}  \end{tabular}}} \\ \\

        \includegraphics[width=0.075\textwidth]{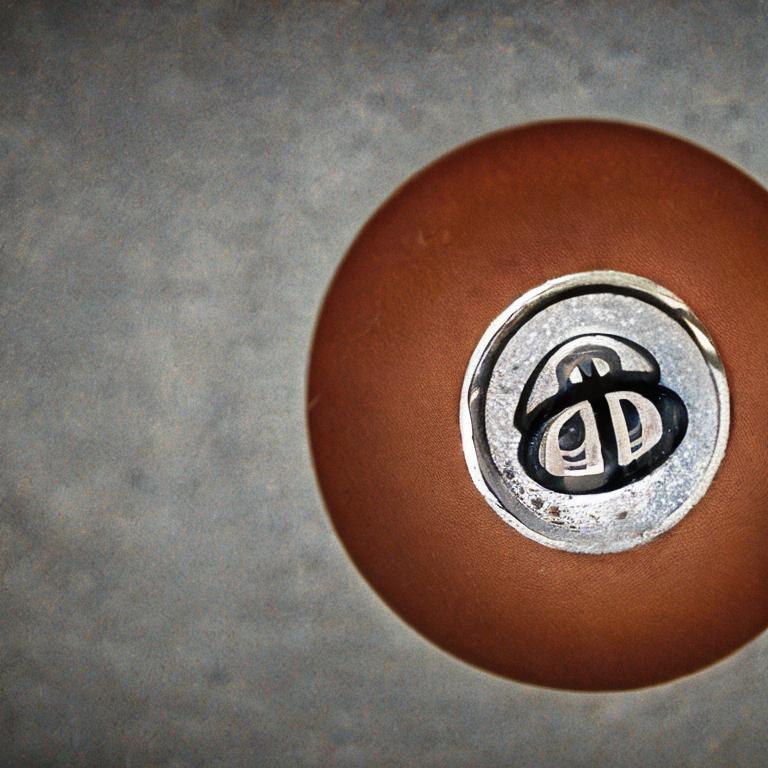} &
        \includegraphics[width=0.075\textwidth]{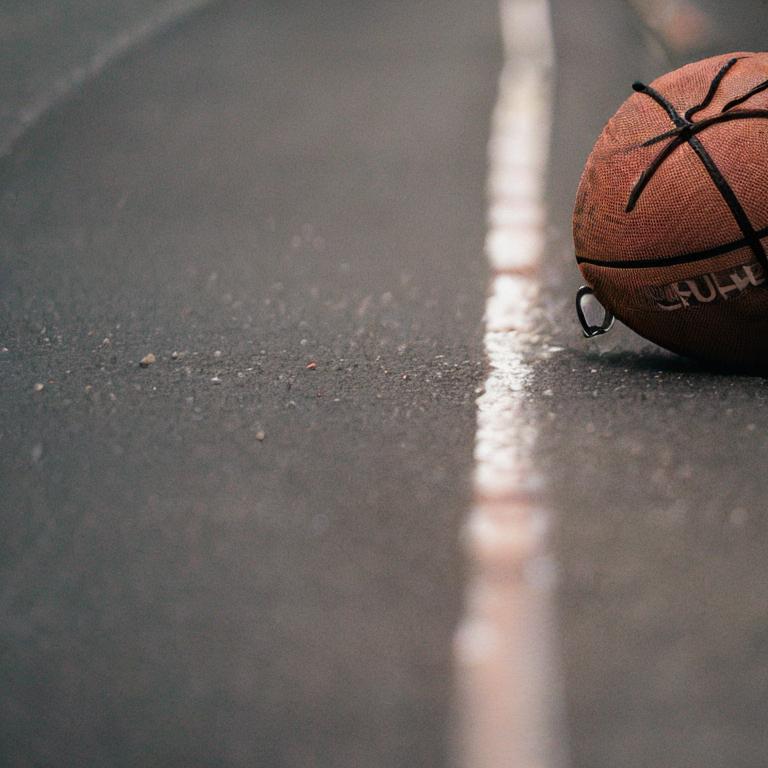} &
        \includegraphics[width=0.075\textwidth]{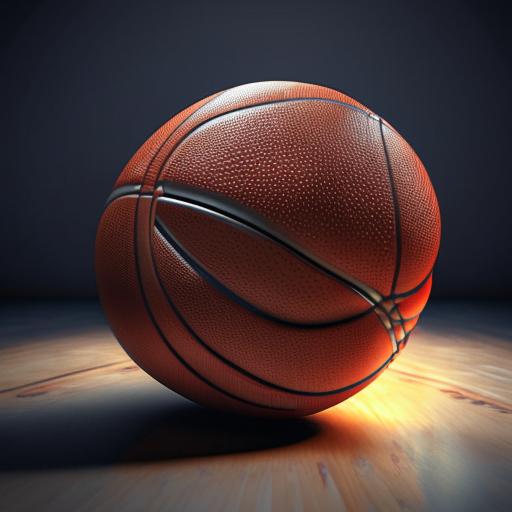} &
        \includegraphics[width=0.075\textwidth]{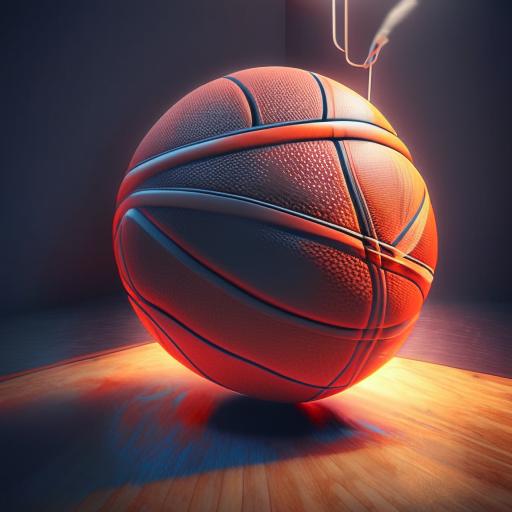} &
        \includegraphics[width=0.075\textwidth]{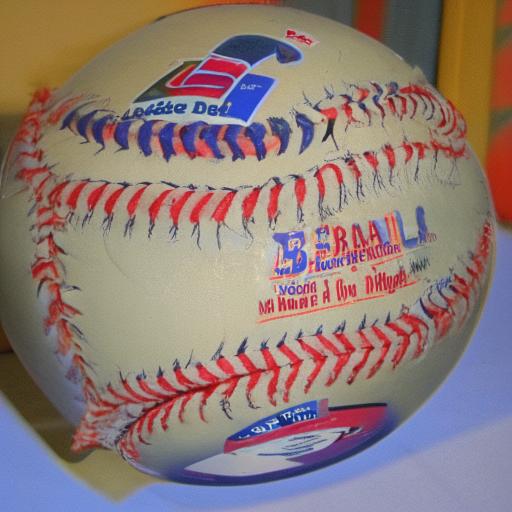} &
        \includegraphics[width=0.075\textwidth]{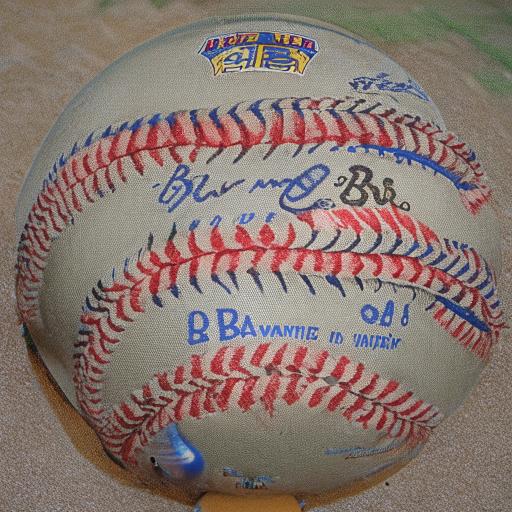} \\
        \multicolumn{6}{c}{\begin{tabular}{c} \textcolor{mygreen}{+ sports ball}, \textcolor{red}{\textendash{} soccer ball, volleyball, basketball,}\\ \textcolor{red}{football, golf ball, tennis ball} \end{tabular}} \\ \\

        \includegraphics[width=0.075\textwidth]{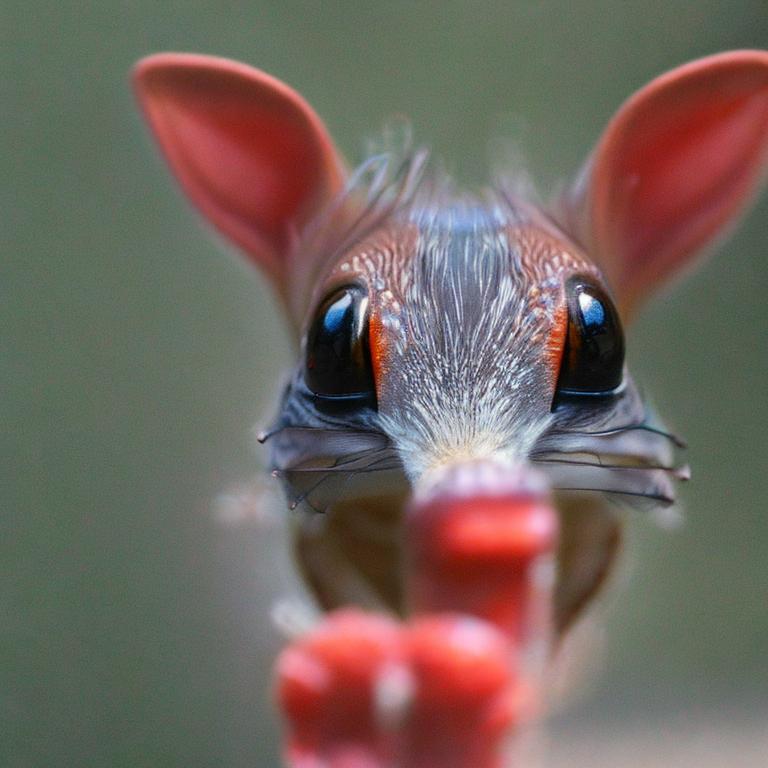} &
        \includegraphics[width=0.075\textwidth]{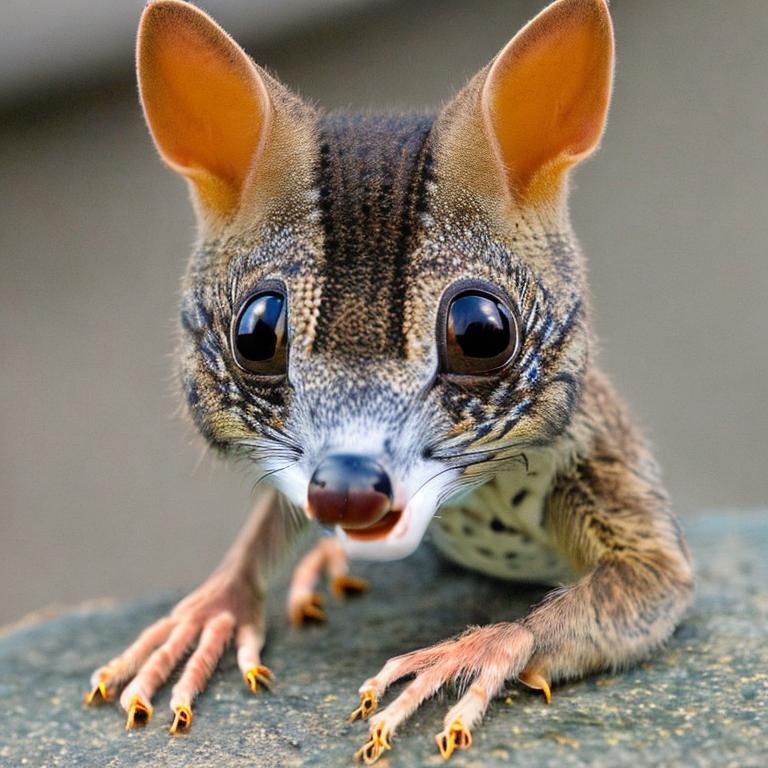} &
        \includegraphics[width=0.075\textwidth]{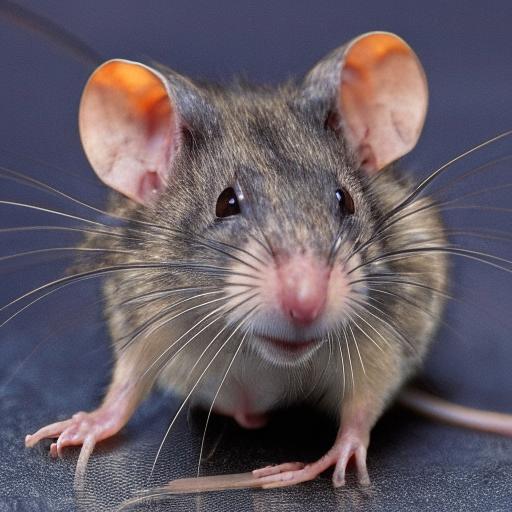} &
        \includegraphics[width=0.075\textwidth]{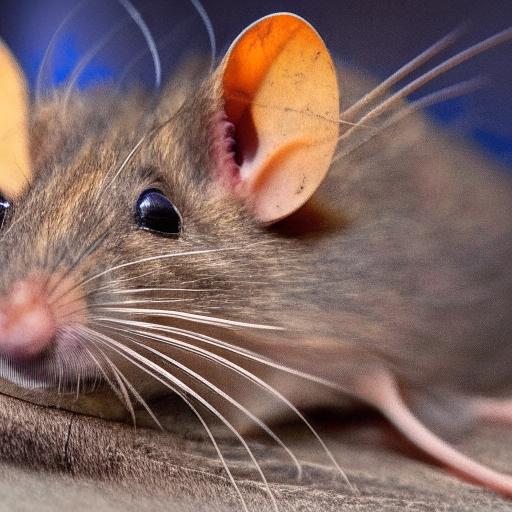} &
        \includegraphics[width=0.075\textwidth]{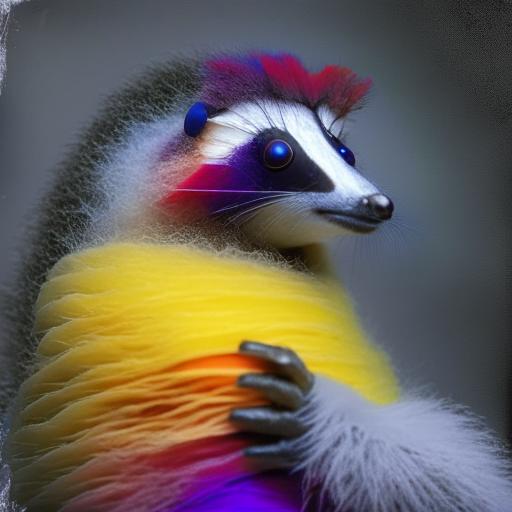} &
        \includegraphics[width=0.075\textwidth]{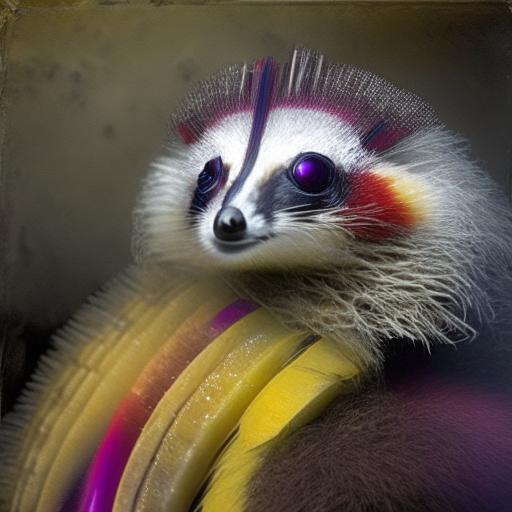} \\
        \multicolumn{6}{c}{\begin{tabular}{c} \textcolor{mygreen}{+ rodent}, \textcolor{red}{\textendash \mbox{} mouse, hamster, rat, beaver, otter} \end{tabular}} \\ \\

        \includegraphics[width=0.075\textwidth]{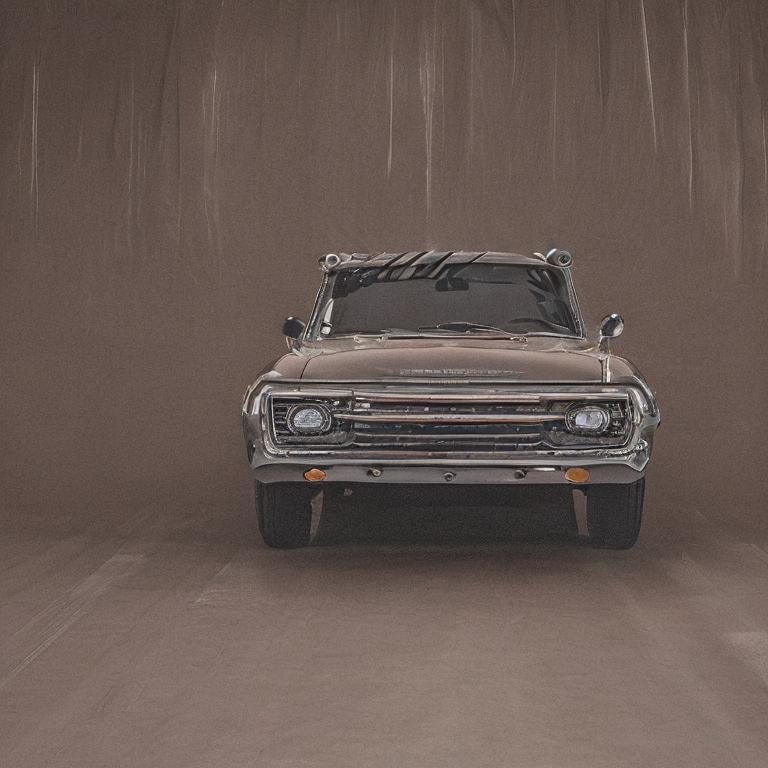} &
        \includegraphics[width=0.075\textwidth]{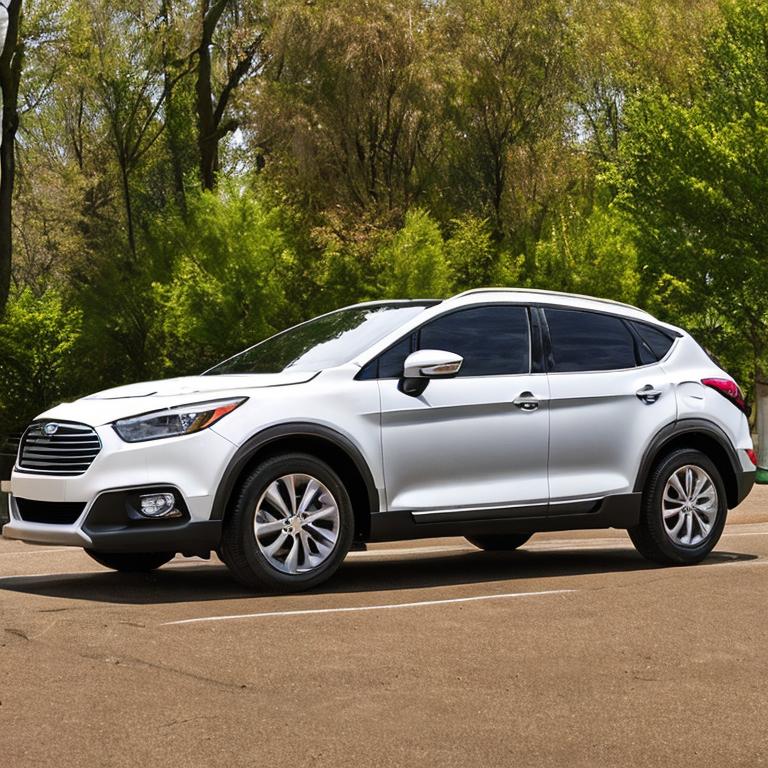} &
        \includegraphics[width=0.075\textwidth]{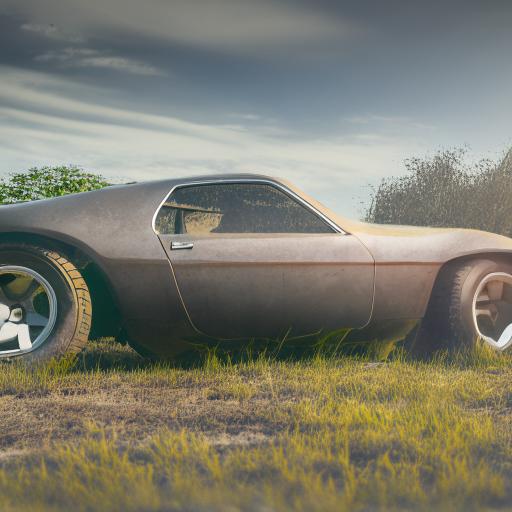} &
        \includegraphics[width=0.075\textwidth]{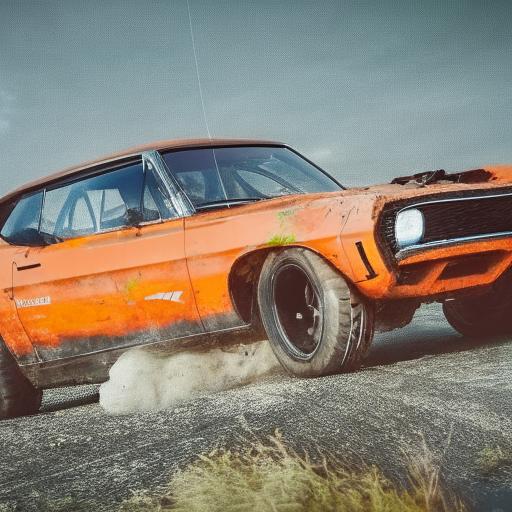} &
        \includegraphics[width=0.075\textwidth]{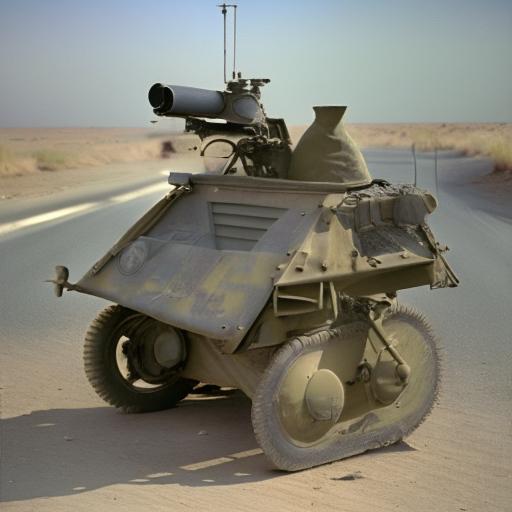} &
        \includegraphics[width=0.075\textwidth]{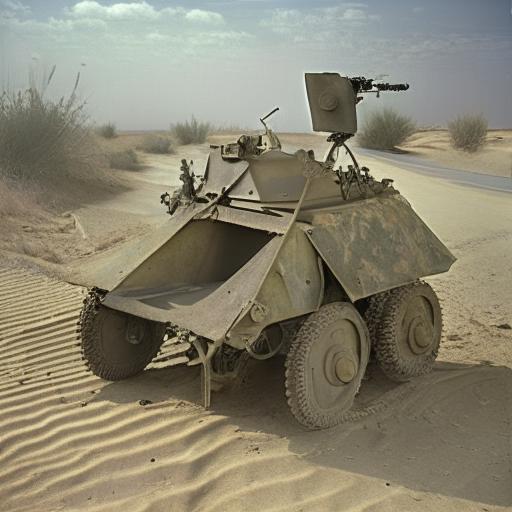} \\
        \multicolumn{6}{c}{\begin{tabular}{c} \textcolor{mygreen}{+ vehicle}, \textcolor{red}{\textendash \mbox{} bus, truck, private car} \end{tabular}} \\ \\

        \multicolumn{2}{c}{{\begin{tabular}{c} Stable Diffusion \end{tabular}}} &
        \multicolumn{2}{c}{{\begin{tabular}{c} Kandinsky  \end{tabular}}} &
        \multicolumn{2}{c}{{\begin{tabular}{c} ConceptLab \end{tabular}}} 

    \end{tabular}
    
    }
    \caption{Comparison to negative prompting. For both Stable Diffusion and Kandinsky, a negative prompt was composed containing all specified classes. 
    }
    \label{fig:ablations_sd_neg_prompts}
\end{figure}

\paragraph{Qualitative Comparisons. }

In~\Cref{fig:ablations_sd_neg_prompts} we compare ConceptLab to the training-free baselines. As can be seen, while negative prompting does work when a single constraint is used, the baselines generally do not perform well when faced with multiple constraints. Specifically, even when tasked with generating a ``pet'' with both ``cat'' and ``dog'' explicitly stated in the negative prompt, both approaches tend to generate images of dogs.
Conversely, ConceptLab is able to consistently align with both the positive token and negative constraints. We further note that the training-free baselines do not learn a consistent representation of a \textit{specific} concept, and hence do not allow for the same editing capabilities as ConceptLab.

\paragraph{Quantitative Comparisons. }
We now turn to quantitatively evaluate the considered methods using a CLIP-based evaluation scheme. Specifically, we evaluate the ability of each method to (1) capture the positive concept while (2) generating images that do not resemble any of the given negative concepts. We consider five broad categories: pets, plants, fruits, furniture, and musical instruments. For each domain, we consider three different pairs of negative concepts (e.g., ``cat'' and ``dog'', ``closet'' and ``bed'', etc.) and train ConceptLab using five random seeds for each combination, resulting in a total of $75$ concepts. For each concept, we then generate $32$ images using the prompt ``A photo of a $S_*$'', resulting in $160$ images for each positive-negative combination. 
For Stable Diffusion and Kandinsky, we use negative prompting and generate $160$ images for the same sets of positive and negative concept pairs.

\begin{figure}[b]
    \centering
    \includegraphics[width=0.475\textwidth]{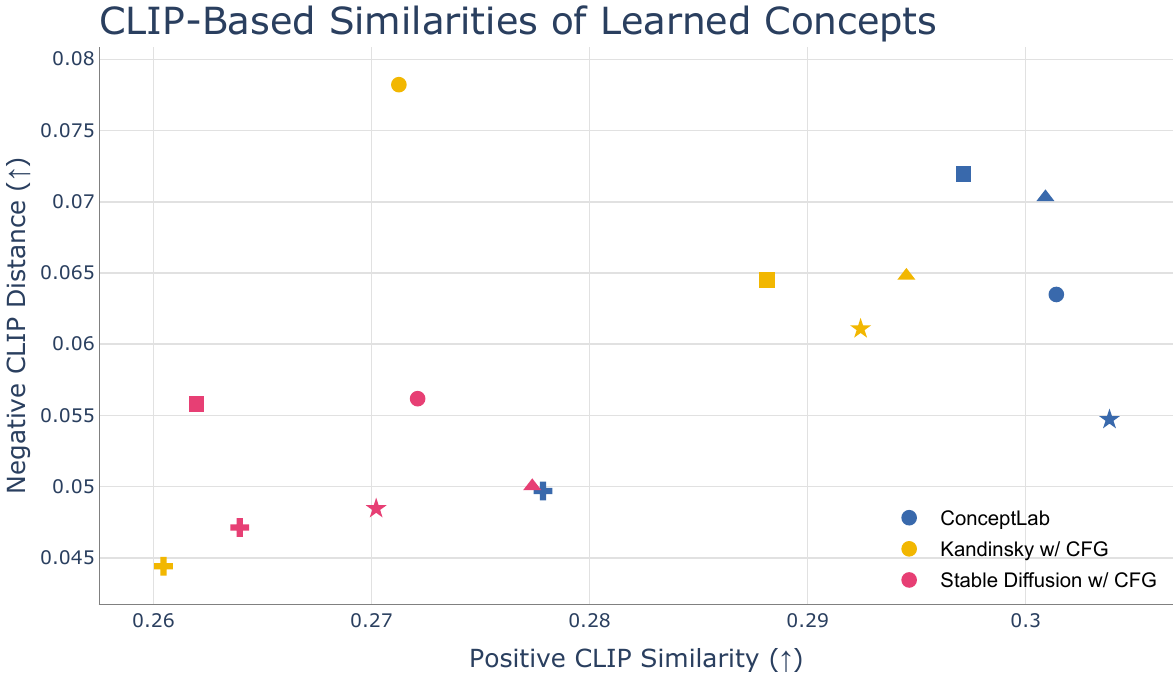}
    \vspace{-0.5cm}
    \caption{Quantitative evaluation. We compare ConceptLab to Kandinsky~\cite{kandinsky2} and Stable Diffusion~\cite{rombach2022high} with classifier-free guidance using negative prompting. 
    For each, we compute (1) the similarity between the generated images and the positive concept, and (2) the difference between the positive similarity and the maximum negative similarity between the generated images and all negative concepts. 
    Results are averaged across each category separately.
    The domains are represented by: pet: $\circ$, plant: $\square$, fruit: $\star$, furniture: $+$, musical instrument: $\bigtriangleup$.    
    }
    \label{fig:quantitative}
    
\end{figure}

We define two measurements that are jointly used to measure and compare the different methods.  First, we compute the positive similarity of each concept to the target category by calculating the CLIP-space similarity between the embeddings of all generated images and the text prompt ``A photo of a $c_{pos}$'', where $c_{pos}$ is our positive concept. 
Next, we compute a measurement of the distance between the positive constraints and the negative constraints. This is done by first calculating the maximum similarity between the generated images and all negative concepts. We then compute the difference between the previously computed positive similarity and the maximum negative similarity. This measures the method's ability to stay away from negative constraints, while also penalizing predictions that are out of distribution. (Consider the case where the target concept is a ``pet'' and the negative constraints are ``cat'' and ``dog'', but the generated images resemble a ``fruit''. The negative similarity between the images and the constraints would be low, but this is still an undesirable solution).
Together, the metrics capture both the ability of the method to remain close to the positive class, while distinguishing its concepts from the negative constraints. 

The results are illustrated in~\Cref{fig:quantitative}.
As can be seen, ConceptLab consistently outperforms both baselines in positive CLIP similarity across all five domains, indicating that ConceptLab is able to faithfully generate images belonging to the target broad category. In terms of our negative distance metric, ConceptLab outperforms Stable Diffusion in all categories while outperforming Kandinsky in four of the five categories. This indicates that ConceptLab is able to generate images that belong to the target category, but differ significantly from existing concepts.

\paragraph{User Study. }
We additionally conduct a user study to compare ConceptLab to the negative prompting techniques. We follow the same evaluation setup as above and generate images using each method belonging to five different broad categories. We then asked respondents to rate the images generated by each method based on their ability to both capture the target broad concept category and differ from the specified negative concepts. Respondents were asked to rate each set of results on a scale from $1$ to $5$. 
Results are shown in~\Cref{tb:user_study}. In total, we had $30$ respondents, for a total of $300$ ratings per method. As shown, participants heavily favored ConceptLab when compared to both baselines.

\begin{table}
\caption{User Study.
We asked respondents to rate images on a scale of $1$ to $5$ based on how well they respect a given set of constraints.
}
\label{tb:user_study}
\small
\centering
\setlength{\tabcolsep}{2pt}   
\begin{tabular}{l c c c} 
    \toprule
    & \begin{tabular}{c} Stable Diffusion \end{tabular} & 
      \begin{tabular}{c} Kandinsky  \end{tabular} &  
      \textbf{ConceptLab}  \\
    \midrule
    Average Rating ($\uparrow$) & 1.90 $\pm$ 1.11 & 1.79 $\pm$ 1.16 & \textbf{3.77} $\pm$ \textbf{1.35} \\
    \bottomrule
\end{tabular}

\end{table}

\subsection{Additional Analysis}

\paragraph{Using the Prior. }
We now turn to validate the use of our prior constraints. To this end, we compare ConceptLab to two baselines. First, we consider ConceptLab \textit{without} passing the text encoding through the Diffusion Prior, a method which we call CLIP-ConceptLab, as all objectives from~\Cref{eq:objective} are computed over the text conditioning space, $E_y(\cdot)$. 
Next, we compare to a variant of ConceptLab using Stable Diffusion~\cite{rombach2022high}. Specifically, we collect images of each negative class and apply our CLIP-space constraints between the collected images and denoised images $x_0$ computed throughout training using a DDIM scheduler~\cite{song2021denoising}. We note this is not an existing method but rather our attempt to ``implement'' ConceptLab with Stable Diffusion, which we call SD-ConceptLab.

\begin{figure}
    \centering
    \setlength{\tabcolsep}{0pt}
    \renewcommand{\arraystretch}{1.0}
    {\small
    \begin{tabular}{c c@{\hspace{0.1cm}} c c@{\hspace{0.1cm}} c c@{\hspace{0.1cm}}}

        \includegraphics[width=0.075\textwidth]{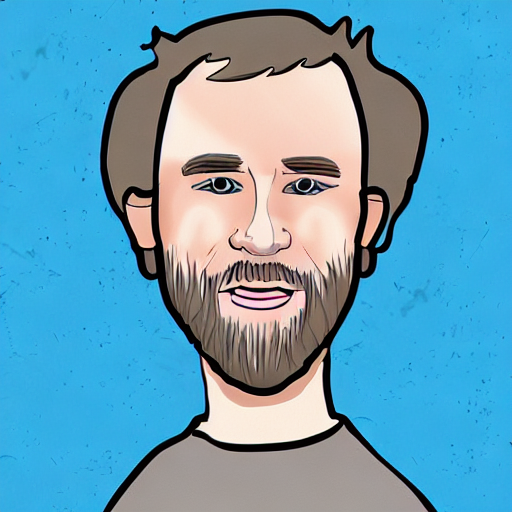} &
        \includegraphics[width=0.075\textwidth]{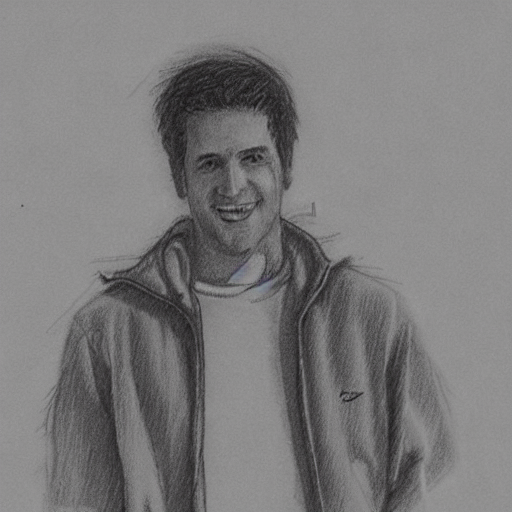} &
        \includegraphics[width=0.075\textwidth]{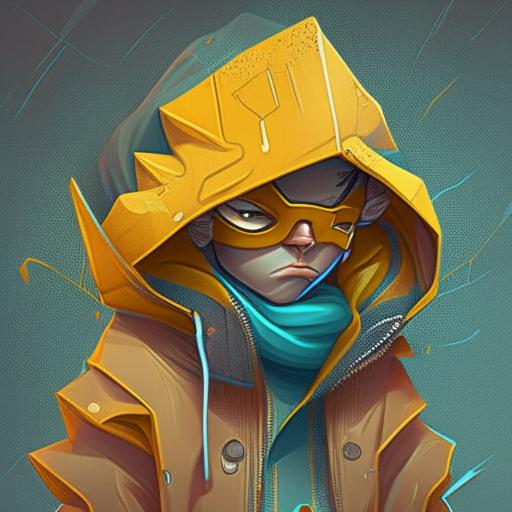} &
        \includegraphics[width=0.075\textwidth]{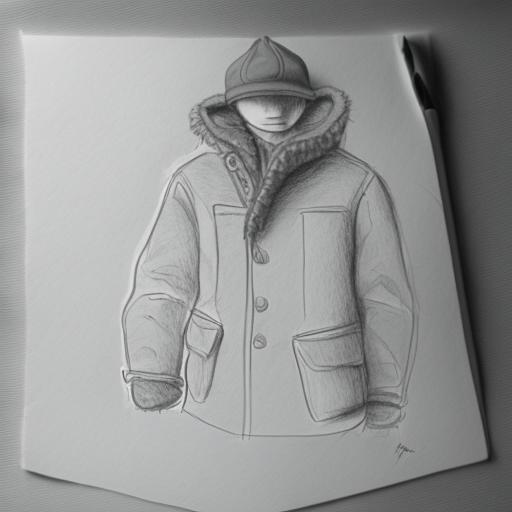} &
        \includegraphics[width=0.075\textwidth]{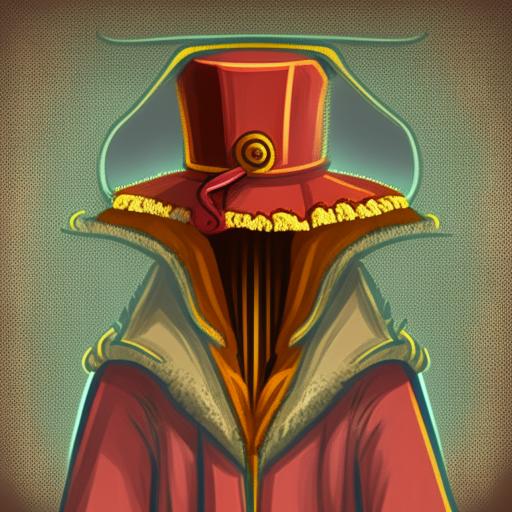} &
        \includegraphics[width=0.075\textwidth]{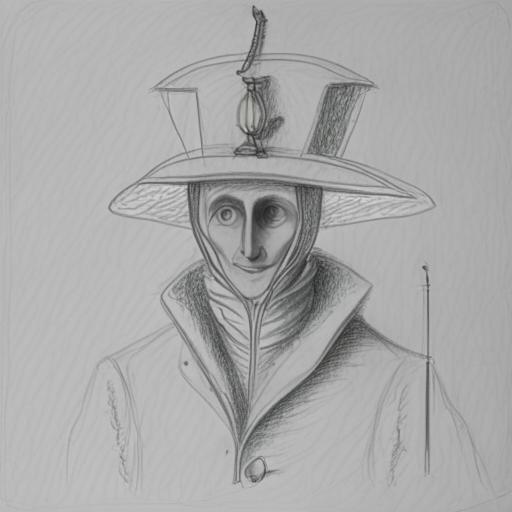} \\
        \multicolumn{6}{c}{\begin{tabular}{c} \textcolor{mygreen}{+ garment}, \textcolor{red}{\textendash \mbox{} shirt, dress, pants, skirt}  \end{tabular}} \\ 

        \includegraphics[width=0.075\textwidth]{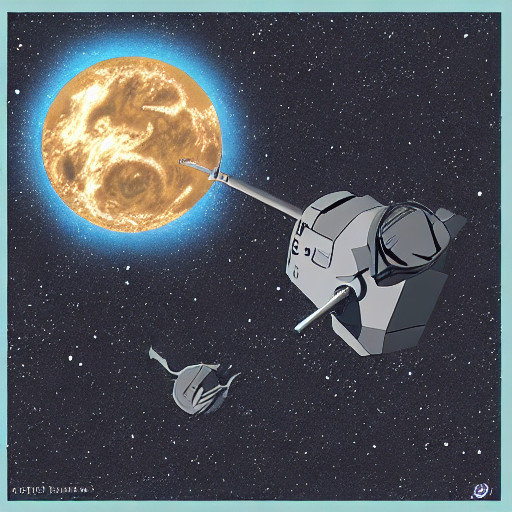} &
        \includegraphics[width=0.075\textwidth]{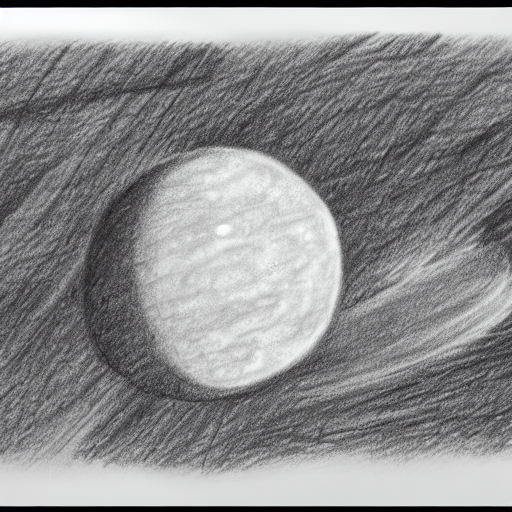} &
        \includegraphics[width=0.075\textwidth]{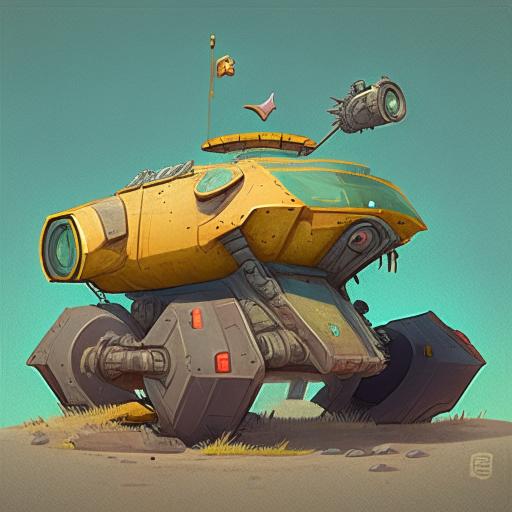} &
        \includegraphics[width=0.075\textwidth]{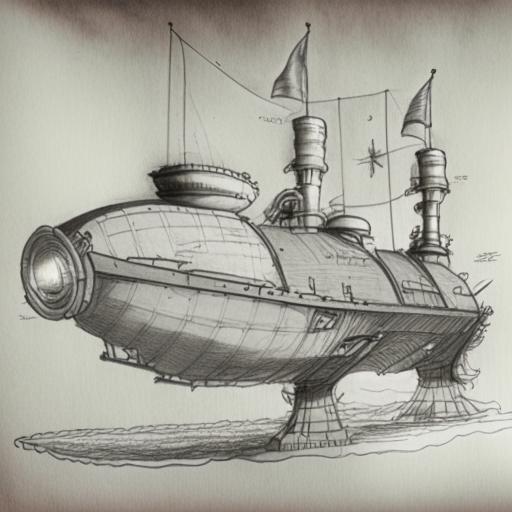} &
        \includegraphics[width=0.075\textwidth]{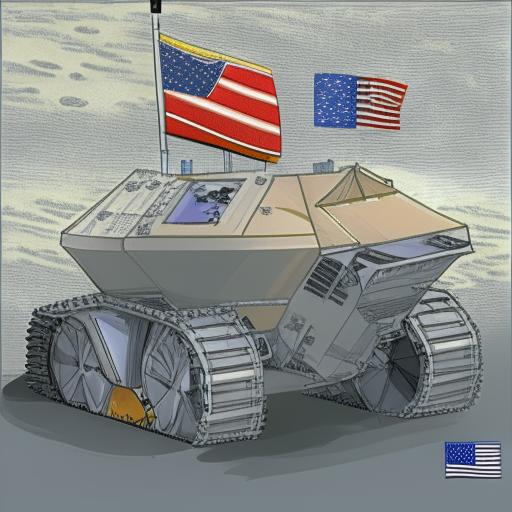} &
        \includegraphics[width=0.075\textwidth]{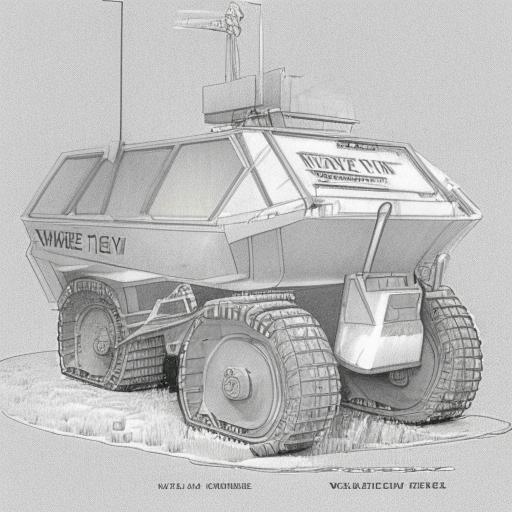} \\
        \multicolumn{6}{c}{{\begin{tabular}{c} \textcolor{mygreen}{+ vehicle}, \textcolor{red}{\textendash \mbox{} car, truck, motorcycle, bus, minibus}  \end{tabular}}} \\

        \includegraphics[width=0.075\textwidth]{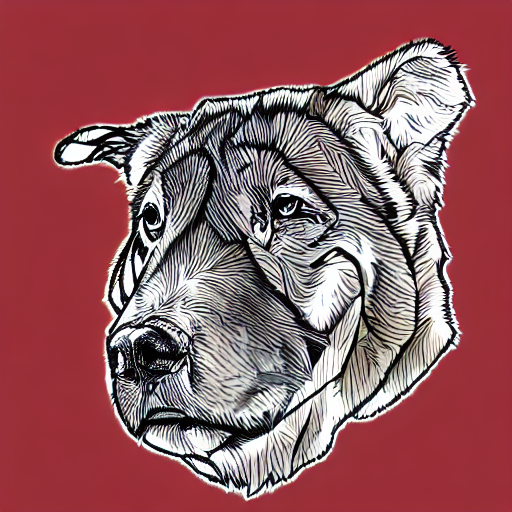} &
        \includegraphics[width=0.075\textwidth]{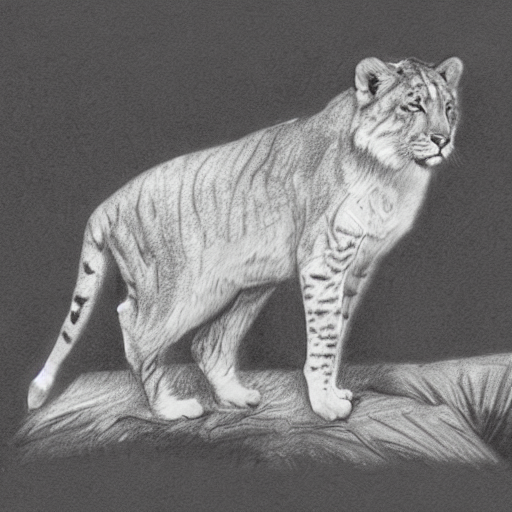} &
        \includegraphics[width=0.075\textwidth]{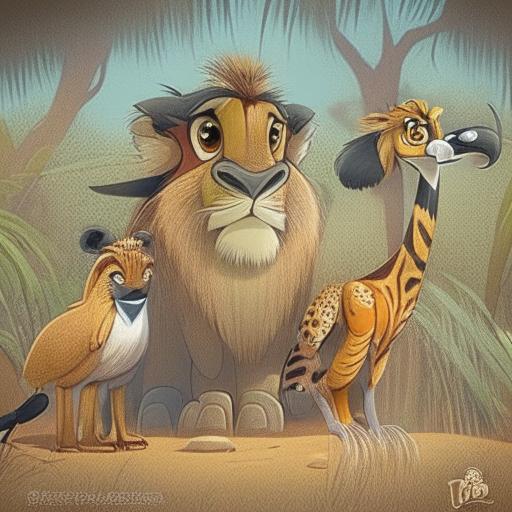} &
        \includegraphics[width=0.075\textwidth]{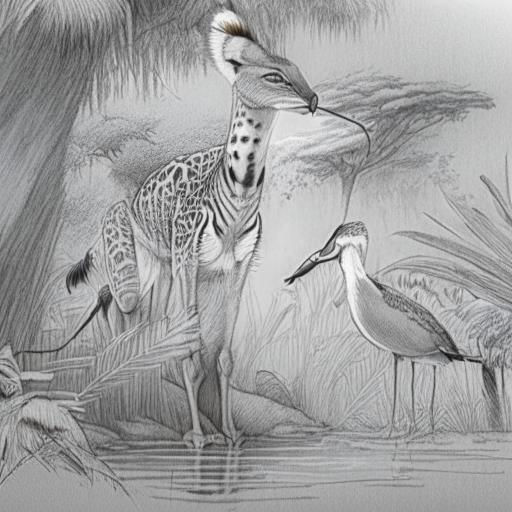} &
        \includegraphics[width=0.075\textwidth]{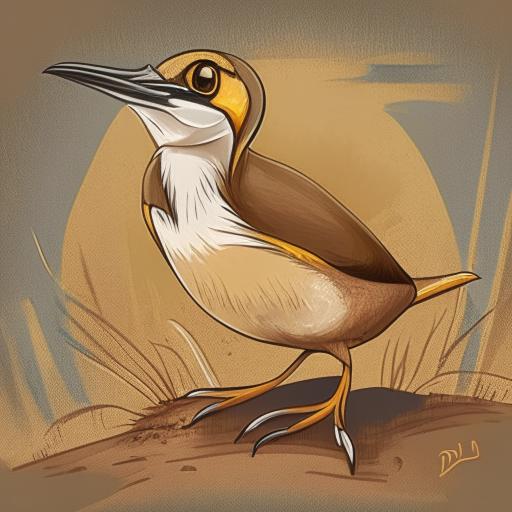} &
        \includegraphics[width=0.075\textwidth]{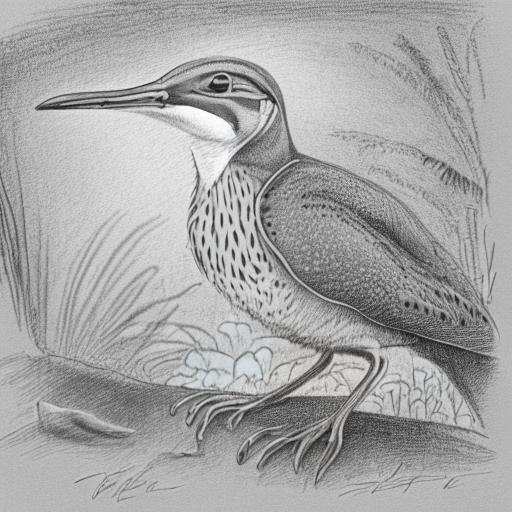} \\
        \multicolumn{6}{c}{{\begin{tabular}{c} \textcolor{mygreen}{+ safari animal}, \textcolor{red}{\textendash \mbox{} elephant, giraffe, lion, rhino, zebra}  \end{tabular}}} \\

        \includegraphics[width=0.075\textwidth]{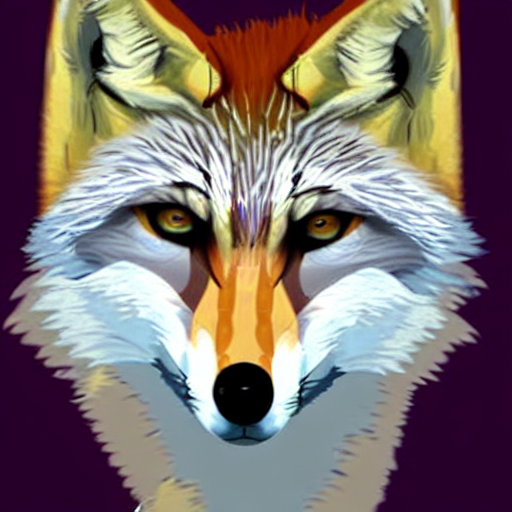} &
        \includegraphics[width=0.075\textwidth]{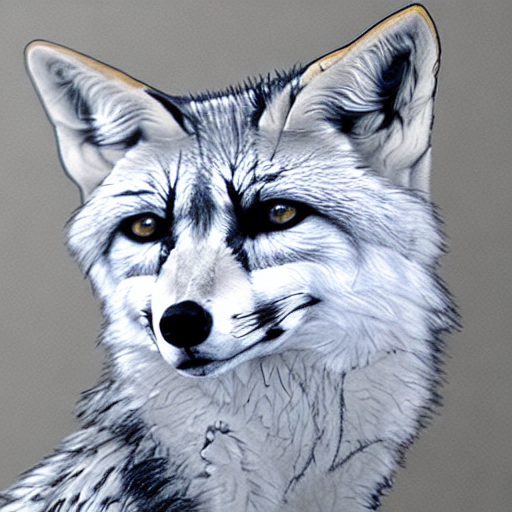} &
        \includegraphics[width=0.075\textwidth]{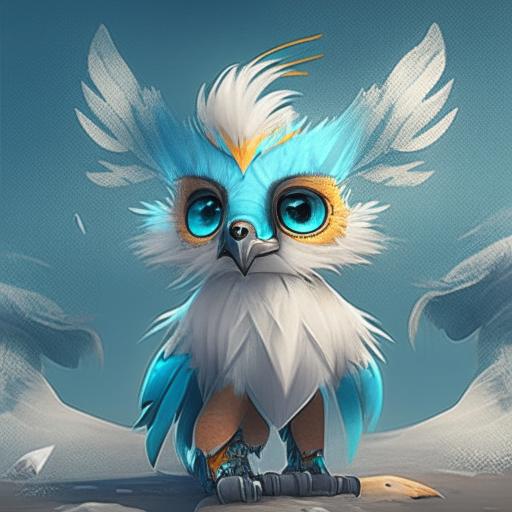} &
        \includegraphics[width=0.075\textwidth]{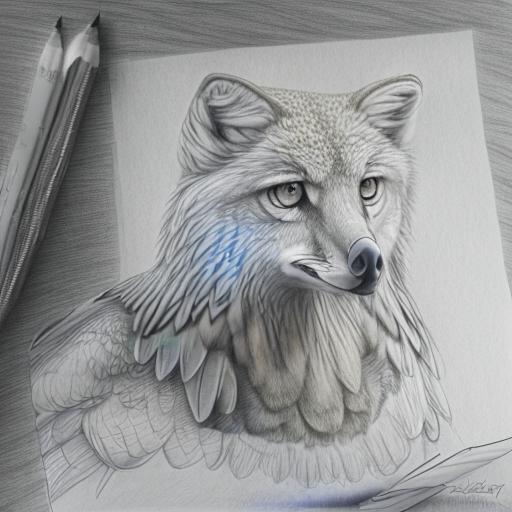} &
        \includegraphics[width=0.075\textwidth]{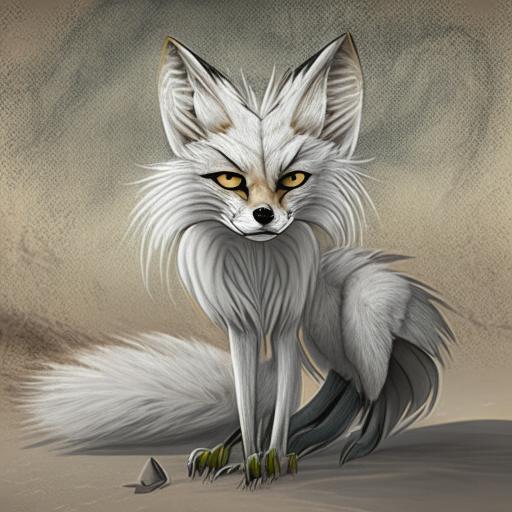} &
        \includegraphics[width=0.075\textwidth]{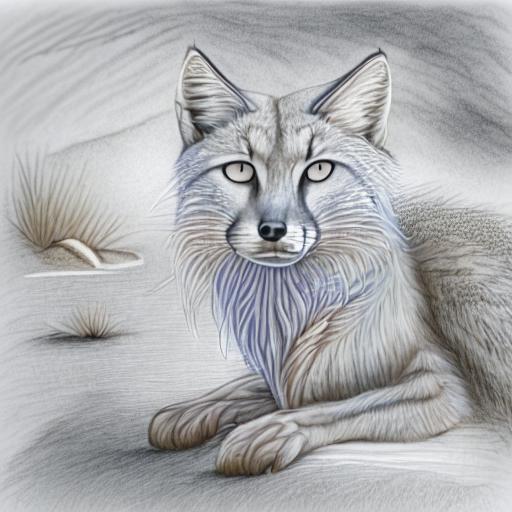} \\
        \multicolumn{6}{c}{\begin{tabular}{c} \textcolor{mygreen}{+ arctic animal}, \textcolor{red}{\textendash \mbox{} polar bear, narwhal, penguin, reindeer}  \end{tabular}} \\  \\[-0.2cm]

        \multicolumn{2}{c}{{\begin{tabular}{c} SD-ConceptLab  \end{tabular}}} &
        \multicolumn{2}{c}{{\begin{tabular}{c}  CLIP-ConceptLab \end{tabular}}} &
        \multicolumn{2}{c}{{\begin{tabular}{c} \textbf{ConceptLab}  \end{tabular}}} 

    \end{tabular}
    
    }
    \caption{Ablation of applying our constraints in the prior space. For SD-ConceptLab we apply constraints over estimated denoised images. For CLIP-ConceptLab we apply the constraints directly on the text encoder output and only use the prior to generate the final images. To highlight our improved consistency, each concept is presented under two prompts: ``A digital cartoon art of ...'' on the right, and ``A pencil sketch of ...'' on the left.}
    \label{fig:ablations_sd_clip_loss}
\end{figure}

The results are illustrated in~\Cref{fig:ablations_sd_clip_loss}. As can be seen, SD-ConceptLab often fails to align with the constraints, as shown in the first two rows, or generates inconsistent images between different prompts featuring the same learned token.
While CLIP-ConceptLab usually does a surprisingly good job at respecting the constraints, it tends to be more inconsistent between different prompts. This aligns well with our insight that applying the Diffusion Prior over $E_y(v_*)$ encourages the generated instances of $v_*$ to better uphold the textual constraints.

\begin{figure}
    \centering
    \setlength{\tabcolsep}{1pt}
    \renewcommand{\arraystretch}{0.5}
    {\small
    \begin{tabular}{c c@{\hspace{0.1cm}} c c@{\hspace{0.1cm}} c c@{\hspace{0.1cm}} c c@{\hspace{0.1cm}} c c@{\hspace{0.1cm}}}

        \includegraphics[width=0.09\textwidth]{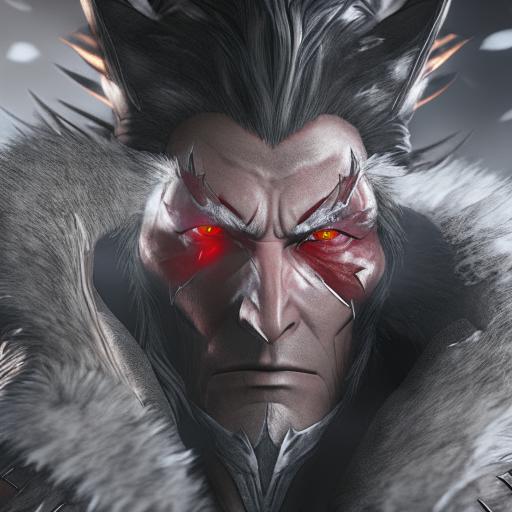} &
        \includegraphics[width=0.09\textwidth]{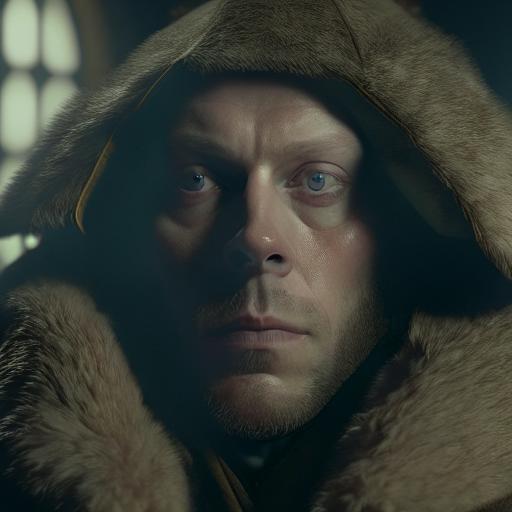} &
        \includegraphics[width=0.09\textwidth]{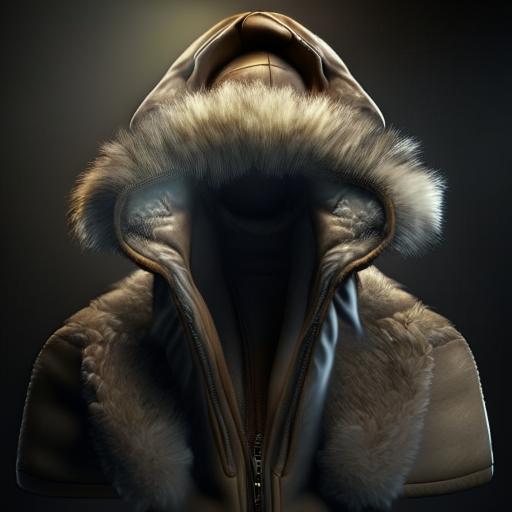} &
        \includegraphics[width=0.09\textwidth]{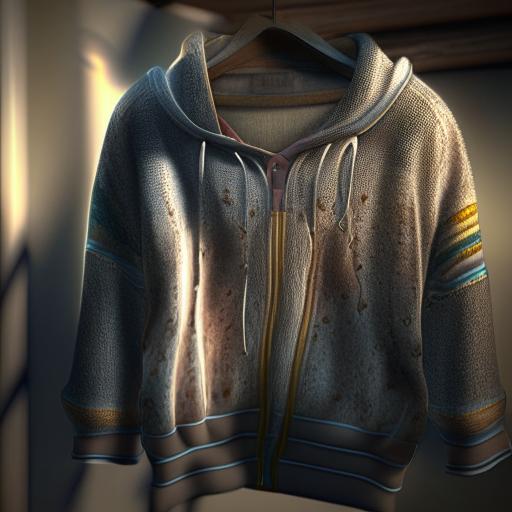} &
        \includegraphics[width=0.09\textwidth]{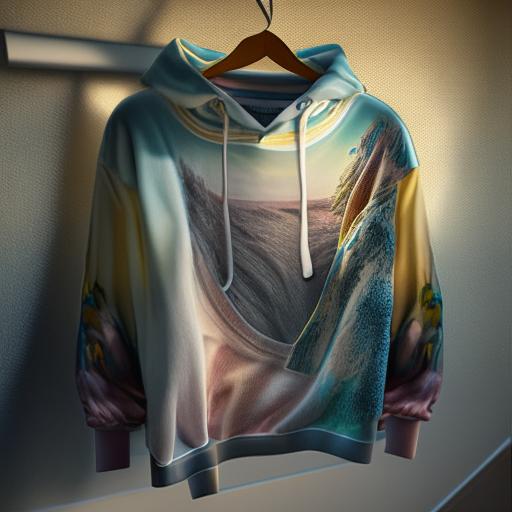} &
        \\
        \multicolumn{6}{c}{{\begin{tabular}{c} \textcolor{mygreen}{+ garment}, \textcolor{red}{\textendash \mbox{} shirt, dress, pants, skirt}  \end{tabular}}} \\ \\
    
        \includegraphics[width=0.09\textwidth]{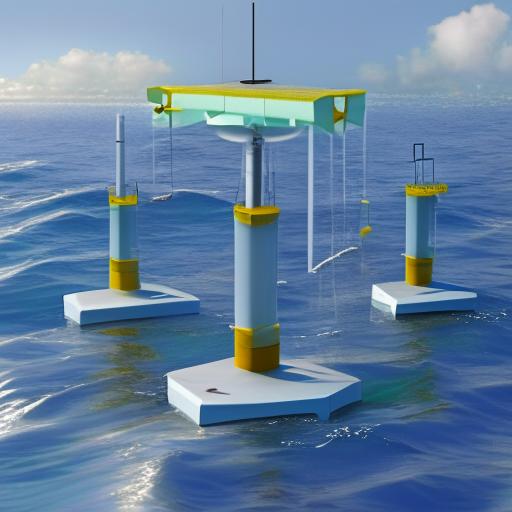} &
        \includegraphics[width=0.09\textwidth]{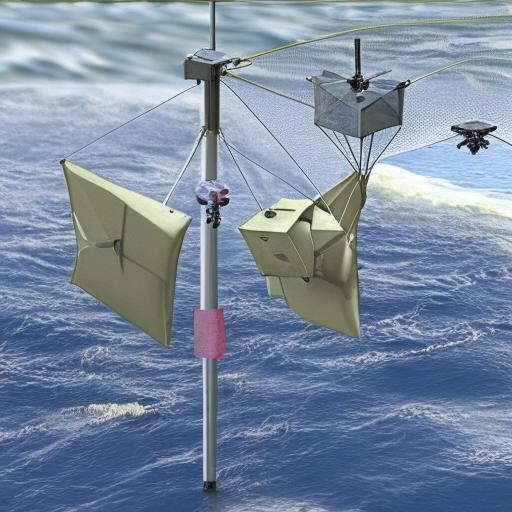} &
        \includegraphics[width=0.09\textwidth]{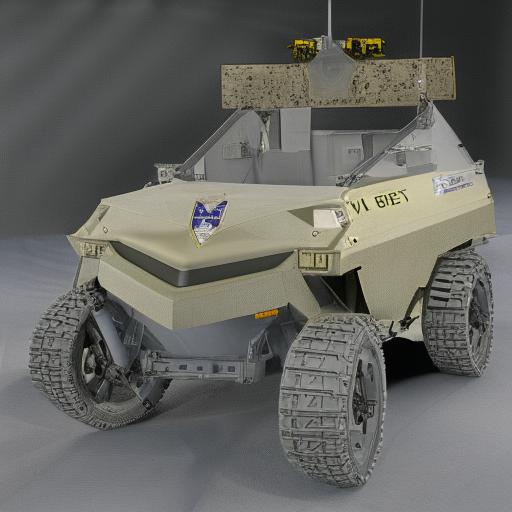} &
        \includegraphics[width=0.09\textwidth]{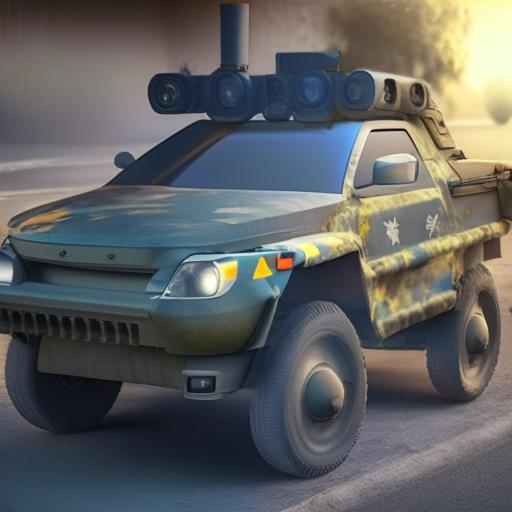} &
        \includegraphics[width=0.09\textwidth]{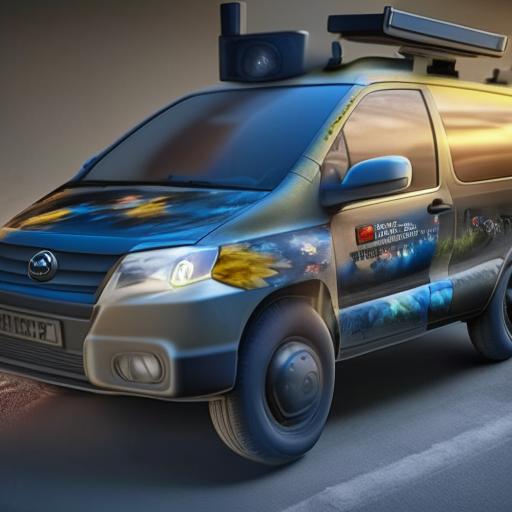} &
        \\
        \multicolumn{6}{c}{{\begin{tabular}{c} \textcolor{mygreen}{+ vehicle}, \textcolor{red}{\textendash \mbox{} car, truck, motorcycle, bus, minibus}  \end{tabular}}} \\ \\

        \multicolumn{1}{c}{{\begin{tabular}{c} $\lambda=0.1$ \end{tabular}}} &
        \multicolumn{1}{c}{{\begin{tabular}{c}  $\lambda=0.5$ \end{tabular}}} &
        \multicolumn{1}{c}{{\begin{tabular}{c}  $\lambda=1.0$ \end{tabular}}} &
        \multicolumn{1}{c}{{\begin{tabular}{c}  $\lambda=2.0$  \end{tabular}}} &
        \multicolumn{1}{c}{{\begin{tabular}{c}  $\lambda=10.0$ \end{tabular}}} 

    \end{tabular}

    }
    \caption{The effect of the relative weighting of our loss between the positive and negative constraints. For small values of $\lambda$ (i.e., low positive weight), the positive constraint is ignored, while for large weights, the negative constraints are largely ignored.}
    \label{fig:experiments_pos_weight}
\end{figure}

\paragraph{Balancing the Constraints. }
In~\Cref{fig:experiments_pos_weight}, we explore the effect of the weighting between the positive and negative constraints as defined in~\Cref{eq:objective}. As shown, when a low weight is given to the positive similarity, the resulting images do not align with the target positive category. Conversely, when the weight is too large, the negative constraints are generally ignored, and the resulting images depict existing concepts found in the list of negative concepts. We find that setting $\lambda=1$  nicely balances both constraints.

\begin{figure}
    \centering
    \setlength{\tabcolsep}{1pt}
    \renewcommand{\arraystretch}{0.7}
    {\small
    \begin{tabular}{c c c c c c}

        \raisebox{0.15in}{\rotatebox{90}{Ours}} &
        \includegraphics[width=0.085\textwidth]{figures/creative_results/sealrat/photo_of_rat_1.jpg} &
        \includegraphics[width=0.085\textwidth]{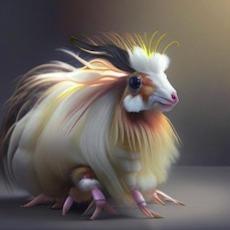} &
        \includegraphics[width=0.085\textwidth]{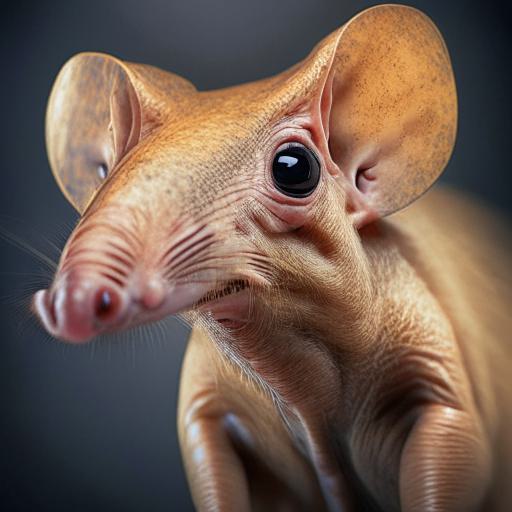} &
        \includegraphics[width=0.085\textwidth,height=0.085\textwidth]{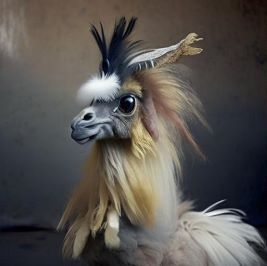} &
        \includegraphics[width=0.085\textwidth]{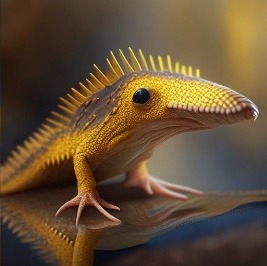} \\

        \raisebox{0.0in}{\rotatebox{90}{\begin{tabular}{c} Prompt \& \\ Kandinsky \end{tabular}}} &
        \includegraphics[width=0.085\textwidth]{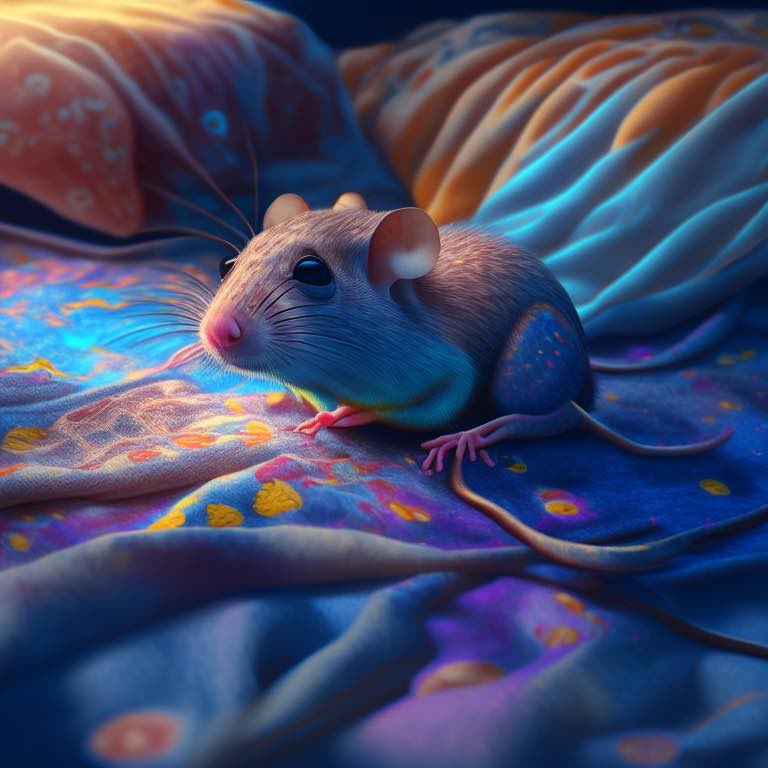} &
        \includegraphics[width=0.085\textwidth]{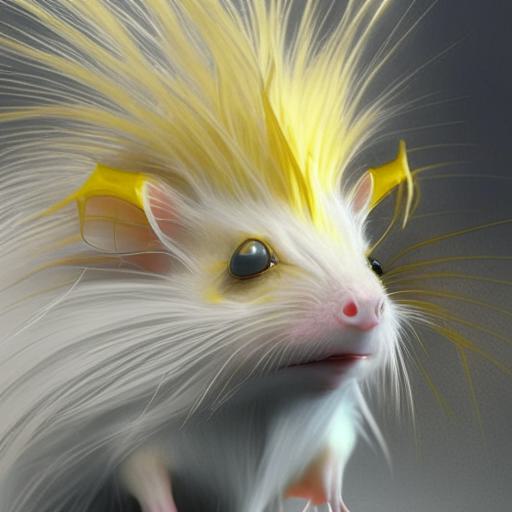} &
        \includegraphics[width=0.085\textwidth]{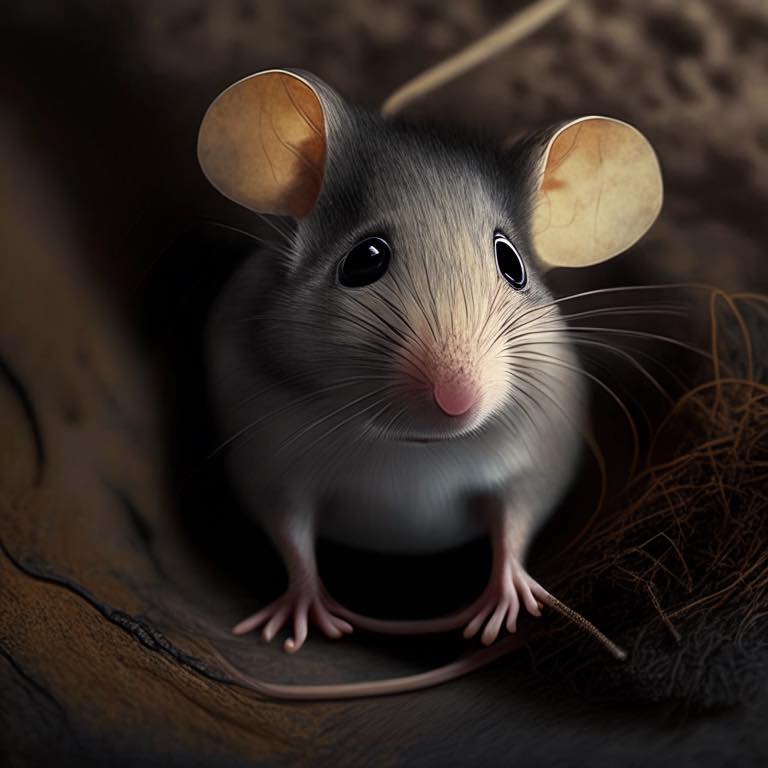} &
        \includegraphics[width=0.085\textwidth]{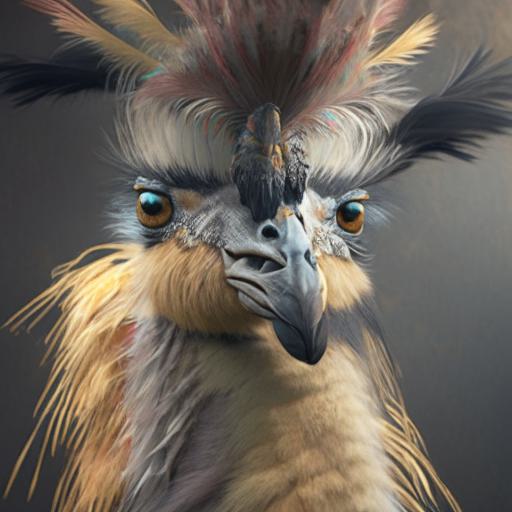} &
        \includegraphics[width=0.085\textwidth]{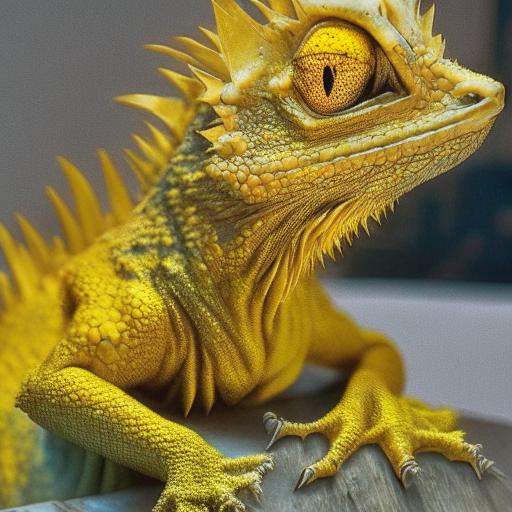} \\

    \end{tabular}
    }
    \caption{Attempting to generate our novel generations with Kandinsky 2~\cite{kandinsky2}. Given an image generated by our method, we use CLIP Interrogator~\cite{clip_interrogator} to compose a prompt describing our concept, which is then used to generate an image. For example, the prompt for the rightmost image is: \textit{``a close up of a lizard on a table, inspired by Bob Eggleton, zbrush central contest winner, yellow spiky hair, photoreal, vivid colours. sharp focus. wow!, realistic gold, great pinterest photo, beautiful, photo realistic''.}}
    \vspace{-0.4cm}
    \label{fig:prompting_comparison}
\end{figure}

\paragraph{Generated Descriptions}
Once a concept has been generated using ConceptLab, an interesting question arises: can this novel idea now be automatically transformed into a text prompt instead of a token? 
To check this, we first pass an image depicting a learned concept to a vision-language model~\cite{clip_interrogator} and ask it to compose a prompt corresponding to the input image. We then pass the generated prompt to Kandinsky~\cite{kandinsky2} and generate a corresponding image. As can be seen in~\Cref{fig:prompting_comparison}, the generated prompt is able to capture the general nature of the concept, but its unique details are mostly missing.
One can potentially manually refine each prompt to better represent some of the missing properties of our generated concepts, but this only further highlights the unique nature of our generated concepts and the benefit of representing them as learnable tokens.
\begin{figure}
     \centering
    \setlength{\tabcolsep}{0pt}
    \renewcommand{\arraystretch}{0.5}
    {\small
    \begin{tabular}{c c c c c c c}
        \multicolumn{7}{c}{\includegraphics[width=0.45\textwidth]{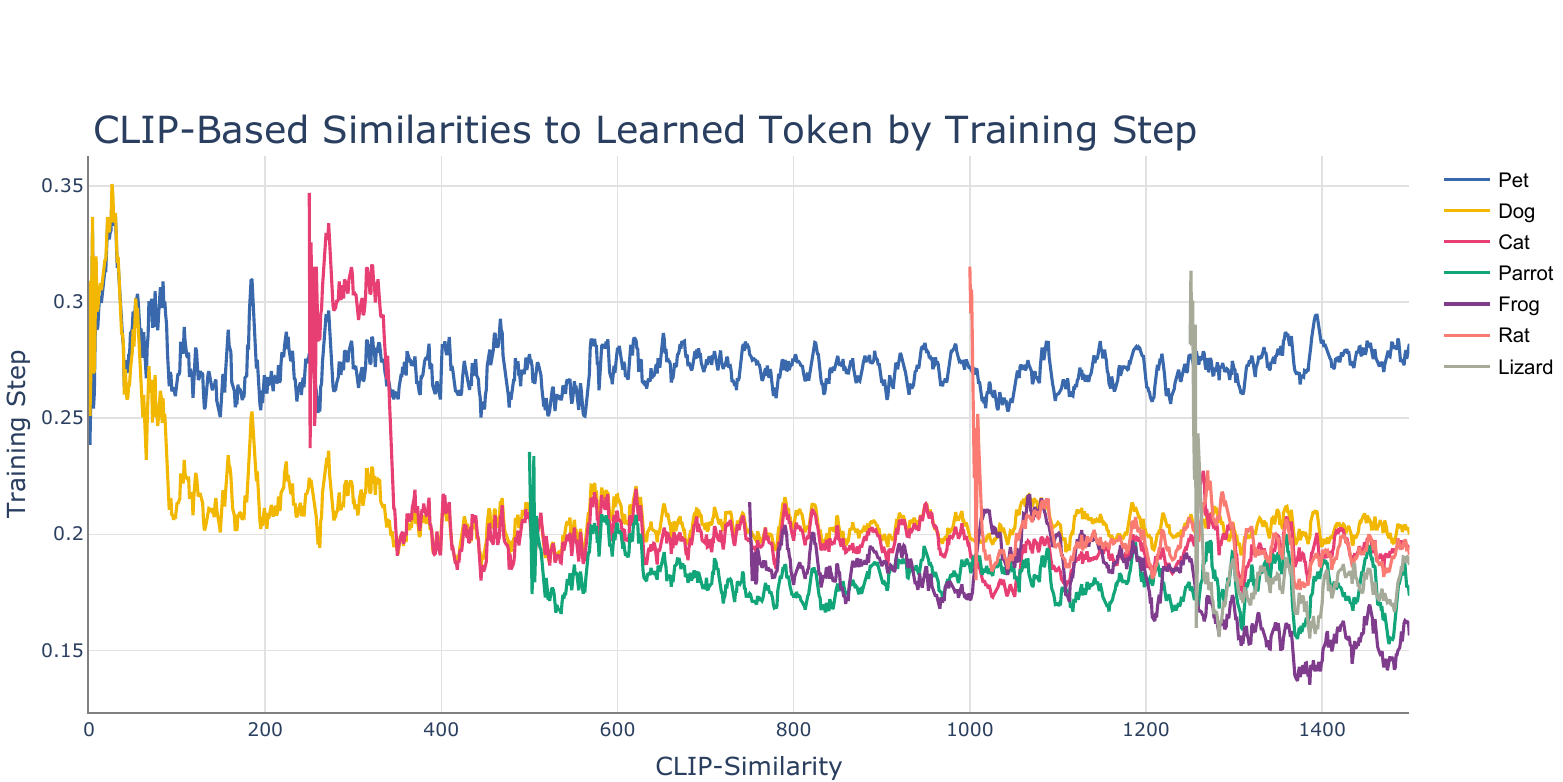}}
        \\
        \includegraphics[width=0.065\textwidth]{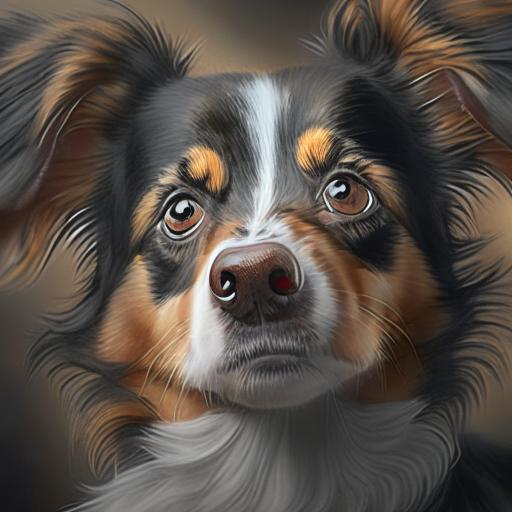}
        &
        \includegraphics[width=0.065\textwidth]{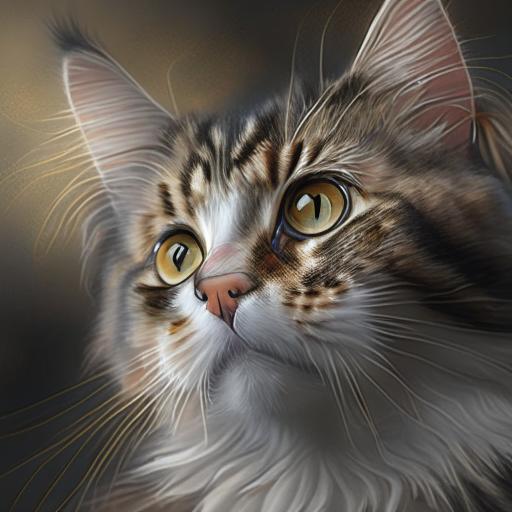}
        &
        \includegraphics[width=0.065\textwidth]{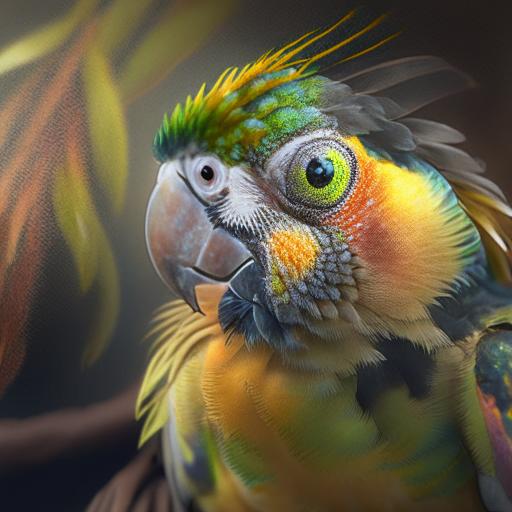}
        &
        \includegraphics[width=0.065\textwidth]{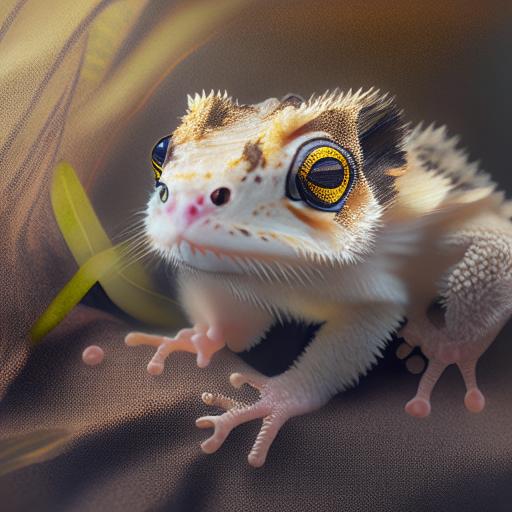}
        &
        \includegraphics[width=0.065\textwidth]{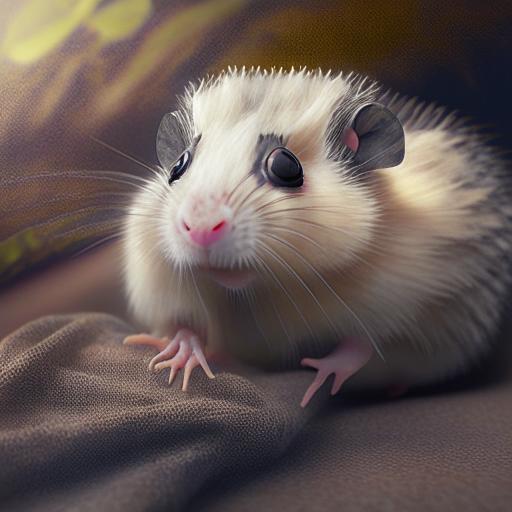}
        &
        \includegraphics[width=0.065\textwidth]{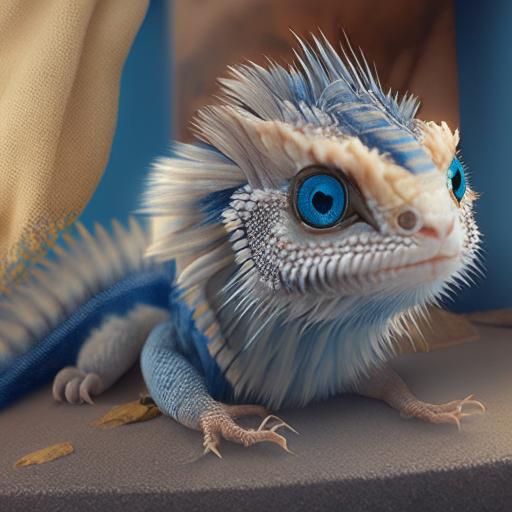}
        &
        \includegraphics[width=0.065\textwidth]{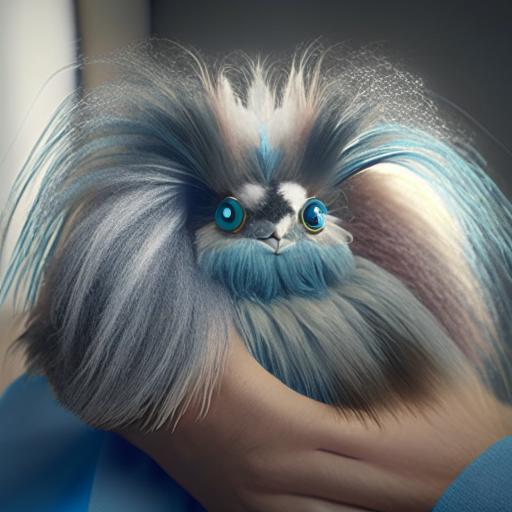} 
        \\
        dog
        &
        cat
        & 
        parrot
        & 
        frog
        & 
        rat
        &
        lizard
        &
        new-pet
    \end{tabular}
    }
    \caption{CLIP-based similarity between our learned concept and the positive and negative constraints throughout training.}
    \label{fig:sim_graph}
\end{figure}

\paragraph{Similarity Analysis}

In \Cref{fig:sim_graph}, we demonstrate how the similarity to different constraints behaves along the optimization process when applying our adaptive negatives scheme. In the upper part of the Figure, we can observe that the similarity to the positive constraint, in this case, ``pet'', remains relatively constant. Every $250$ iterations, a new negative constraint is added based on BLIP-2's predictions, and one can observe how the similarity to the new constraint decreases over time. At the bottom, we present the rendered images from which BLIP-2 inferred the new negative member to add to our list of constraints.

\section{Limitations}
\begin{figure}
    \centering
    \setlength{\tabcolsep}{0pt}
    \renewcommand{\arraystretch}{0.6}
    {\footnotesize
    \begin{tabular}{c c c@{\hspace{0.1cm}} c c}

                \includegraphics[width=0.0915\textwidth]{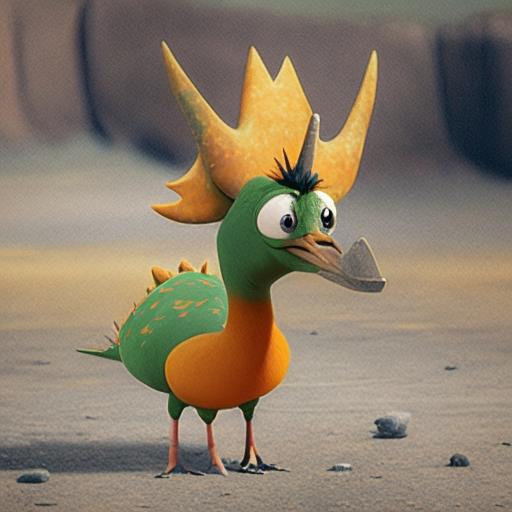} &
        \includegraphics[width=0.0915\textwidth]{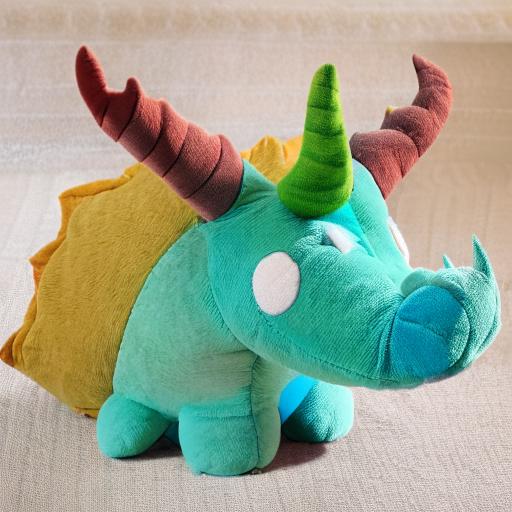} &
        \includegraphics[width=0.0915\textwidth]{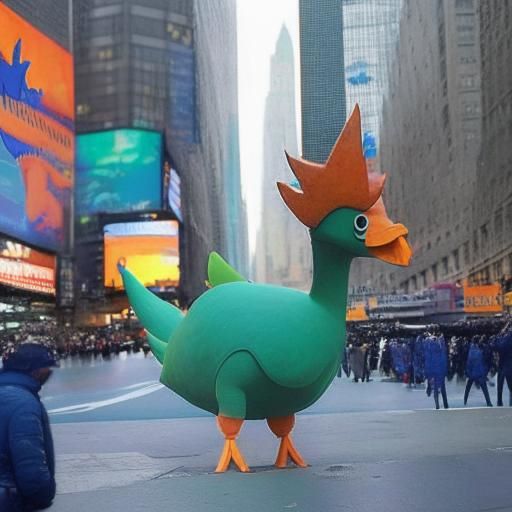} &
        \includegraphics[width=0.0915\textwidth]{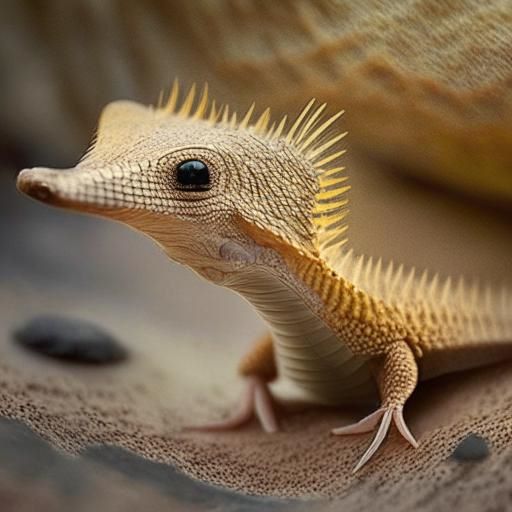} &
        \includegraphics[width=0.0915\textwidth]{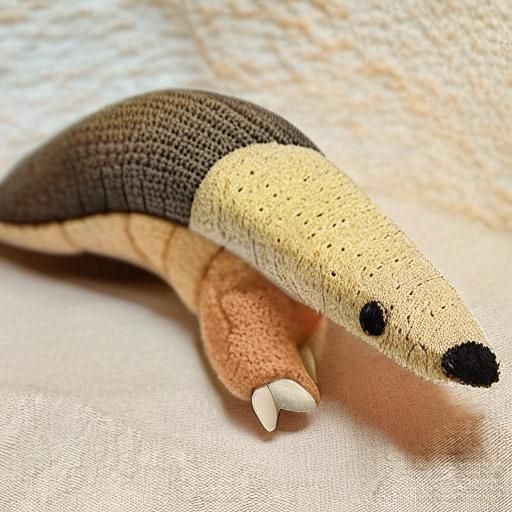} \\

 \textcolor{mygreen}{+ dino} & 
        ``plush'' & 
        {\begin{tabular}{c} ``in Times \\ Square''\end{tabular}} & 
        \textcolor{mygreen}{+ reptile} & 
        {\begin{tabular}{c} ``plush''\end{tabular}} \\

        \includegraphics[width=0.0915\textwidth,height=0.0915\textwidth]{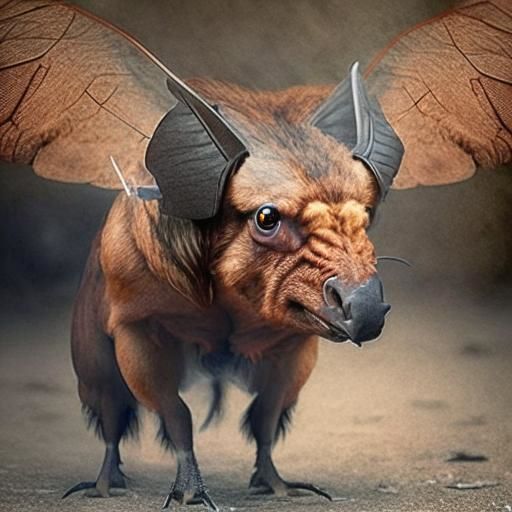} &
        \includegraphics[width=0.0915\textwidth]{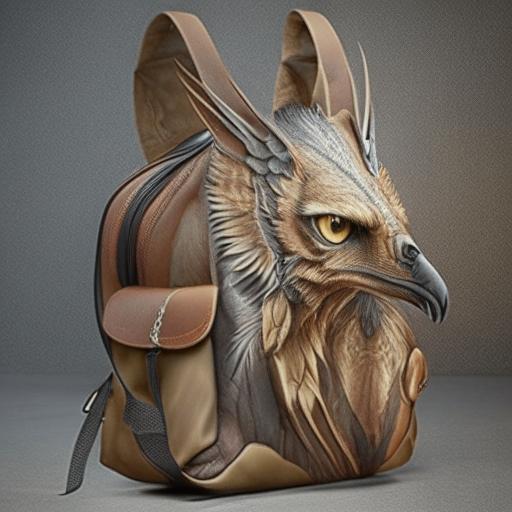} &
        \includegraphics[width=0.0915\textwidth]{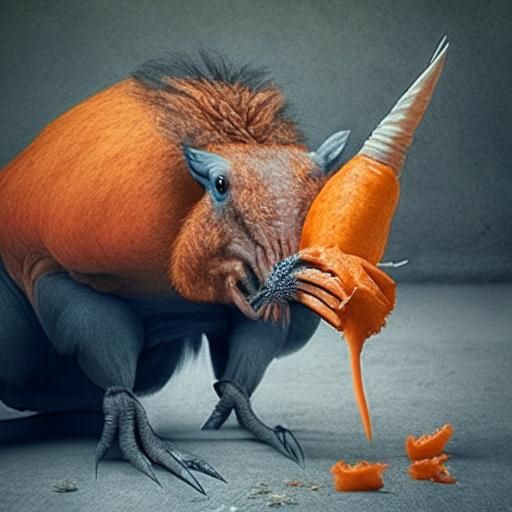} &
        \includegraphics[width=0.0915\textwidth]{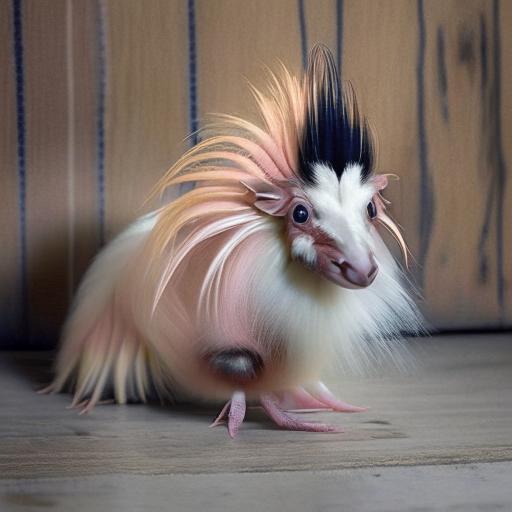} &
        \includegraphics[width=0.0915\textwidth]{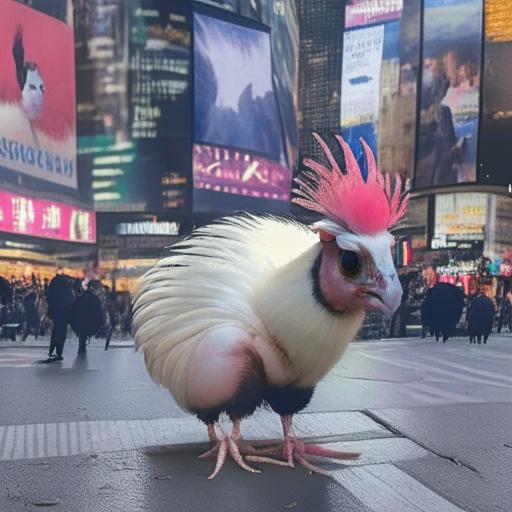} \\

       \textcolor{mygreen}{+ beast} & 
        {\begin{tabular}{c} ``a backpack''\end{tabular}} & 
        {\begin{tabular}{c} ``eating \\ a carrot''\end{tabular}} &
        \textcolor{mygreen}{+ pet} & 
        {\begin{tabular}{c} ``in Times \\ Square''\end{tabular}} \\

        \includegraphics[width=0.0915\textwidth]{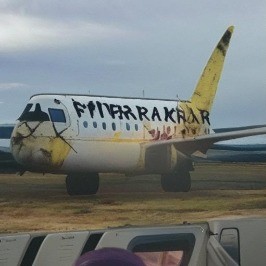} &
        \includegraphics[width=0.0915\textwidth]{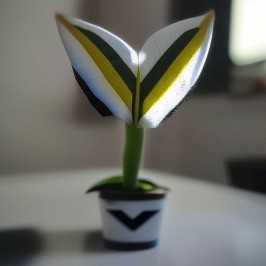} &
        \includegraphics[width=0.0915\textwidth]{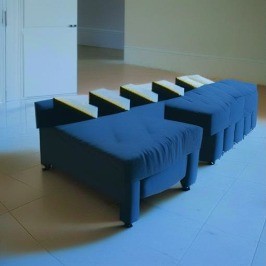} &
        \includegraphics[width=0.0915\textwidth]{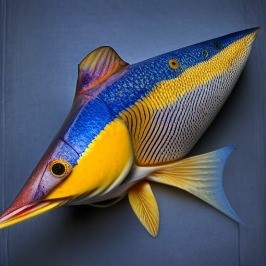} &
        \includegraphics[width=0.0915\textwidth]{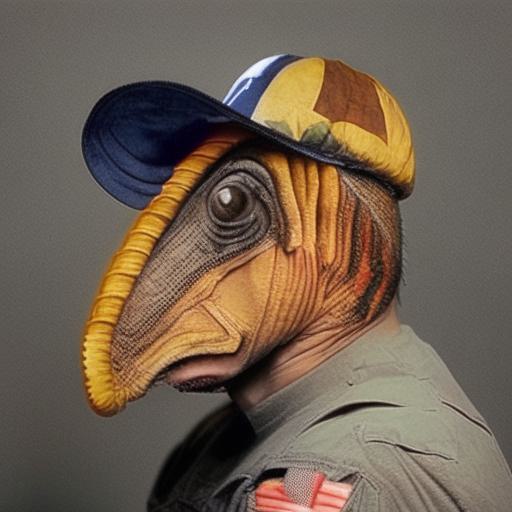} \\

        \textcolor{mygreen}{+ airplane} & 
        \textcolor{mygreen}{+ plant} & 
        \textcolor{mygreen}{+ furniture} & 
       \textcolor{mygreen}{+ fish} & 
        \textcolor{mygreen}{+ reptile}   \\

    \\[-0.3cm]
    \end{tabular}
    }
    \caption{Limitations of ConceptLab. Some edits do not respect all of the concept properties, resulting in more generic outputs. Some learned concepts are not creative or do not respect the positive constraint well enough.}
    \label{fig:limitations}
\end{figure}

Our method is generally capable of learning novel concepts that follow the given constraints. However, it is important to acknowledge  its limitations. First, similar to personalization methods, creating new images with different prompts that include the learned concept does not always preserve the concept's properties. We illustrate such examples in the first two rows of~\Cref{fig:limitations}. Second, the optimization process itself does not always yield the desired outcomes. For some classes, such as ``airplane'' or ``fish'', ConceptLab struggles to generate creative concepts. We empirically  observe that this is often related to negatives generated by BLIP-2. For instance, in some categories, BLIP-2 tends to produce highly specific negatives (e.g., a particular airplane model) that do not serve as a strong constraint.

\section{Conclusions}
We introduced a novel approach for creative generation using text-to-image diffusion models. Specifically, we proposed to use Diffusion Prior models to learn novel concepts that belong to a given broad category. To optimize our learned concept we introduced ``prior constraints'', a set of positive and negative constraints applied over the Diffusion Prior output. By integrating a question-answering VLM into the optimization process we encouraged uniqueness while ensuring distinctness from existing category members. Our experiments demonstrate the effectiveness of our method, producing visually diverse and appealing concepts, and further showcasing the effectiveness of ``prior constraints'' for concept mixing. We hope that our approach will open up exciting possibilities for generating creative content using text-to-image models.

\begin{acks}
We would like to give a special thanks to Hao Zhang for inspiring and encouraging us throughout this work. We would also like to thank Gal Metzer and Rinon Gal for their valuable feedback and suggestions. This work was supported by the Israel Science Foundation under Grant No. 2366/16 and Grant No. 2492/20.
\end{acks}

\bibliographystyle{ACM-Reference-Format}
\bibliography{main}

\appendix

\begin{figure*}
    \centering
    \setlength{\tabcolsep}{0.1pt}
    {\footnotesize
    \begin{tabular}{c c@{\hspace{0.25cm}} c@{\hspace{0.25cm}} c@{\hspace{0.25cm}} c@{\hspace{0.25cm}} c@{\hspace{0.25cm}} c@{\hspace{0.25cm}}}

        \raisebox{0.65in}{\multirow{2}{*}{
            \rotatebox{90}{\begin{tabular}{c} \textcolor{mygreen}{+ super hero}  \end{tabular}}
        }} &
        \includegraphics[width=0.15\textwidth]{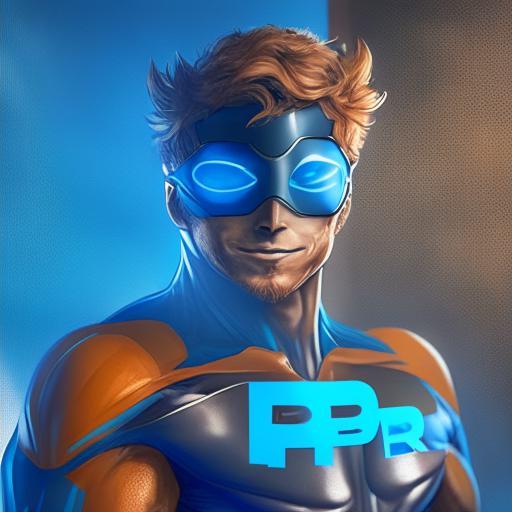}
        \includegraphics[width=0.15\textwidth]{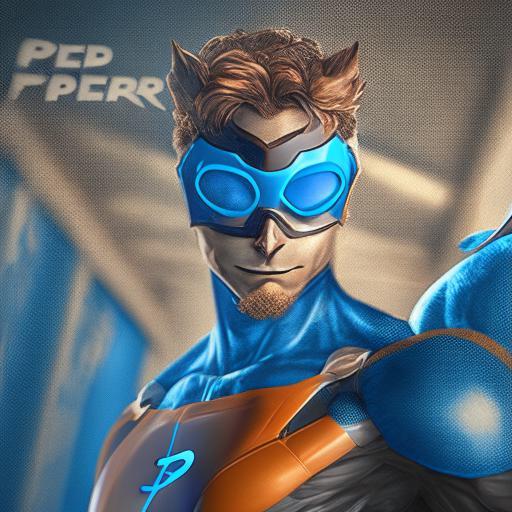} &
        \includegraphics[width=0.15\textwidth]{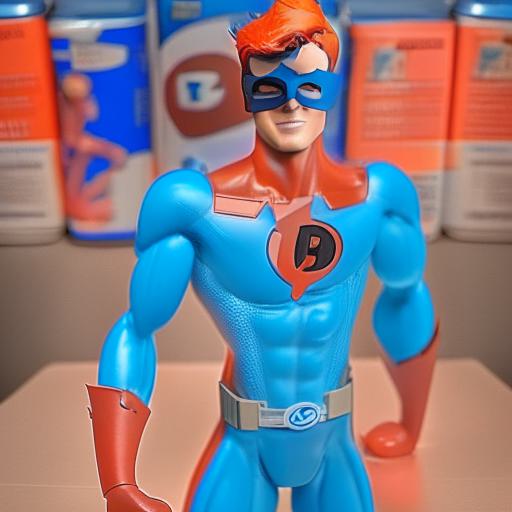} &
        \includegraphics[width=0.15\textwidth]{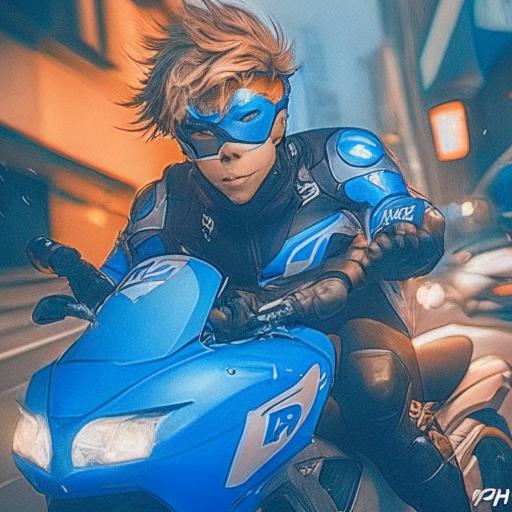} &
        \includegraphics[width=0.15\textwidth]{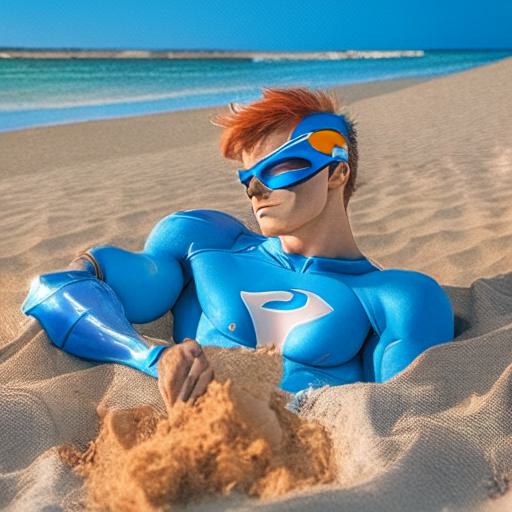} &
        \includegraphics[width=0.15\textwidth]{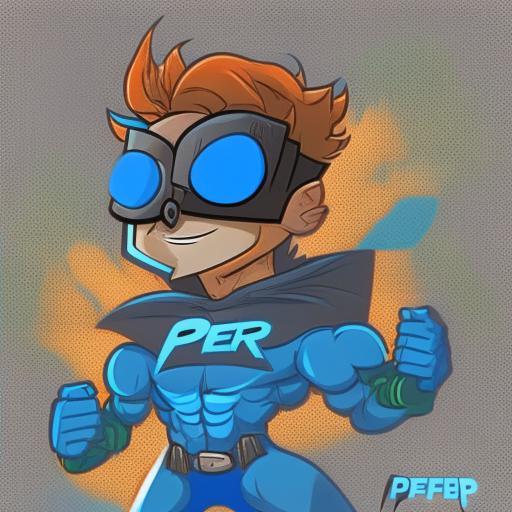} \\

        &
        \begin{tabular}{c} ``Professional high-quality photo of \\ a $S_*$. photorealistic, 4k, HQ'' \end{tabular} &
        \begin{tabular}{c} ``An action figure of $S_*$'' \end{tabular} &
        \begin{tabular}{c} ``$S_*$ riding his \\ motorcycle'' \end{tabular} &
        \begin{tabular}{c} ``A photo of $S_*$ \\  relaxing on the beach'' \end{tabular} &
        \begin{tabular}{c} ``A cartoon of $S_*$'' \end{tabular} \\

        \raisebox{0.65in}{\multirow{2}{*}{
            \rotatebox{90}{\begin{tabular}{c} \textcolor{mygreen}{+ building}  \end{tabular}}
        }} &
        \includegraphics[width=0.15\textwidth]{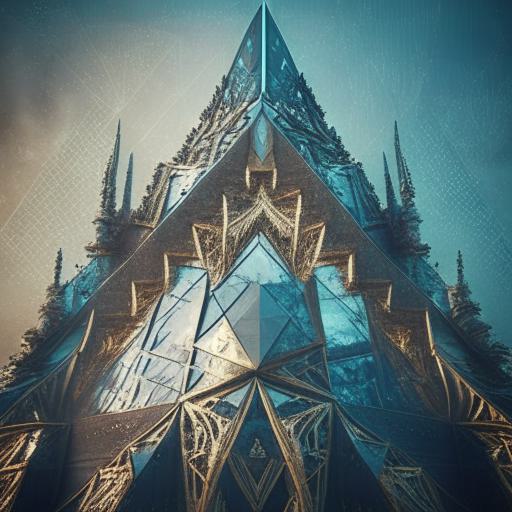}
        \includegraphics[width=0.15\textwidth]{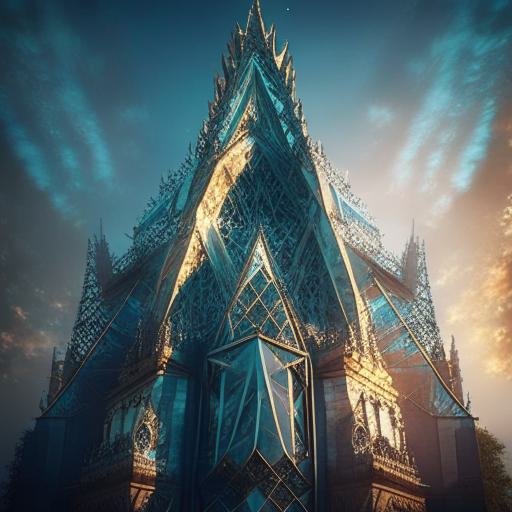} &
        \includegraphics[width=0.15\textwidth]{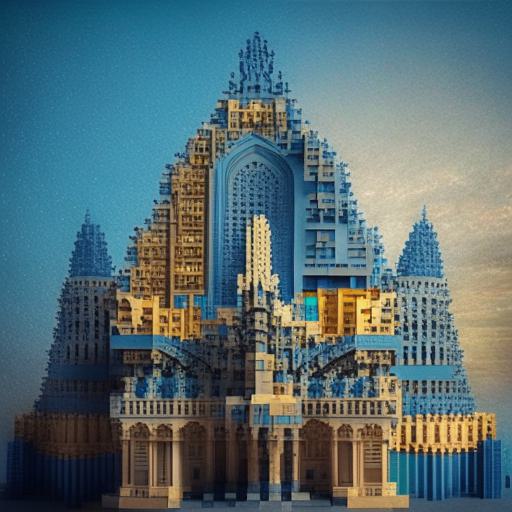} &
        \includegraphics[width=0.15\textwidth]{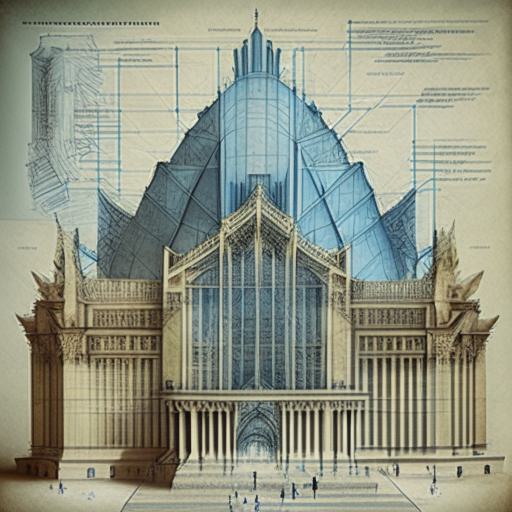} &
        \includegraphics[width=0.15\textwidth]{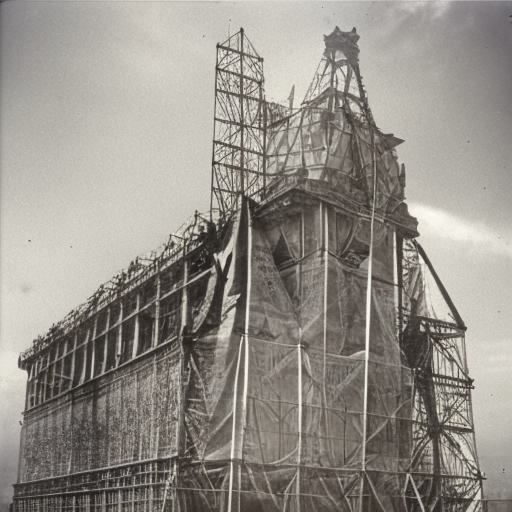} &
        \includegraphics[width=0.15\textwidth]{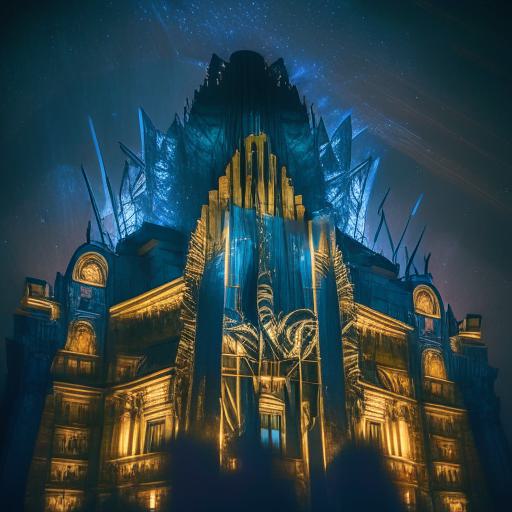} \\

        &
        \begin{tabular}{c} ``Professional high-quality photo of \\ a $S_*$. photorealistic, 4k, HQ'' \end{tabular} &
        \begin{tabular}{c} ``A photo of a $S_*$ \\ made out of legos'' \end{tabular} &
        \begin{tabular}{c} ``An architectural \\ blueprint of $S_*$'' \end{tabular} &
        \begin{tabular}{c} ``A photo of $S_*$ \\ during construction'' \end{tabular} &
        \begin{tabular}{c} ``A photo of $S_*$'' \\ glowing at night \end{tabular} \\

        \raisebox{0.55in}{\multirow{2}{*}{
            \rotatebox{90}{\begin{tabular}{c} \textcolor{mygreen}{+ reptile}  \end{tabular}}
        }} &
        \includegraphics[width=0.15\textwidth]{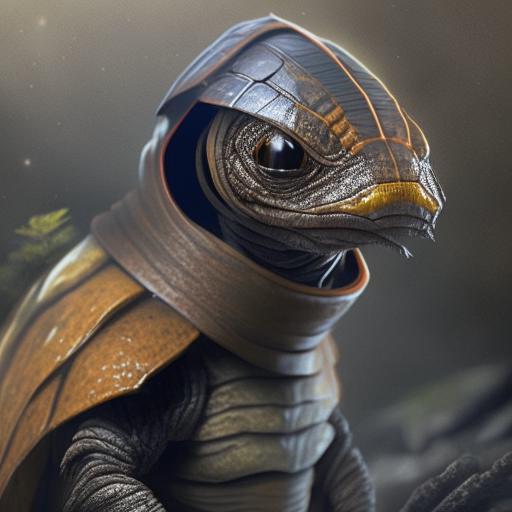}
        \includegraphics[width=0.15\textwidth]{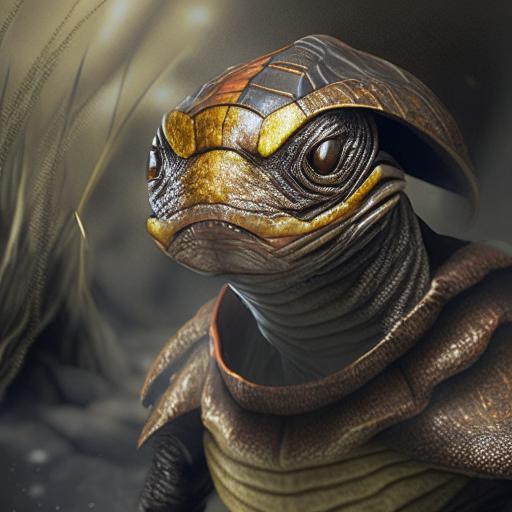} &
        \includegraphics[width=0.15\textwidth]{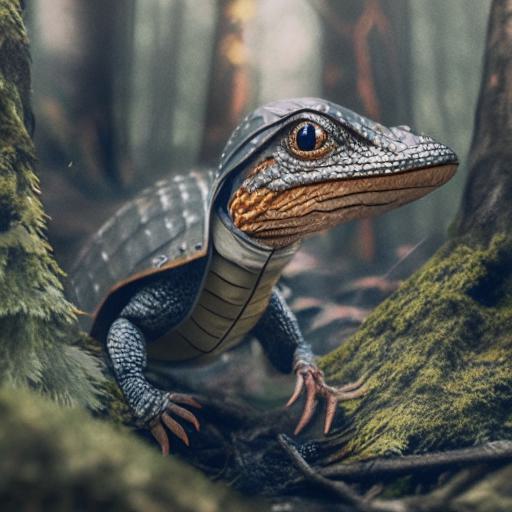} &
        \includegraphics[width=0.15\textwidth]{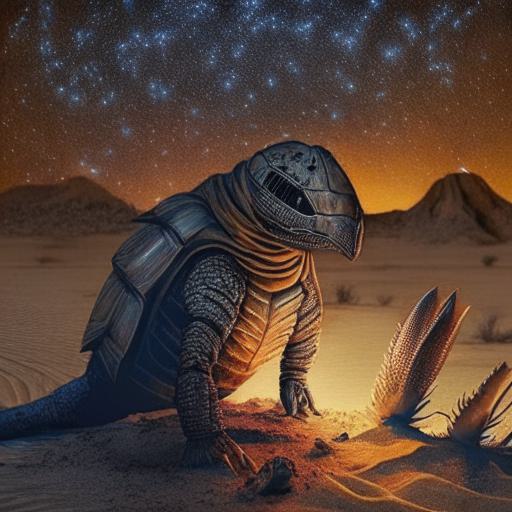} &
        \includegraphics[width=0.15\textwidth]{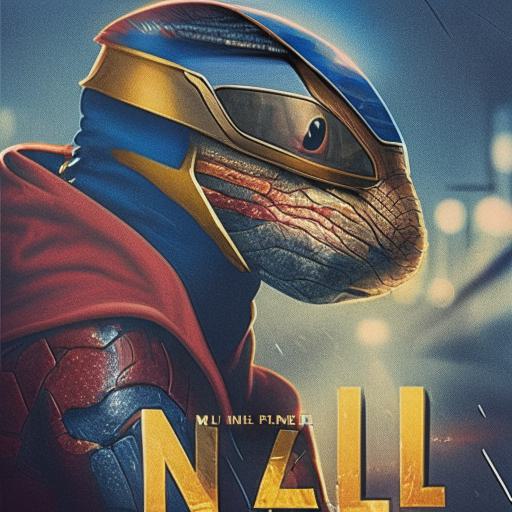} &
        \includegraphics[width=0.15\textwidth]{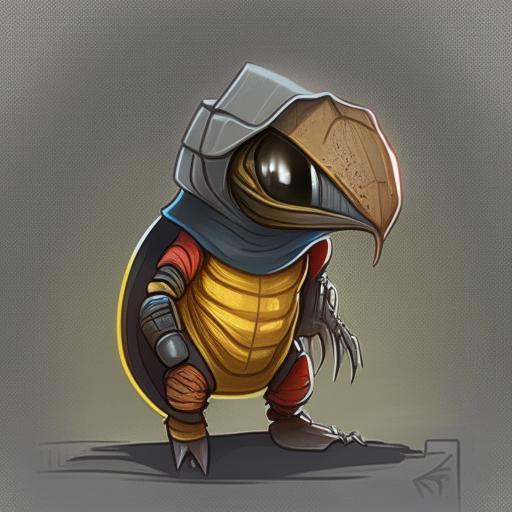} \\ 

        &
        \begin{tabular}{c} ``Professional high-quality photo of \\ a $S_*$. photorealistic, 4k, HQ'' \end{tabular} &
        \begin{tabular}{c} ``A photo of a $S_*$ \\ in the forest'' \end{tabular} &
        \begin{tabular}{c} ``A photo of $S_*$ in \\ the desert at night'' \end{tabular} &
        \begin{tabular}{c} ``A movie poster \\ featuring a $S_*$'' \end{tabular} &
        \begin{tabular}{c} ``A cartoon of a $S_*$'' \end{tabular} \\

        \raisebox{0.5in}{\multirow{2}{*}{
            \rotatebox{90}{\begin{tabular}{c} \textcolor{mygreen}{+ fruit}  \end{tabular}}
        }} &
        \includegraphics[width=0.15\textwidth]{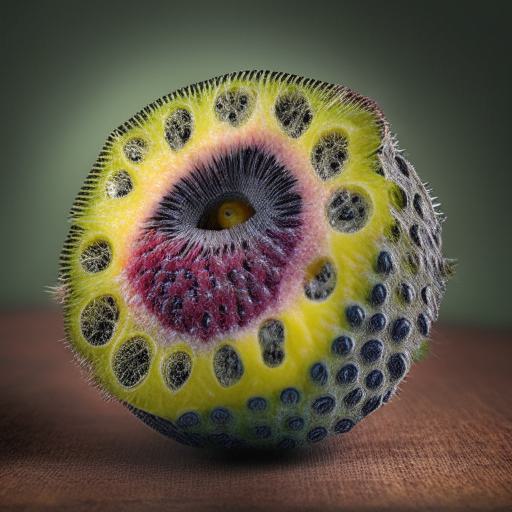}
        \includegraphics[width=0.15\textwidth]{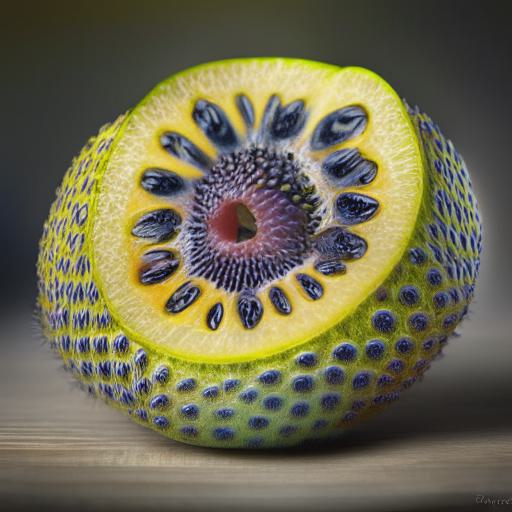} &
        \includegraphics[width=0.15\textwidth]{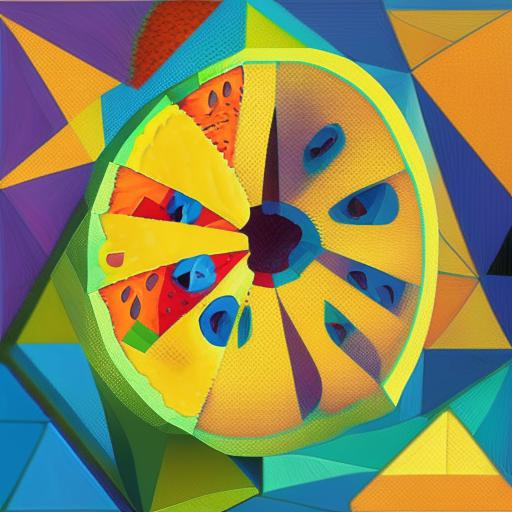} &
        \includegraphics[width=0.15\textwidth]{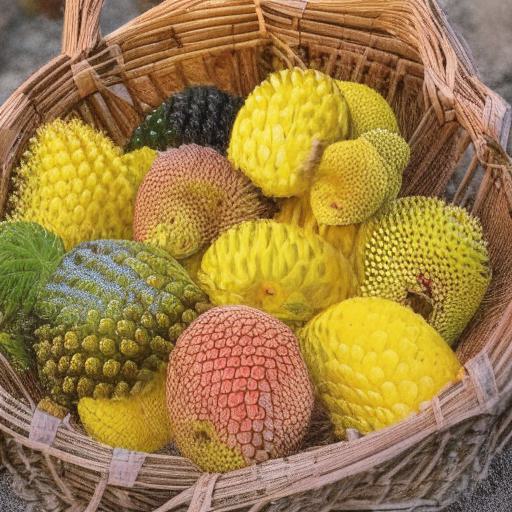} &
        \includegraphics[width=0.15\textwidth]{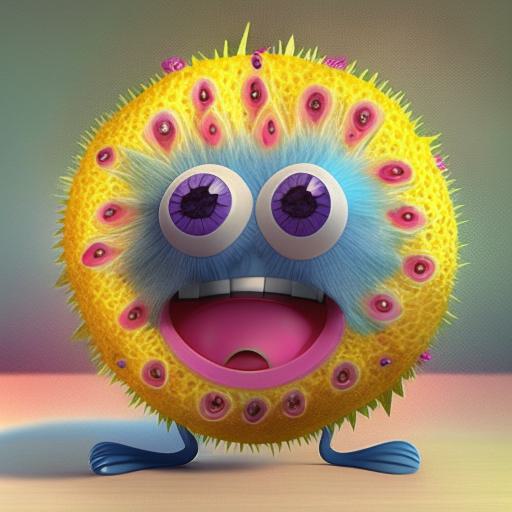} &
        \includegraphics[width=0.15\textwidth]{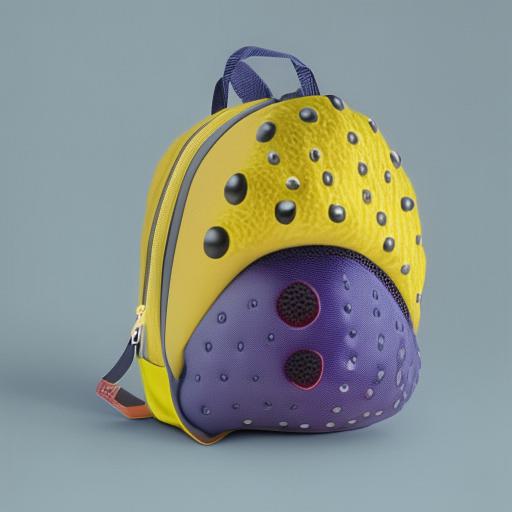} \\

        &
        \begin{tabular}{c} ``Professional high-quality photo of \\ a $S_*$. photorealistic, 4k, HQ'' \end{tabular} &
        \begin{tabular}{c} ``An cubism painting \\ of a $S_*$'' \end{tabular} &
        \begin{tabular}{c} ``A basket of freshly \\ picked $S_*$'' \end{tabular} &
        \begin{tabular}{c} ``An animated character \\ of a $S_*$'' \end{tabular} &
        \begin{tabular}{c} ``A backpack \\ of a $S_*$'' \end{tabular} \\  

        \raisebox{0.45in}{\multirow{2}{*}{
            \rotatebox{90}{\begin{tabular}{c} \textcolor{mygreen}{+ pet}  \end{tabular}}
        }} &
        \includegraphics[width=0.15\textwidth]{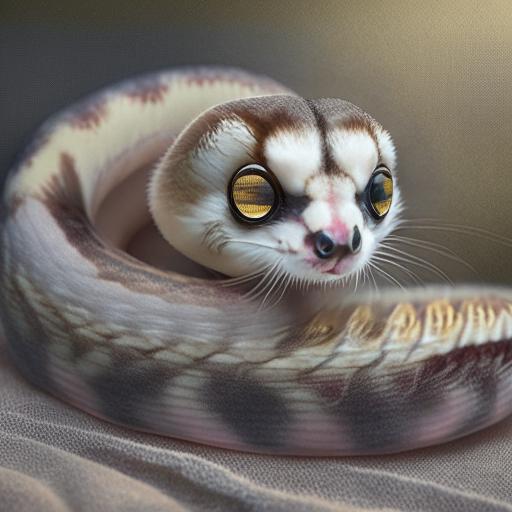}
        \includegraphics[width=0.15\textwidth]{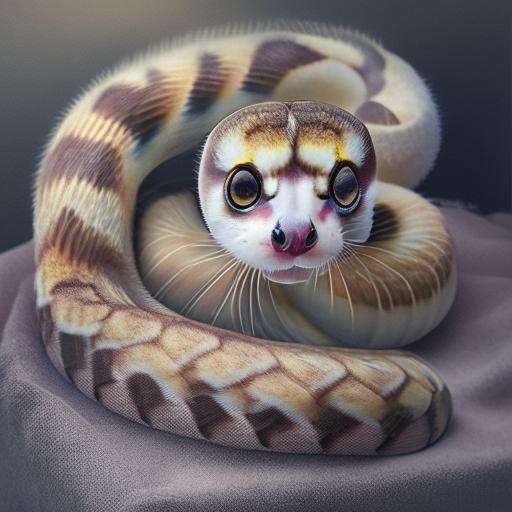} &
        \includegraphics[width=0.15\textwidth]{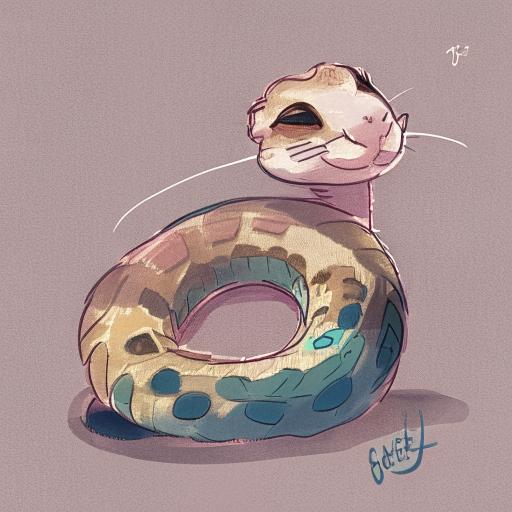} &
        \includegraphics[width=0.15\textwidth]{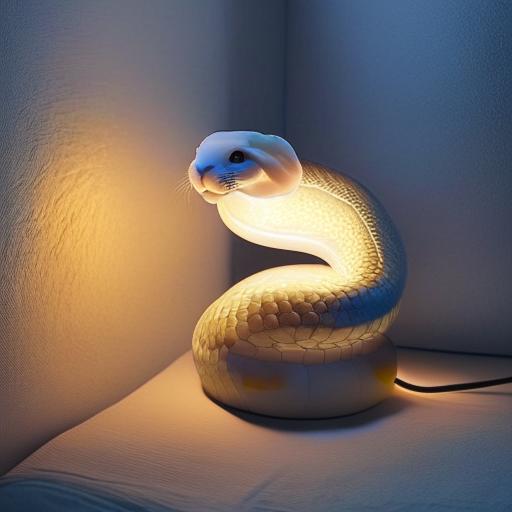} &
        \includegraphics[width=0.15\textwidth]{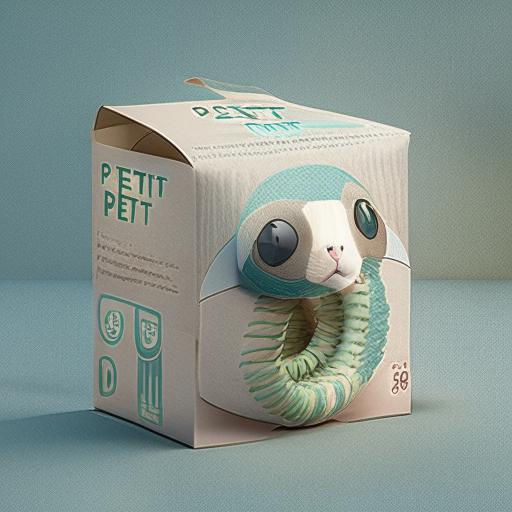} &
        \includegraphics[width=0.15\textwidth]{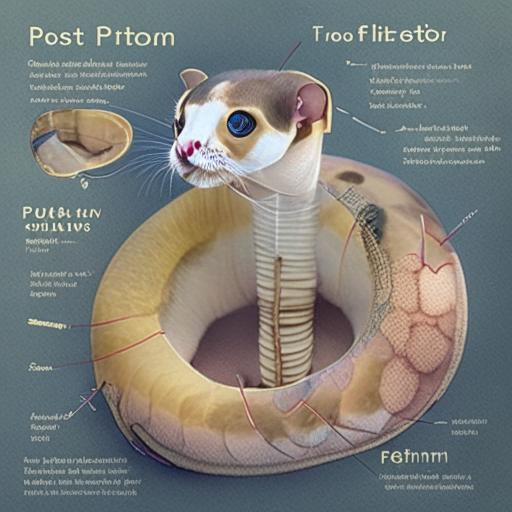} \\ 
        &
        \begin{tabular}{c} ``Professional high-quality photo of \\ a $S_*$. photorealistic, 4k, HQ'' \end{tabular} &
        \begin{tabular}{c} ``A doodle of a $S_*$'' \end{tabular} &
        \begin{tabular}{c} ``A $S_*$ in the  shape \\ of a night lamp'' \end{tabular} &
        \begin{tabular}{c} ``A package design for \\  a  $S_*$ toy'' \end{tabular} &
        \begin{tabular}{c} ``The anatomy of  $S_*$'' \end{tabular} \\ 

        \raisebox{0.55in}{\multirow{2}{*}{
            \rotatebox{90}{\begin{tabular}{c} \textcolor{mygreen}{+ insect}  \end{tabular}}
        }} &
        \includegraphics[width=0.15\textwidth]{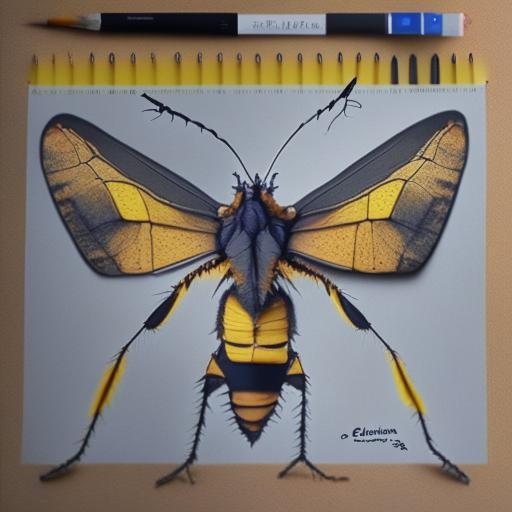}
        \includegraphics[width=0.15\textwidth]{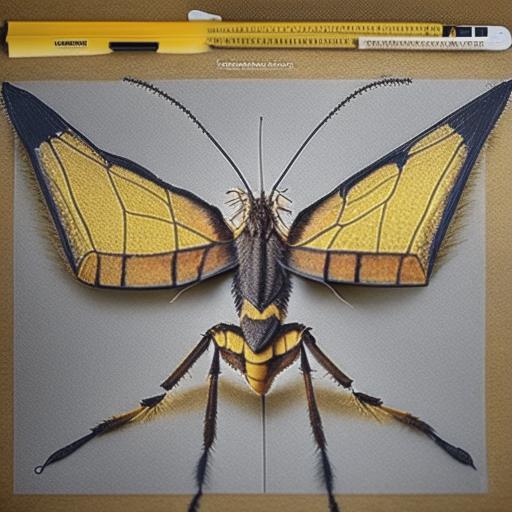} &
        \includegraphics[width=0.15\textwidth]{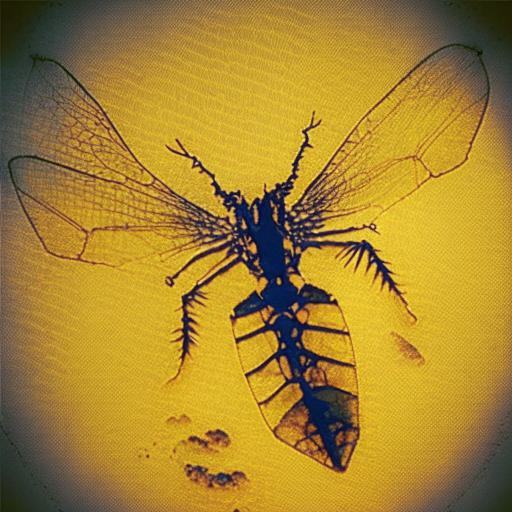} &
        \includegraphics[width=0.15\textwidth]{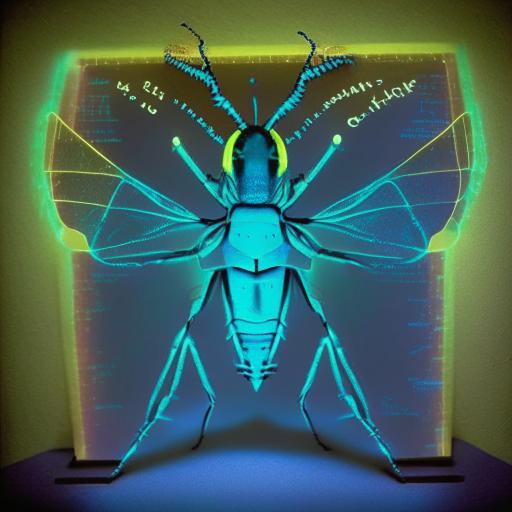} &
        \includegraphics[width=0.15\textwidth]{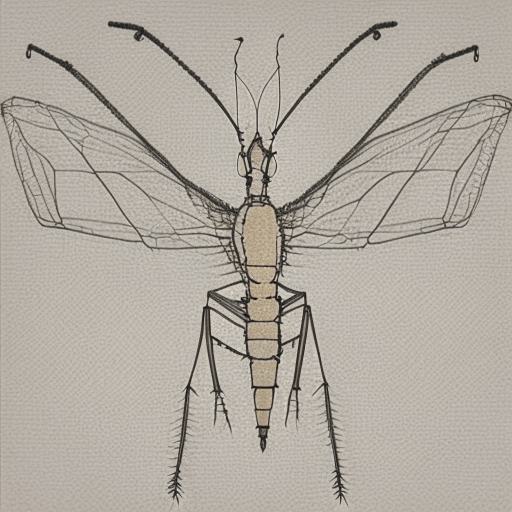} &
        \includegraphics[width=0.15\textwidth]{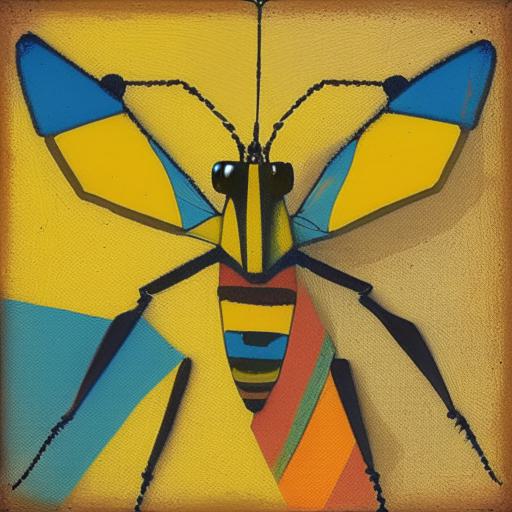} \\

        &
        \begin{tabular}{c} ``Professional high-quality photo of \\ a $S_*$. photorealistic, 4k, HQ'' \end{tabular} &
        \begin{tabular}{c} ``An amber fossil \\ of a $S_*$'' \end{tabular} &
        \begin{tabular}{c} ``A futuristic hologram \\ of a $S_*$'' \end{tabular} &
        \begin{tabular}{c} ``A line drawing \\ of a $S_*$'' \end{tabular} &
        \begin{tabular}{c} ``A painting of a $S_*$ \\ in the style \\ of Kandinsky'' \end{tabular}

        \\

    \end{tabular}
    }
    \caption{Sample text-guided creative generation results obtained with ConceptLab. The positive concept used for training is shown to the left. All results are obtained using our adaptive negative technique.}
    \label{fig:our_results_2}
\end{figure*}

\end{document}